\documentclass[times,twocolumn,final]{elsarticle}

\usepackage{medima}
\usepackage{framed,multirow}
\usepackage{float}
\usepackage{amssymb}
\usepackage{latexsym}
\usepackage{xcolor}

\usepackage[printonlyused]{acronym}
\usepackage{url}
\usepackage{xcolor}
\usepackage{graphicx}
\usepackage{subcaption}
\usepackage{adjustbox}
\usepackage{caption}
\usepackage{amsfonts} 
\usepackage{amsmath}
\usepackage{hyperref}

\usepackage[colorinlistoftodos,prependcaption,textsize=tiny]{todonotes}
\definecolor{newcolor}{rgb}{.8,.349,.1}

\usepackage{titlesec}
\titleclass{\subsubsubsection}{straight}[\subsection]

\newcounter{subsubsubsection}[subsubsection]
\renewcommand\thesubsubsubsection{\thesubsubsection.\arabic{subsubsubsection}}

\titleformat{\subsubsubsection}
  {\normalfont\normalsize}{\thesubsubsubsection}{1em}{}
\titlespacing*{\subsubsubsection}
{0pt}{3.25ex plus 1ex minus .2ex}{1.5ex plus .2ex}

\makeatletter
\renewcommand\paragraph{\@startsection{paragraph}{5}{\z@}%
  {3.25ex \@plus1ex \@minus.2ex}%
  {-1em}%
  {\normalfont\normalsize\bfseries}}
\renewcommand\subparagraph{\@startsection{subparagraph}{6}{\parindent}%
  {3.25ex \@plus1ex \@minus .2ex}%
  {-1em}%
  {\normalfont\normalsize\bfseries}}
\def\toclevel@subsubsubsection{4}
\def\toclevel@paragraph{5}
\def\toclevel@paragraph{6}
\def\l@subsubsubsection{\@dottedtocline{4}{7em}{4em}}
\def\l@paragraph{\@dottedtocline{5}{10em}{5em}}
\def\l@subparagraph{\@dottedtocline{6}{14em}{6em}}
\makeatother

\journal{Medical Image Analysis}

\begin{document}

\verso{João D. Nunes \textit{et~al.}}

\begin{frontmatter}

\title{A Survey on Cell Nuclei Instance Segmentation and Classification: Leveraging Context and Attention}%

\author[1,2]{João D. \snm{Nunes}\corref{cor1}}
\cortext[cor1]{Corresponding author: 
  e-mail.: joao.d.fernandes@inesctec.pt}
\author[3,4,5]{Diana \snm{Montezuma}}
\author[3]{Domingos \snm{Oliveira}}
\author[1,6]{Tania \snm{Pereira}}
\author[1,2]{Jaime \snm{S. Cardoso}}

\address[1]{INESC TEC - Institute for Systems and Computer Engineering, Technology and Science, R. Dr. Roberto Frias, Porto, 4200-465, Portugal}
\address[2]{University of Porto - Faculty of Engineering, R. Dr. Roberto Frias, Porto, 4200-465, Portugal}
\address[3]{IMP Diagnostics, Praça do Bom Sucesso, 4150‑146 Porto, Portugal}
\address[4]{Cancer Biology and Epigenetics Group, Research Center of IPO Porto (CI-IPOP) /
RISE@CI-IPOP (Health Research Network), Portuguese Oncology Institute of Porto (IPO Porto) / Porto Comprehensive Cancer Center (Porto.CCC), R. Dr. António Bernardino de Almeida, 4200-072, Porto, Portugal}
\address[5]{Doctoral Programme in Medical Sciences, School of Medicine and Biomedical
Sciences – University of Porto (ICBAS-UP), Porto, Portugal}
\address[6]{FCTUC - Faculty of Science and Technology, University of Coimbra, Coimbra, 3004-516, Portugal}
\received{1 May 2013}
\finalform{10 May 2013}
\accepted{13 May 2013}
\availableonline{15 May 2013}
\communicated{S. Sarkar}

\begin{abstract}
Nuclear-derived morphological features and biomarkers provide relevant insights regarding the tumour microenvironment, while also allowing diagnosis and prognosis in specific cancer types. However, manually annotating nuclei from the gigapixel Haematoxylin and Eosin (H\&E)-stained Whole Slide Images (WSIs) is a laborious and costly task, meaning automated algorithms for cell nuclei instance segmentation and classification could alleviate the workload of pathologists and clinical researchers and at the same time facilitate the automatic extraction of clinically interpretable features for artificial intelligence (AI) tools. But due to high intra- and inter-class variability of nuclei morphological and chromatic features, as well as H\&E-stains susceptibility to artefacts, state-of-the-art algorithms cannot correctly detect and classify instances with the necessary performance. In this work, we hypothesise context and attention inductive biases in artificial neural networks (ANNs) could increase the performance and generalization of algorithms for cell nuclei instance segmentation and classification.  To understand the advantages, use-cases, and limitations of context and attention-based mechanisms in instance segmentation and classification, we start by review works in computer vision and medical imaging. We then conduct a thorough survey on context and attention methods for cell nuclei instance segmentation and classification from H\&E-stained microscopy imaging, while providing a comprehensive discussion of the challenges being tackled with context and attention. Besides, we illustrate some limitations of current approaches and present ideas for future research. As a case study, we extend both a general instance segmentation and classification method (Mask-RCNN) and a tailored cell nuclei instance segmentation and classification model (HoVer-Net) with context- and attention-based mechanisms, and do a comparative analysis on a multi-centre colon nuclei identification and counting dataset. Although pathologists rely on context at multiple levels while paying attention to specific \acp{RoI} when analysing and annotating WSIs, our findings suggest translating that domain knowledge into algorithm design is no trivial task, but to fully exploit these mechanisms in ANNs, the scientific understanding of these methods should first be addressed.
\end{abstract}


\end{frontmatter}

\section{Introduction}

Cancer is a very heterogeneous disease, and as it manifests differently in each patient, conventional general treatments such as surgery, chemotherapy, radiation, and immunotherapy may not be effective for every cancer subtype, aside from the fact that these treatments are generally considered aggressive and also target healthy tissue. Particularly, cancer subtypes are associated with distinct genetic mutations and molecular and protein alterations. Therefore, when combined with proteomics, transcriptomics, and/or genomics, the visual analysis of tumour microenvironment from digitised \ac{HE}-stained \acp{WSI} is vital to help clinicians make informed decisions. Analysis of \ac{HE}-stained \ac{WSI} allows the characterization of tumour microenvironment and is of high clinical value for cancer diagnosis, treatment assessment, and prognosis. More specifically, precision and personalized medicine treatments have increasingly benefited from -omics biomarkers identified using these modalities. Furthermore, since \ac{HE}-stained \acp{WSI} are part of routine clinical practice, image-based phenotyping has the potential to replace complex and expensive diagnostic and prognostic solutions (e.g., molecular profiling) in precision and personalized medicine. In particular, the increasing migration of modern pathology towards digital workflows has allowed the introduction of novel solutions, supported by \ac{AI} algorithms, with unprecedented results \citep{Cui2021, Baxi2022, Rantalainen2020}.  But \ac{CPath} is a broad field that involves many tasks, including diagnosis, tissue characterisation, staging, grading, and disease prognosis \citep{Esteva2021}. Automatic tools that assist in these tasks usually call for \ac{CV} and \ac{ML} algorithms entailing regression analysis, classification, semantic segmentation, object detection, or instance segmentation and classification.  Depending on the objective of the task, multiple \ac{CV} and \ac{ML} models can be considered, but deep neural networks usually yield better performance and allow generalization to larger and more complex datasets \citep{Baxi2022}. Nonetheless, a promising strategy might be to incorporate prior clinical knowledge \citep{Marini2022, Neto2022, Aristizabal2022} into algorithm design. \ac{CPath} is thus a multidisciplinary field where pathologists and \ac{ML} scientists must work in close collaboration towards the design of algorithms with clinical value and that best exploit the potentialities of \ac{WSI} data with human-interpretable features. 

One of these clinically important tasks is the cell nuclei instance segmentation and classification\footnote{Throughout this work, we use "cell nuclei instance segmentation and classification" as an umbrella term that encompasses several tasks, namely: object detection, instance counting, localization, semantic segmentation, and instance segmentation and classification.}. This is relevant, for example, to the quantification of cells, characterization of the tumour microenvironment, and, in certain cancer types (e.g., breast cancer, lung cancer), nuclei-derived features have shown useful as diagnosis and prognosis biomarkers \citep{Lu2020, Yan2022}. However, nuclei instance segmentation and classification is not a trivial task. In the first place, there is intrinsically high intra- and inter-subject variability in tissue appearance and nuclei organization, even if we consider nuclei of the same cell type. And as different nuclei types can be morphologically similar, this task becomes even more challenging. Moreover, the digitised histology slides are highly prone to stain variability \citep{Tellez2018} and artefacts that negatively affect the generalization \citep{Campanella2019, Rolls2008}. Another bottleneck resides in the dimensionality of \acp{WSI}, which can be in the order of gigapixels, an undesirable attribute in hardware-limited settings. Moreover, each \ac{WSI} could have thousands of nuclei instances, which in voluminous datasets could scale to millions of nuclei (or billions of pixels) to annotate. Therefore, manually annotating every nuclei instance is practically impossible, aside from high inter-observer variability and annotation inconsistencies that hinder \ac{DL} models performance. To cope with these challenges, alternative and more label-efficient paradigms, like \ac{SSL}, \ac{WSL}, or \ac{SmSL} naturally emerge in an attempt to design \ac{ML} models robust to partially or incorrectly labelled data. 

In their daily practice, to overcome most of the previously described challenges in \ac{WSI} analysis and nuclei quantification, clinicians usually resort to \ac{WSI} visualization tools that allow them to scroll through the tissue image while zooming in and out, and while paying attention to different \acp{RoI} and tissue concepts (e.g., cells, nuclei, etc.). Through this process, clinicians aggregate information at multiple levels to make a decision regarding nuclear boundaries and types. In essence, this is a context-based decision process that we argue translates to increased annotation quality and that we hypothesise should be exploited to devise inductive biases in neural networks. Therefore, this survey reviews works with inductive biases for the spatial and semantic correlation between tissue concepts, i.e., with context inductive biases or that explore inter-instance relations. A representation encoding the co-occurrence and long-range dependencies of different objects (i.e., instances) could be more discriminative than a representation that only considers intra-instance features. Especially in the presence of ambiguities (e.g., due to noisy or blurred data, occluded objects, small instances, etc.). These ideas have already been investigated in the \ac{CV} field \citep{Naseer2021}. In turn, some context variables are confounders that bias \ac{DL} models to non-stable features \citep{Dreyer2023} (e.g., sky or grasslands to discriminate between birds and cows). But by using the self-attention mechanism to learn long-range dependencies (i.e., contextual information) between patch features, \ac{ViT}-based models could be less sensitive to spurious correlations when compared with \acp{ConvNet} \citep{Zhang2022ViTGen}. Overall, it is not clear that contextual information will always improveobject identification performance \citep{Zhang2020Causal, Liu2022ContextDebiasing, Ouyang2023}. When and how to use context priors is, in general, an unsolved topic in \ac{CV} and related subfields like cell nuclei instance segmentation and classification from \ac{HE}-stained brightfield microscopy images.  This survey is thus an attempt to help further researchers clarify the role of context and attention in nuclei instance segmentation and classification from \ac{HE}-stained brightfield microscopy imaging. Such knowledge could then be used to design more robust \ac{DL} models.

Besides this introduction, this work is structured as follows: In section \ref{sec:related_work}, we review previous related works and present our main contributions. In section \ref{sec:clinical_perspective}, we provide a clinical motivation for cell nuclei instance segmentation and classification, and in section \ref{sec:preliminaries}, we review the main metrics, datasets, and fundamental neural network architectures used in \ac{CPath} for  cell nuclei instance segmentation and classification. But for a comprehensive analysis of context- and attention-based mechanisms in  cell nuclei instance segmentation and classification, in section \ref{sec:context_attention_review}, we start by defining these two terms, review the taxonomies proposed in the literature for each paradigm, and discuss types and levels of context and attention. Following these definitions, we then review fundamental context- and attention-based mechanisms in \ac{CV}, including neural network modules, mechanisms, and even complete architectures that depend on the attention mechanism to incorporate multiple context levels, i.e., \acp{ViT}. The strategies suggested in these publications are frequently cornerstones of algorithms designed for specific domains and applications. But as these strategies are general, and often optimized for everyday scenes and objects (e.g., images of cats, cars, houses, etc.), due to the concept and covariate shift, they frequently fail in medical imaging data without fine-tuning or minor modifications. Therefore, we believe it is crucial to also understand how context and attention-based strategies are being adapted to solve medical imaging tasks. In particular, to help our discussion on  cell nuclei instance segmentation and classification, in section \ref{subsec:context_medical_imaging}, we review works that resemble that task, hoping that knowledge from other domains could help \ac{CPath} researchers understand the potentialities of context and attention-based mechanisms. Ultimately, that knowledge could help design novel solutions tailored for \ac{CPath} applications. However, the main scope of this work is a systematic review of context- and attention-based mechanisms in  cell nuclei instance segmentation and classification from \ac{HE}-stained brightfield microscopy imaging (section \ref{sec:context_attention_conic}). We provide a detailed analysis of state-of-the-art works, discuss how context- and attention-based mechanisms have been used to address key challenges in  cell nuclei instance segmentation and classification, identify promising research directions, and present alternative research ideas. In fact, for a more complete understanding of the impact of context- and attention-based mechanisms    in nuclei instance segmentation and classification, in section \ref{sec:comp_study}, we provide practical insights from a case study of how adapting Mask-RCNN \citep{He2017} and HoverNet \citep{Graham2019b} with context- and attention-based mechanisms affects the generalization of a nuclei instance segmentation and classification task. In section \ref{sec:conclusion}, we conclude with some final remarks regarding our findings and future directions.

\section{Related Work}
\label{sec:related_work}

Medical imaging is an immensely broad field and there is an increasing interest in \ac{CV} algorithms to analyse medical data, primarily owing to the growing availability of these imaging modalities, especially in fields like dermatology, radiology, and pathology \citep{Esteva2021}. An exhaustive review of context-, and attention-based \ac{CV} algorithms for medical data is thus impractical and beyond the scope of this work.  Furthermore, there is already an extensive body of research discussing \ac{DL} for medical image analysis \citep{Goncalves2022, He2022, Zhao2022, vanderVelden2022, Rao2021, Esteva2021, Lundervold2019, Liu2019}, but we differentiate from the prior art by, in addition to reviewing nuclei instance segmentation and classification methods, summarizing works on context- and attention-based mechanisms that have been successful in medical imaging tasks that share similarities with nuclei instance segmentation and classification. More specifically, we focus on methods that incorporate several of the following features: object detection, object segmentation (instance and semantic), instance classification, imbalanced instances, small and dense object detection or segmentation, or 2D inputs. We argue that such methods can serve as motivation for solutions tailored to pathology. Nonetheless, the main focus of the present review is to provide a systematic survey on context and attention-based mechanisms in \ac{DL}-based nuclei instance segmentation and classification from \ac{HE}-stained brightfield microscopy imaging. This review arises from a reflection on the methods used by pathologists during the analysis and annotation of tissue slides, as well as from domain knowledge on nuclei and tissue microenvironment in \ac{HE}-stained \acp{WSI}, which are context-rich.

Several authors aggregate and summarize the main works on  cell nuclei instance segmentation and classification, complementing their review with a critical analysis and discussion of future directions. \citet{Basu2023} review, from 2017 to 2021, \ac{DL}-based algorithms for cell nuclei segmentation with an emphasis on analyzing and comparing several models, namely, U-Net \citep{Ronneberger2015}, SCPP-Net \citep{Chanchal2021}, Sharp U-Net \citep{Zunair2021}, and LiverNet \citep{Aatresh2021}. Furthermore, the work highlights the importance of nuclei segmentation for \ac{CPath} including cell counting, diagnosis, and morphological analysis \citet{Basu2023}. \citet{Nasir2022}, on the other hand, discuss the main public datasets, grand-challenge algorithms, and future directions in gland and nuclei instance segmentation. The authors address works on both conventional \ac{ML} with manual feature extraction, and \ac{DL}, including encoder-decoder neural networks and generative models. Similarly to our work, they review several attention-based \ac{DL} methods, but no focus is given to context-based strategies or vision transformers.  \citet{Srinidhi2021deep} provide a comprehensive review of deep neural networks for \ac{CPath}. Indeed, the scope of their work is large, and although the paper mentions a few works on context- and attention-based mechanisms in \ac{DL} for \ac{CPath}, the scientific understanding of these mechanisms is still limited, and there is also the need to assess the usefulness of these methods specifically in nuclei instance segmentation and classification. \citet{Lagree2021}, on the other hand,  highlight several \ac{DL} methods for cell nuclei segmentation in breast tissue slides, trained on a multi-organ dataset. The methodology considers U-Net \citep{Ronneberger2015}, Mask-RCNN \citep{He2017}, and an ensemble of various U-Net-like models. But as their results underline the challenges of highly crowded nuclei, the authors propose an ensemble network (GB-UNet) and show that transfer learning is an effective strategy for nuclei segmentation in breast tissue WSIs \citep{Lagree2021}. Furthermore, \citet{Krithiga2021} review different algorithms in breast \ac{CPath}, namely methods for cell and nuclei instance segmentation and classification, feature extraction, image enhancement, and tissue classification, while \citet{Hayakawa2021} review nuclei instance segmentation and classification works with prominence on conventional \ac{ML} and on pre-processing and post-processing techniques.

In essence, our work differentiates from related reviews as we focus on surveying context and attention-based mechanisms not only in cell nuclei instance segmentation and classification but also in related medical imaging subdomains which could provide new insights regarding the effectiveness of these methods and serve as inspiration for the development of novel solutions. We consider this to be relevant as, regardless of the imaging technique, some common challenges of the medical imaging domain make solutions tailored to other imaging modalities of potential value to \ac{HE}-stained brightfield microscopy imaging. Nonetheless, not only are digitized \ac{HE}-stained brightfield microscopy images (mostly \ac{HE}-stained \acp{WSI}), widely available, as also \ac{HE} staining is one of the brightfield microscopy imaging techniques that provide the most detailed information, allowing clear visualization of tissue structures like the cytoplasm, nucleus, organelles and extra-cellular components. Therefore, this modality offers the best cost-effectiveness when compared with other microscopy techniques, which translates to a \ac{CPath} research focus on methods for \ac{HE}-stained image analysis. For these reasons, we give special attention to this imaging technique. Namely, we do a systematic review of context and attention mechanisms for \ac{DL}-based cell nuclei instance segmentation and classification from \ac{HE}-stained brightfield microscopy images, including \acp{WSI} and tiles, and provide a comprehensive discussion of how the different context levels have been addressed in the literature for cell nuclei instance segmentation and classification from \ac{HE}-stained brightfield microscopy images, while illustrating the different attention mechanisms that have been explored and their intended use like enriching the estimated feature maps with contextual information or reducing sensitivity to irrelevant background information.

Following the limitations of prior art, our main contributions are as follows:

\begin{itemize}
    \item A review of the main context- and attention-based \ac{DL} algorithms for \ac{CV} applications.

    \item A review of context- and attention-based works in medical imaging dense prediction tasks\footnote{A set of paradigms in computer vision that demand the mapping from input images to elaborate output structures, including instance segmentation and classification, semantic segmentation, and object detection and localization.}.
    
    \item A survey of 44 papers on context- and attention-based methods in cell nuclei instance segmentation and classification from \ac{HE}-stained brightfield microscopy (including \acp{WSI} and tiles). 

    \item A case study to empirically evaluate how extending reference cell nuclei instance segmentation and classification models with context and attention-based mechanisms affects generalization.

\end{itemize}

\section{A Clinical Perspective}
\label{sec:clinical_perspective}

Nuclei evaluation has long been relevant in Anatomical Pathology, as nuclei hold critical information regarding the identity and characteristics of the cell. Different cell types display unique nuclear features that help pathologists in classifying cells accurately. Furthermore, morphological alterations in the nucleus occur frequently in cancer, and other diseases, and can provide important information during diagnostic interpretation \citep{chow2012, gurcan2009}. In fact, for more than a century, pathologists have relied on morphological abnormalities of the nucleus as a crucial feature differentiating benign from malignant cells \citep{fischer2020}. These abnormalities comprise nuclear enlargement, elevated nuclear-to-cytoplasmic ratio, nuclear membrane irregularities, hyperchromasia, and altered chromatin distribution \citep{fischer2020}. Still today, the Nottingham histologic score for breast cancer grading relies on three parameters: tubule formation, mitotic activity and nuclear pleomorphism \citep{WHO2019}. The higher the score (1-3), the more aggressively the tumour is expected to behave. Recent work, by \citet{jaroensri2022}, has investigated the use of DL for histologic grading in breast cancer . Another example where nuclei evaluation is important is renal cell carcinoma, as the ISUP/WHO score is based on nucleolar prominence/eosinophilia for grades 1-3, and nuclear anaplasia for grade 4 \citep{delahunt2019}. In reality, nuclei evaluation is extremely relevant for most neoplastic diagnoses and plays a role in assessing disease severity, even if not included in a formal score. Moreover, nuclei evaluation holds importance beyond grading in diagnostics. For instance, diagnosing papillary thyroid carcinoma requires identifying distinctive nuclear features, concretely enlarged nuclei with irregular membranes, nuclear grooves and the characteristic optically clear nuclei (“Orphan Annie Eye” nuclei) \citep{Ibrahim2020}. There are already studies investigating \ac{ML} methods for thyroid cancer classification, applying nuclear segmentation \citep{bohland2021}. Apart from epithelial cells, nuclei evaluation is also relevant to assess/distinguish other cell types. In this context, scoring \acp{TIL} has proven prognostic value for breast cancer \citep{salgado2015} and also in other tumour types \citep{hendry2017}. As \acp{TIL} visual scoring is a time-consuming task, limited by intra and interobserver variability, software-guided \ac{TIL} analysis is currently under research, with promising results \citep{sun2021, aung2022, zhang2022}. Another domain that relies on nuclear assessment is \ac{IHC} stain scoring, as some \ac{IHC} stains target and color the cellular nuclei. Pathologists visually evaluate immunostains and estimate the percentage and intensity of staining, which encompasses diagnostic and predictive value. As such, exploring nuclear segmentation tasks has particular relevance in this setting, with several studies addressing this topic \citep{priego2022,mahanta2021,lejeune2013}. Moreover, multiple \ac{AI}/\ac{ML}-tools to automatically analyze immunostains are already commercially available. Lastly, nuclear evaluation relevance exceeds oncologic diseases, as Pathologists use nuclear evaluation as part of their diagnostic toolkit in a multitude of diseases (inflammatory, infectious, degenerative, etc.), beyond just cancer. In conclusion, by examining the cell nucleus, pathologists uncover valuable insights into disease processes, enabling accurate diagnoses and appropriate follow-up recommendations for patients, underscoring the importance of \ac{ML} methods focusing on nuclear segmentation and classification. 

\section{Preliminaries}
\label{sec:preliminaries}
In this section, we briefly discuss the main datasets, and relevant  cell nuclei instance segmentation and classification works that, although not explicitly incorporating context and attention mechanisms, are notable in the field. We also present the most common metrics to evaluate the quality of algorithms to solve dense prediction tasks like cell nuclei instance segmentation and classification (e.g., object detection, semantic segmentation, instance segmentation and classification, etc.).

\subsection{Datasets}

Table \ref{tab:datasets_nuclei_seg} summarizes the main annotated datasets for  cell nuclei instance segmentation and classification. Most of these datasets were initially proposed as part of \ac{ML} competitions \citep{Kumar2020, Sirinukunwattana2017, graham2023conic, Caicedo2019, Verma2020}, whose primary goal is to gather a community of \ac{ML} researchers to work on the same problem and to advance the state-of-the-art on nuclei instance segmentation and classification. In fact, data and coding competitions often translate to algorithmic performance breakthroughs, but a limitation is that little attention is paid to solving clinically relevant problems \citep{Varoquaux2022}. Nonetheless, the competition data is often publicly shared after the end of the challenges, meaning datasets become easily accessible for future research and benchmarking.  Top-performing \ac{ML} models are thus often trained and evaluated in one or several of these datasets. Moreover, although most works resorting to these data focus on the task of nuclei instance segmentation and classification, there are other emerging use cases like self-supervised representation learning \citep{Ciga2022, Song2023} (KUMAR \citep{Kumar2020}; CoNSeP \citep{Graham2019b} ), disease diagnosis \citep{Tourniaire2023, Bai2022, Zhou2019cgcnet} (DigestPath \citep{Da2022}; PanNuke \citep{Gamper2020pannuke}; CoNSeP \citep{Graham2019b}) and prognosis \citep{Dodington2021, Xie2022} (KUMAR  \citep{Kumar2020}), characterization of the tissue and tumour microenvironment \citep{Lu2020a, Zhao2023} (PanNuke \citep{Gamper2020pannuke}), \ac{DA} \citep{Li2022} (DigestPath \citep{Da2022}), explainable and interpretable \ac{AI} \citep{Graziani2020, Jaume2021, Mahapatra2022} (GlaS \citep{Sirinukunwattana2017}; KUMAR \citep{Kumar2020}), and biomarker discovery \citep{Fang2023} (PanNuke \citep{Gamper2020pannuke}). Nonetheless, as these data originate from diverse geographic locations, clinical sites, scanners, and tissue types, these are also of potential value to develop and evaluate \ac{DL} methods for \ac{DG}, continual learning, and even federated learning. Other promising use cases for \ac{HE}-stained images and corresponding nuclei annotations are hardware-efficient \ac{DL}, multimodal \ac{ML} (e.g., paired text-imaging data), and identification of imaging biomarkers.

Regarding the evaluation protocol, although the scientific publications that first introduce these data and corresponding nuclei instance segmentation and classification algorithms report the metrics they use, we identify two limitations, namely, some metrics  may not be appropriate to assess the segmentation quality of small instances \citep{Cheng2021boundary, Foucart2023}, and the heterogeneity across datasets of the metrics used to assess the quality of nuclei instance segmentation and classification methods. Besides, although splits are chosen carefully to not leak data from the training set to the validation and test splits, for most datasets only a portion of the tiles from the \ac{WSI} are selected. Following this observation, and noticing that a common practice in nuclei instance segmentation and classification algorithms is to quantify tile-level results, the question occurs regarding the performance for the complete \ac{WSI}. Besides, assuming not all tiles are effectively required to achieve a diagnosis, the question occurs regarding the optimal tile selection strategy for both training and testing. Although there is some work on sampling strategies for training \ac{DL} models for colorectal dysplasia grading \citep{Neto2023cad}, to the best of our knowledge, there is no similar strategy in other cancer types, especially considering segmentation-oriented paradigms. This is a relevant remark as reducing the subset of tiles used for training decreases a lot the computation costs, but if strong assumptions are imposed, it could also hinder \ac{DL} models generalization, especially in \ac{OOD} data. Answering these questions could be of service to the development of nuclei instance segmentation and classification algorithms with increased performance and generalization, while also increasing data and models' reusability and trustworthiness. 

Another concern regarding the data is sample selection bias, an undesirable phenomenon that occurs when certain elements or groups from a population are more likely to be included in a sample than others. This translates to datasets that are not representative of the target population, leading to flawed \ac{ML} algorithms that favour spurious correlations instead of stable features. In \ac{CPath}, batch effects are a common form of sample selection bias. Batch effects refer to variation between two batches of data due to differences in slide preparation, or digitizer. For instance, as different providers adopt different hardware, algorithms for image manipulation, and file formats \citep{Janowczyk2019}, it is expected digitizer-related batch effects, including variations in scale and colour. Besides, colour batch effects can also occur due to differences in staining protocol, room temperature during stain preparation, the clinician that prepared the slide, variations of thickness in tissue sections, and different lots or manufacturers of staining reagents \citep{Niazi2019}. These batch effects can therefore be site-specific, thus contributing to sampling bias, aside from different patient demographics. Indeed, recent work \citep{AsilianBidgoli2022} demonstrates that even in datasets with samples from multiple sources, \ac{DL} models become biased towards site-specific signatures, learning to discriminate the cancer type from spurious associations. To help identify batch effects and sample selection bias, we suggest future dataset releases should be accompanied by detailed reporting of data acquisition protocols, along with information on preprocessing, and storage procedures. Ultimately, sample selection bias is one of the potential causes of hidden stratification. Hidden stratification occurs when a model has good average performance in the entire test set, but fails on some specific subgroups characterized by a combination of low prevalence of disease subtype, incorrect or inconsistent annotations, sparse discriminative features, and spurious correlations \citep{Oakden2020}. Hidden stratification is thus closely related to domain shift, helping clarify why models with good performance in the \ac{iid} setting fail on \ac{OOD} samples. The most effective system to mitigate hidden stratification is to provide, along with the data, metadata to help identify the relevant subsets. For instance, certain samples that originate from the TCGA database \citep{Clark2013} have some demographic variables available like gender, race, age at diagnosis, and smoking habits. These variables could help identify model biases, however, these are not provided along with the annotated nuclei subsets from TCGA \citep{Clark2013}, a major drawback impeding a thorough evaluation of nuclei instance segmentation and classification algorithms. In fact, for safe \ac{ML} deployments in medicine, measurement and reporting of hidden stratification effects should be part of the models' evaluation protocol \citep{Oakden2020}. But we observe this is not addressed on datasets for cell nuclei instance segmentation and classification, as these lack metadata to help detect biases and identify relevant subsets. One form of overcoming this limitation is unsupervised learning, namely clustering techniques  which could be used to both identify relevant subgroups of the data, or as part of evaluation protocols to assess human annotation of existing subclasses \citep{Oakden2020}.

A final consideration regards the annotation pipeline. While most datasets report the annotations were performed, refined, and reviewed by multiple experienced clinical annotators, \ac{DL} for nuclei instance segmentation (and for medical image analysis in general) is a multidisciplinary field that demands iterative, and close collaboration between \ac{ML} researchers and pathologists \citep{Montezuma2023, Graham2021a} so as to reduce mislabelling or systematic errors that undermine the performance of \ac{AI} algorithms. On the other hand, the lack of standard annotation protocols could translate to annotation bias as a result of the annotator's (e.g., pathologist) experience, judgment, level of fatigue, difficulty in annotating specific tissue concepts (e.g., cell nuclei) and regions \citep{Wahab2022}. Moreover, while \ac{AI}-driven semi-automatic tools are a valuable resource in reducing the annotation burden, these could bias the pathologist to agree with the tool on tissue regions and concepts that humans cannot easily discriminate \citep{Ibrahim2020}. Moreover, following the FAIR data principles \citep{Wilkinson2016} and guidelines for \ac{AI} in medical imaging \citep{Mongan2020, Montezuma2023}, qualitative and quantitative evaluations of annotation quality should be done, as well as reporting of quality control measures. Most of the discussed datasets describe some quality control steps, but we find varied practices with different levels of quality assessment, meaning there is no standard protocol to maximize confidence in the annotations \citep{Wahab2022, Montezuma2023}, nor standard quantitative measures to estimate data and annotation quality    in nuclei instance segmentation and classification datasets. In essence, this could translate to limitations in the knowledge drawn from algorithms trained on those data. This is something we argue should be more thoroughly addressed in future public dataset releases.

\begin{table*}[t]
\captionsetup{font=normalsize}
\begin{adjustbox}{width=\textwidth,center}
\begin{tabular}{cccccccccc}
\hline
\textbf{Dataset} & \textbf{Pub. Test Set $\dagger$ } & \textbf{\# WSIs} & \textbf{\# Annot. Nuclei} & \textbf{Source}                                    & \textbf{Scanner} & \textbf{Mag.} & \textbf{Ref.} & \textbf{\# Cit. $\ddagger$} & Year of Pub. \\ \hline
CoNSeP           & yes & 16              & 24,319                      & Univ. Hosp. Coventry and Warwickshire, UK & Omnyx VL120  & $ 40 \times $ &  \citep{Graham2019b} & 687 & 2018 \\
KUMAR         & yes & 30                & 21,623                      & TCGA        &          Multiple        & $ 40 \times $   &  \citep{Kumar2020} & 776 & 2018 \\
GlaS             & no & 16              &     55,364*    &    Univ. Hosp. Coventry and Warwickshire, UK & Zeiss MIRAX MIDI &   $ 20 \times $   &  \citep{Sirinukunwattana2017} & 671 & 2016 \\
DigestPath       & no &  476               & 168,510*  & Multiple Hospitals (4), China & KFBIO FK-Pro-120 &   $ 20 \times $  & \citep{Da2022} & 28 & 2019 \\
TNBC             & yes & 11              & 4,022      &   Curie Institute, Paris, France  &  Philips Ultra Fast Scanner 1.6RA            &  $ 40 \times $ &  \citep{Naylor2018} & 468 & 2018 \\
CPM-17            & yes & 32              &            7,570                 & TCGA                                               &    Multiple  &    $ 20/40 \times $  &  \citep{Vu2019} & 218 & 2017 \\
CCa              & yes & 10              & 29,756                      &   Univ. Hosp. Coventry and Warwickshire, UK & Omnyx VL120      &  $ 20 \times $  & \citep{Sirinukunwattana2016} & 1235 & 2016 \\
Lizard & yes &   NA & 431,913                     & Multiple                                           & Multiple                &  $ 20 \times $  &  \citep{Graham2021a} & 87 & 2019 \\
MoNuSAC     & yes &  46  &  46,909  &  TCGA  &  Multiple   &  $ 40 \times $  &  \citep{Verma2020} & 86 & 2020 \\
PanNuke     & yes &   481    &   205,343 &   Multiple &   Multiple  &  $ 40 \times $  & \citep{Gamper2020pannuke} & 286 & 2020 \\ 
\hline
\end{tabular}
\end{adjustbox}
\captionof{table}{Overview of nuclei instance segmentation and classification datasets. \label{tab:datasets_nuclei_seg} * Annotations provided by the Lizard dataset. $\dagger$ \ac{HE}-stained images and cell nuclei annotations. $\ddagger$ Google Scholar citations (Last reviewed December 2023). }
\end{table*}

\subsection{Evaluation Metrics}

\begin{itemize}
    \item \textbf{Intersection Over Union}
\end{itemize}

The \ac{IOU}, also known as Jaccard Index, quantifies the overlap between the detected/segmented objects and the ground truth. It is defined as 

\begin{equation}
    IoU = \frac{|P \cap G|}{ |P \cup G|}
\end{equation}

\noindent with $P$ and $G$ predicted and target objects.

\begin{itemize}
    \item \textbf{Aggregated Jaccard Index}
\end{itemize}

The \ac{AJI} is an extended version of the Jaccard Index designed for instance segmentation tasks. Aside from considering the overlap between each instance in the prediction and target images, it contains a term to penalize false positive detections. It therefore simultaneously penalizes object-level (missed detection, false positive detection) and pixel-level (under/over-segmentation) errors. Formally, we have

\begin{equation}
    AJI = \frac{\sum_{i=1}^N |P_i\cap G_i|}{\sum_{i=1} |P_i\cup G_i| + \sum_{j=1}^M |FP_j|}
\end{equation}

\noindent with $P_i$ and $G_i$ predicted and target objects out of $N$ instances, and $FP_j$ a false positive detection out of $M$ predicted instances.

\begin{itemize}
    \item \textbf{Panoptic Quality}
\end{itemize}

The \ac{PQ} \citep{Kirillov2019} is a metric to jointly evaluate the quality of instance and semantic segmentation. It is composed of two terms, one to assess the detection quality, and the other to evaluate the segmentation quality. For a target class, $t$, \ac{PQ} is formulated as

\begin{equation}
    \mathcal{PQ}_t = \underbrace{\frac{|TP_t|}{|TP_t|+ \frac{1}{2}|FP_t|+\frac{1}{2}|FN_t|}}_\text{Detection \: Quality (DQ)} \times \underbrace{\frac{\sum_{(x_t, y_t) \in TP} IoU (x_t, y_t)}{|TP_t|}}_\text{Segmentation \: Quality (SQ)}
\end{equation}

\noindent with $TP_t$ the set of predicted instances of class $t$ that is correctly classified, $FN_t$ the set of instances predicted as negative for class $t$ but that should be positive, and $FP_t$ the set of instances predicted as class $t$ but that are not correctly classified.  

\ac{PQ} is computed as the average of $\ac{PQ}_t$ over all target classes. For a test dataset, $\mathcal{D}$, \ac{PQ} is averaged over all images in $\mathcal{D}$. Another version of \ac{PQ} exists, the multiclass $\mathcal{PQ}$, referred to as $m\mathcal{PQ}^+$, which is the average of $\mathcal{PQ}_t$  computed over all instances in a dataset. This metric is more robust to inconsistencies in images where a particular class is not present, thus naturally resilient to imbalanced data.

\begin{itemize}
    \item \textbf{Mean Average Precision}
\end{itemize}

The \ac{mAP}, for a $n$ class problem, is formally described as

\begin{equation}
    mAP = \frac{1}{n}\sum_i AP_i,
\end{equation}

\noindent where the \emph{AP} is computed as the area under the \emph{precision-recall}, $p(r)$, curve

\begin{equation}
    AP = \int_0^1 p(r) dr
\end{equation}

\noindent the precision defined as the fraction of true positives among the detected instances

\begin{equation}
    \label{eq:precision}
    Precision = \frac{TP}{TP+FP}
\end{equation}

\noindent and recall as the fraction of true positives that are detected:

\begin{equation}
    \label{eq:recall}
    Recall = \frac{TP}{TP+FN} \cdot
\end{equation} 

The precision-recall curve represents the trade-off between Precision and Recall at different thresholds. A high \ac{AUC} means high precision and high recall. The \ac{mAP} metric is widely adopted in object detection and segmentation tasks, where a \ac{mAP} at $50\%$ \ac{IoU} means that a detected/segmented object is accepted as a true positive if at least $50\%$ of the area covered by both objects represents their intersection.

\begin{itemize}
    \item  \textbf{Dice Similarity Coefficient}
\end{itemize}

The \ac{DC}, or S{\o}rensen-Dice index, is a measure of the overlap between the prediction and ground truth segmentation maps, one of the most common metrics used to assess the quality of image segmentation algorithms. Formally, the \ac{DC} is defined as

\begin{equation}
    DC = \frac{2|P \cap G|}{ |P| + |G|}
\end{equation}

\noindent with $P$ and $G$ the predicted and target segmentation maps.

\begin{itemize}
    \item \textbf{Symmetric Hausdorff Distance}
\end{itemize}

The symmetric \ac{HD} derives from the \ac{dH}, an asymmetric measure of the distance between two finite sets of points. Given two finite sets of points (e.g., predicted and ground truth segmentation maps), $ \mathcal{A} =\{ a_1, a_2, ..., a_n\}$, and $ \mathcal{B} = \{b_1, b_2, ..., b_n\}$, the \ac{dH} between $\mathcal{A}$ and $ \mathcal{B} $ is defined as the maximum distance between each point of $ \mathcal{A} $ to its closest point in $ \mathcal{B} $. More formally, we have

\begin{equation}
    \label{eq:hd}
    d_{H}(\mathcal{A}, \mathcal{B}) = \max_{a \in \mathcal{A}} \{ \min_{b \in \mathcal{B}} \{ d(a, b) \} \}
\end{equation}

\noindent where $ d(a, b) $ is a metric between these points, typically the Euclidean distance. The symmetric \ac{HD} can then be expressed in terms of \ref{eq:hd}:

\begin{equation}
    \label{eq:ahd}
    HD(\mathcal{A}, \mathcal{B}) = \max \{d_H(\mathcal{A}, \mathcal{B}), d_H(\mathcal{B}, \mathcal{A})\}
\end{equation}

This metric is useful to assess the quality of predicted contours and, thus, particularly useful to evaluate the segmentation quality of highly overlapped instances and other fine-grained nuclei segmentation tasks.

\subsubsection{Discussion}

These metrics are frequently used in the literature to evaluate the performance of  cell nuclei instance segmentation and classification algorithms. Nonetheless, there is no gold standard for model evaluation, which represents a bottleneck to directly comparing the results of related works. Moreover, as nuclei instances are frequently small and/or overlapped, there are a few limitations with some of these metrics. The \ac{IoU} over-penalizes errors in small objects \citep{Cheng2021boundary}, is asymmetric with regard to penalizing over and under-estimation \citep{Foucart2023}, and does not consider object shape, i.e., nuclei morphology, which is a relevant feature in \ac{CPath}. On the other hand, this also affects the \ac{PQ} metric, which is based on the \ac{IoU} \citep{Foucart2023}. Other limitations of the \ac{PQ} include the attribution of a more aggressive penalty for good but misclassified detections than for missed detections \citep{Foucart2023}, whereas summarizing the results of a complex multi-objective task into a single metric hinders interpretability of the results. In turn, the \ac{DC} does not consider the correctness of object boundary \citep{Reinke2023}, and is more sensitive to misclassifications in small instances \citep{Reinke2021}. Therefore, per task and per class metrics are strongly recommended \citep{Foucart2023}, while other more qualitative approaches could be considered. For instance, recent work has suggested a more robust and clinically significant evaluation protocol \citep{graham2023conic} where interpretable features (morphological, colocalization, and density features) extracted from the segmented nuclei at the \ac{WSI}-level are used for dysplasia grading and survival analysis.

\subsection{Neural Network Architectures}

In general, context and attention-based neural networks, namely \acp{ViT} \citep{coco_pwcode2023, cityscapes_2023, GuKong2022, Jung2022, Khan2022} frequently correspond to top-performing methods for object detection, semantic segmentation, and even instance segmentation and classification. However, in cell nuclei instance segmentation and classification, we verify that specialized neural networks based on conventional \acp{ConvNet} are the most successful approaches \citep{graham2023conic, Weigert2022, Doan2022}, although \acp{ViT} have gained increasing interest \citep{Haq2022, LuoSelvan2022, Zhang2023}. In this subsection, we review some of these works.

HoVer-Net \citep{Graham2019b} is a dedicated multi-branch neural network for nuclei instance segmentation and classification. It consists of a Pre-Activated ResNet-50 \citep{He2016} encoder, also called the trunk, and three decoding branches, each responsible for a different prediction. The \ac{NC} branch predicts a nuclei class for each pixel, the \ac{HoVer} branch is responsible for regressing the horizontal and vertical distance maps of each pixel to its object centre of mass, and the \ac{NP} branch assigns each individual pixel to either belonging to a nucleus (1) or background (0). A Sobel filter on the output of the \ac{HoVer} branch is reflective of individual nuclei instances. These gradient maps have high values at the border of neighbouring nuclei and, therefore, after adequate thresholding can be used in conjunction with the NP branch output to compute markers and an energy landscape for a marker-based watershed algorithm \citep{Graham2019b}. The instance segmentation predictions are then combined with the pixel class predictions, through majority voting, to achieve nuclei instance segmentation and classification. 

SONNET \citep{Doan2022} is a tri-decoder architecture that predicts the \ac{NF} and \ac{NT}, as well as performs \ac{NO}. The ground truth for the \ac{NO} task consists of a \ac{DDD} map, computed from the Euclidean distance of each pixel to its nucleus center of mass. The \ac{DDD} map discretizes the distances into $K$ sub-intervals, with the length of the sub-intervals an exponentially decreasing function of the pixel distance to the nucleus center of mass. This approach serves a two-fold purpose. First, the pixels near the center can be used to separate overlapping instances. Second, the pixels near the boundary indicate more uncertain regions, therefore, the authors propose a loss weighting scheme such that a misclassification in pixels further from the origin is more penalized. Besides, to obtain the final instance segmentation map, the design also includes a post processing strategy on the outputs of the three decoders.

\cite{Liu2022spn_ien} introduce a weakly supervised two stages method (SPN+IEN) that leverages estimated semantic segmentation maps to generate instance level pseudo-masks. It consists of a \ac{SPN} to estimate the background and cell nuclei masks, and an \ac{IEN} which outputs instance-level embeddings. Both IEN and SPN consist of U-Net-like architectures. A k-means clustering algorithm is used to generate cell nuclei pseudo-labels to train the SPN, while Voronoi partitioning of the SPN output is used to get instance pseudo-masks used in optimizing the IEN module. The final instance segmentation result derives from mean-shift clustering the instance embeddings aggregated with the SPN output.

StarDist \citep{Schmidt2018, Weigert2022} supports itself on the principle that star-convex-polygons are well-suited to approximate cellular and nuclear shapes. Namely, for every pixel in the input image, the network regresses its distance to the furthest point in the object it belongs, along predefined equidistant radial directions. Besides, the network simultaneously assigns a probability for every pixel reflecting if it is a foreground object or background. The neural network output, after \ac{NMS}, thus consists of polygon candidates, expressing cell/nuclei instances, with underlying object probabilities. The authors suggest this approach is well suited for dense nuclei instance segmentation, while having also extended the work to perform classification \citep{Weigert2022}.

CIA-Net \citep{ZhouBao2019} addresses several of the nuclei instance segmentation challenges, primarily overlapping nuclei, noisy inputs, and stain variability. The proposed neural network architecture is based on an encoder-decoder network but considers two decoder branches and dense convolutional blocks in the encoder path. While one of the decoders is responsible for predicting nuclei segmentation, the other predicts nuclei contours. However, the model assumes features learned for each individual task are useful for the other task, where a learnable bidirectional \ac{IAM} \citep{ZhouBao2019} between the two branches is designed to improve the network ability to jointly refine nuclei and contour predictions. A custom loss function is also proposed, which conditions the model to focus on informative samples to reduce sensitivity to input perturbations (i.e. noise).

Although typical segmentation algorithms achieve good results in nuclei semantic segmentation tasks, they become error-prone on instance segmentation, especially in the presence of highly overlapped nuclei. This condition is typical of advanced cancer stages \citep{Koohbanani2019}. SpaNet is thus an instance segmentation and classification design \citep{Koohbanani2019} that leverages spatial information to compensate lack of spatial awareness in conventional algorithms \citep{Koohbanani2019}. The model first predicts a binary nuclear segmentation map and nuclei centroids, using a dual-head neural network. These outputs are subsequently combined with spatial encoding, and 6 colour channels, namely in RGB and HSV colour spaces, and used as input to SpaNet to make instance segmentation predictions. SpaNet is an encoder-decoder network, with skip connections from the encoder to the decoder, but that replaces the typical U-Net \citep{Ronneberger2015} convolutional blocks with  \ac{MSDU} \citep{Koohbanani2019} to encode multi-resolution information into the feature maps. A post-processing routine based on clustering methods is then employed on the SpaNet output to produce the final instance segmentation maps. 

As encoder-decoder neural networks are prone to loss of spatial information due to pooling operations, \citet{Qu2019} propose a full-resolution network (FullNet), composed of only dense layers with no stride or pooling and, subsequently, without any upsampling module. As a mimicry for the pooling operation, the authors suggest dilated convolutions that also allow increasing the receptive field size. Besides, the optimization procedure resorts to variance cross-entropy loss, which considers intra-instance statistics, thus working as a spatial constraint on pixels belonging to the same instance \citep{Qu2019}. The authors suggest this loss function promotes the network to be sensitive to object shapes. 

Another specialized neural network is the MicroNet model \citep{Raza2019}, which is introduced as a versatile algorithm for the segmentation of several entities in microscopic images, at multiple scales, and in different pathology applications. The network receives multiple inputs, each corresponding to a downsampled version of the same tissue image, thus encoding multi-scale context to predict the underlying segmentation maps \citep{Raza2019}. More precisely, the model contains 5 multi-branch groups. The first group is an encoder module responsible for extracting multi-scale feature maps. Group 2 is a typical bridge layer between the encoder and decoder modules, whereas group 3 corresponds to the upsampling path, with lateral connections to the encoder. Group 4, on the other hand, outputs multi-scale feature maps that are subsequently aggregated and processed in group 5 to predict the segmentation masks. Overall, this model can be fine-tuned for several cell/nuclei segmentation tasks \citep{Raza2019}.

\section{Context And Attention-based Methods for Dense Prediction Tasks}
\label{sec:context_attention_review}

This section reviews context- and attention-based mechanisms in \ac{CV} applications, specifically in dense prediction tasks, including medical imaging applications. Aside from summarizing the main methods, we comment on how the reviewed publications address critical points in \ac{ML} research, namely considering methodological best practices, reproducibility, \ac{OOD} generalization, and validity of the drawn conclusions. 

\subsection{Building Blocks: Computer Vision in General}

 The idea of incorporating attention mechanisms in deep neural networks derives from the intuition that, to recognize complex objects or scenes, biological beings with intelligence subconsciously pay more attention to discriminative parts of the visual field while disregarding redundant or irrelevant information. Several works have reviewed attention in \ac{CV} but the work of \cite{Guo2022} has proposed a way of schematizing the different attention mechanisms in \ac{CV}. They start by proposing a general definition for attention, A,  

\begin{equation}
    \label{eq:attention}
    A = f (g(x), x),
\end{equation}

\noindent where $ g(x) $ represents attention and $ f(g(x), x) $ means processing input $x$ based on attention. A taxonomy is then suggested to organize different kinds of attention mechanisms  in \ac{CV} \cite{Guo2022}:

\begin{itemize}
    \begin{item}
    \textbf{Channel Attention}: In deep neural networks, we can interpret each feature map, at multiple scales, as a hierarchy of visual representations, where each channel is a unique semantic encoding of the input. As such, a channel attention mechanism defines what to pay attention to, i.e., it intends to recalibrate the importance of each semantic representation depending on what is being represented.
    \end{item}
    \begin{item}
    \textbf{Spatial Attention}: This term is more related to the structure and spatial context than to the semantics. The spatial attention mechanism intends to refine the importance of different visual regions for representing the input by exploring where to pay attention. 
    \end{item}
    \begin{item}
    \textbf{Temporal Attention}: For time-dependent input representations (e.g., video, time-series, etc.), this mechanism is a means to establish when to pay attention, i.e., it seeks to reinforce selectivity for representations at specific time points. 
    \end{item}
    \begin{item}
    \textbf{Branch Attention}: Specific to branched neural networks, it seeks to refine the importance of each branch, thus answering which to pay attention to.
    \end{item}
    \begin{item}
    \textbf{Spatial and Temporal/Channel Attention}: Tries to simultaneously refine spatial and temporal/channel representations. 
    \end{item}
\end{itemize}

\begin{figure}[h]
  \centering
 \includegraphics[scale=0.6]{"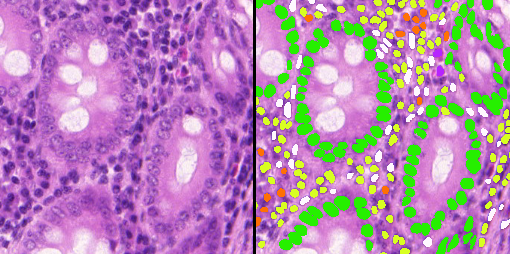"}
  \caption{Example of a \ac{WSI} tile from the Lizard dataset (\ac{CRC}) \citep{Graham2019a}. Left: Original tile. Right: Cell nuclei overlay. We observe that digital pathology images have rich contexts, where cells organize by type in cellular communities with relatively well-defined geometries and orientations. For instance, we observe that nuclei from epithelial cells (green) are organized in an oval geometry and that all other cells are located outwards to these, and orientated in different directions.}
  \label{fig:context_digipath}
\end{figure}

In turn, visual context helps recognize objects more accurately and faster. Context regards all critical information of a visual scene that helps identify an object, such as location, relative size and position to other objects, co-occurrence, or scene semantics. Pathologists heavily rely on contextual information when identifying nuclei in a \ac{WSI}. As cell nuclei present high intra and inter-class variability, context facilitates decision-making as nuclei rarely appear in isolation, but rather in cellular communities, with moderately well-defined positions and orientations to one another and in semantically rich scenes. Fig. \ref{fig:context_digipath} represents an example of colon cell nuclei in a tile extracted from a \ac{HE}-stained \ac{WSI}. For instance, we observe that nuclei from epithelial cells (green) are organized in an oval geometry and that all other cells are located outwards to these and orientated in multiple directions.  

But generalizing to the \ac{CV} domain, a survey by \cite{XWang2022} suggests a structured way of defining context. The authors propose context can be organized into levels and types. Levels include prior knowledge, global context, and local context. We propose to extend the subdivision of context levels also to include long-range and multi-scale contexts. Considering a visual scene, where each object and object parts can be described by a set of independent visual representations, i.e. features, we define context in \ac{CV} as another set of visual representations that encodes co-occurrence, relative position, and appearance to other objects and scene and that affect the semantic meaning of the visual scene and its parts. Following this definition, we propose the following taxonomy for context levels:

\begin{itemize}
    \item \textbf{Prior knowledge:} Refers to \emph{a priori} knowledge obtained independently of any particular observation. In computer vision, it represents what to expect in a particular scene or domain \citep{XWang2022}. 
    \item \textbf{Global Context:} Concerns visual representations of the scene as a whole, namely, global semantic and spatial features \citep{XWang2022}.
    \item \textbf{Local Context:} Considers short-range spatial and semantic relations between the visual representations of an instance or neighbouring instances \citep{XWang2022}.
    \item \textbf{Long Range Context:} Applies to the relations between the visual representations of distant objects, and object parts.
    \item \textbf{Multiscale Context:} Defined as the relation between the representations of the parts and the whole and between the representations of the same concept observed at different levels of magnification (e.g., relations between the visual representations of a constellation observed with $20$ and $30 \times$ per inch of aperture, or the relationship between the features of a \ac{HE}-stained \ac{WSI} at $10$ and $ 20 \times$ magnification.
\end{itemize}
    
\begin{figure*}[h]
\centering
    \begin{subfigure}[t]{0.45\textwidth}
    \centering
    \includegraphics[scale=0.25]{"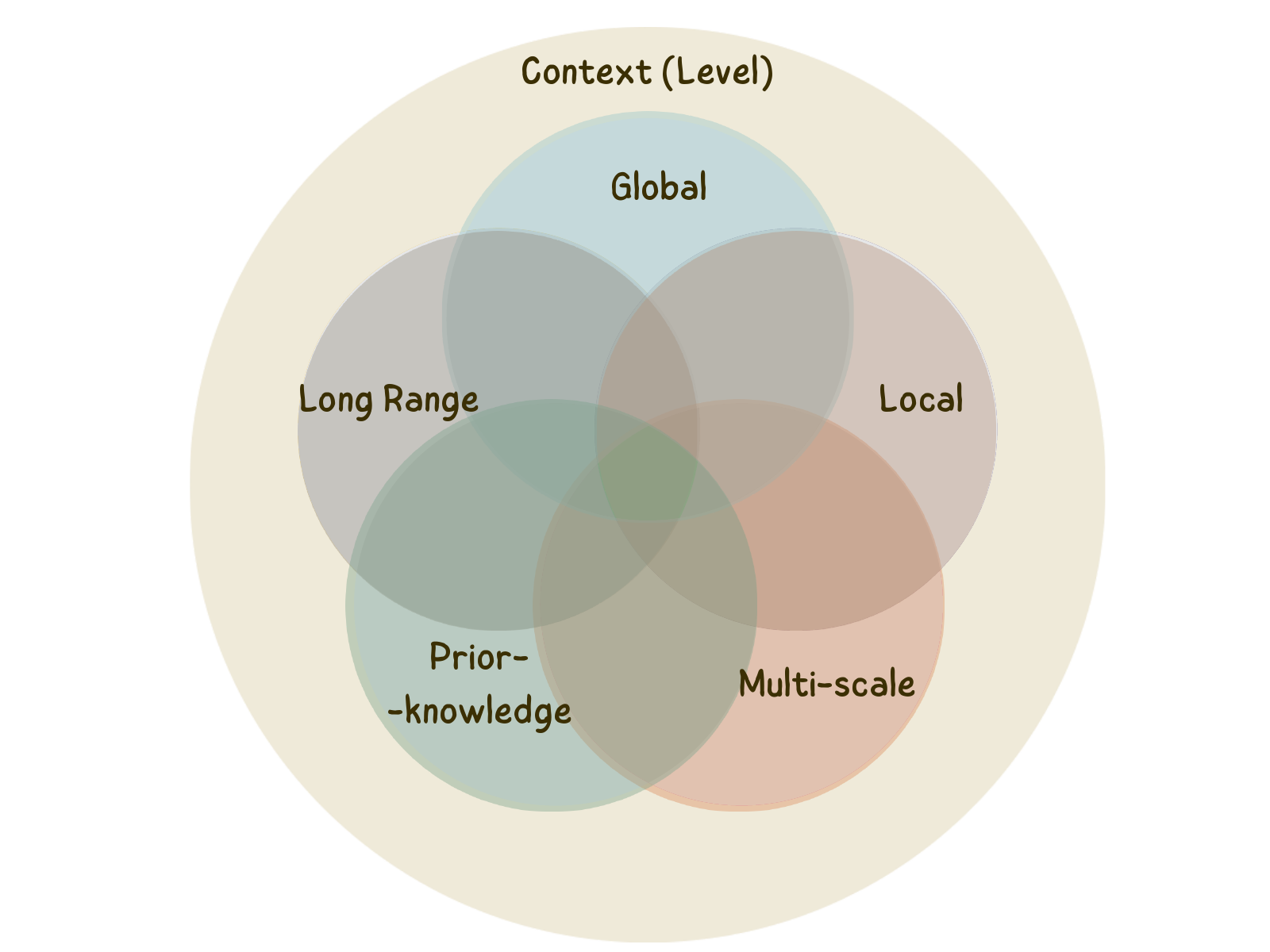"}
    \caption{Context by levels.}
    \end{subfigure}
    \begin{subfigure}[t]{0.45\textwidth}
    \centering
    \includegraphics[scale=0.25]{"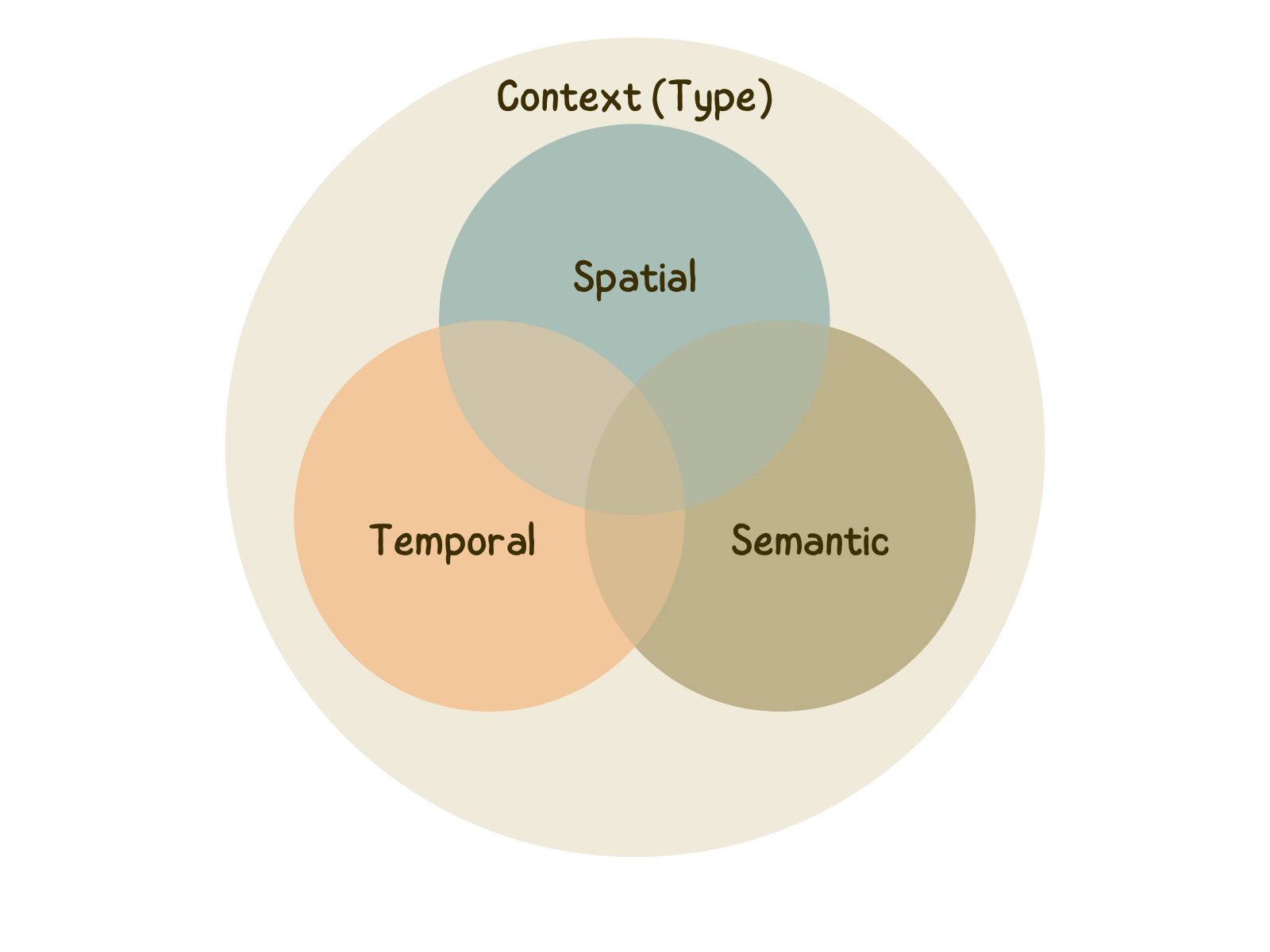"}
    \caption{Context by types.}
  \end{subfigure}
     \caption{A structured way of defining context in computer vision. \label{fig:context_types}}
\end{figure*}

In turn, concerning type, context can be aggregated into spatial, temporal, and semantic context \citep{XWang2022}. Spatial context regards the location of an object in a scene and the relative position of other objects and visual regions. It relates to the co-occurrence of different objects, and of object parts, as well as to spatial relations, including directions (e.g., object A is left of object B), and topological relations (e.g., object A contains object B) \citep{XWang2022}. Spatial semantic context, in turn, considers the likelihood of an object appearing in a particular scene and the specific semantic relationships with other objects (e.g., a pedestrian is likely to appear on top of a marked crosswalk if the traffic light for pedestrians is green). Beyond the scope of the present survey, there are also temporal and semantic temporal contexts, which consider time-dependent relationships between the features of a visual scene or its parts. Fig. \ref{fig:context_types} summarizes the characterization of context levels and types in \ac{CV}.

An important caveat of the present survey is what we define as context-based mechanisms in \ac{DL}. To make clear the notion of context-based mechanisms that we intend to discuss, we define it as \emph{a set of ANN architectures, layers, modules, training recipes, or regularization strategies designed with the goal of maximizing contextual information in learned representations}. This definition stands from the need to distinguish between the implicit context learned by common ConvNets like DeepLabv3 \citep{Chen2017} or U-Net \citep{Ronneberger2015}, and other more context-tailored strategies. In the following subsections, we review what we consider the main works in \ac{CV} that propose context and attention-based mechanisms for image classification and object detection and segmentation tasks. Many other works extend these building blocks to propose specialized architectures.

\subsubsection{Squeeze and Excitation Block}

Motivated by the idea that the representation capacity of \acp{ConvNet} could be increased by capturing correlations between features, \cite{Hu2018} introduce an explicit mechanism to model inter-channel dependencies, the so-called \ac{SE} Block. The central idea of this mechanism is to perform feature refinement, supported by global information, where the most discriminative features are enhanced, and less informative features suppressed. To circumvent the intrinsic locality of convolution operations, the authors propose to squeeze the spatial dimension of the feature vectors to a single unit, through Average Pooling. The resultant feature vector, $F_{avg}$, is thus an aggregation of local descriptors that encodes the whole image. Then, they perform a feature recalibration operation on $F_{avg}$. The authors describe this step as “a gating mechanism with sigmoid activation” \citep{Hu2018}. The operation is thus an adaptive recalibration mechanism, capable of capturing channel-wise non-linearities, and non-mutually exclusive dependencies. Equation \ref{eq:se} offers a mathematical description of the \ac{SE} Block, Where $A_{SE}$

\begin{equation}
\label{eq:se}
    A_{SE}(F) = \sigma(\psi (f_{avg}(F)))
\end{equation}

\noindent means squeeze and excitation attention. $F \in \mathbb{R}^{H \times W \times C}$ is an input feature map, $f_{avg}$ the Average Pooling operator, and $\psi$ a two-layered \ac{MLP} with \ac{ReLU} hidden activation, hidden representation $ h \in \mathbb{R}^{ 1 \times 1 \times \frac{C}{r}}$, with $r$ the reduction ratio, and output $ f \in R^{1 \times 1 \times C}$.

\subsubsection{Convolution Block Attention Module}

\begin{figure}[h]
  \centering
 \includegraphics[scale=0.45]{"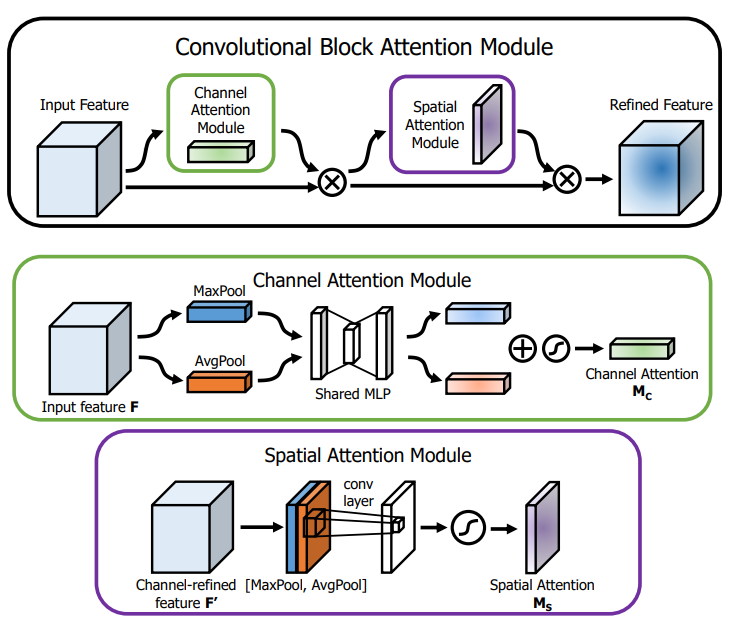"}
  \caption{Overview of \ac{CBAM}. The module considers both channel, and spatial attention to refine feature maps. The spatial attention module applies $ 7 \times 7 $  convolutions over features pooled along the channel dimension, whereas the channel attention module applies a shared \ac{MLP} to average and max-pooled features along the spatial domain. Reproduced from \cite{Woo2018}, Doi: \href{https://doi.org/10.1007/978-3-030-01234-2\_1}{10.1007/978-3-030-01234-2\_1}}
  \label{fig:cbam}
\end{figure}

Building on the principles of spatial and channel attention, \cite{Woo2018} introduced the \ac{CBAM} that simultaneously answers “what” is meaningful and “where” to pay attention. The basic idea is to obtain channel, and spatial-wise attention scores that are subsequently used to refine the original feature map. Fig. \ref{fig:cbam} gives an overview of the proposed \ac{CBAM}. 

The channel attention module starts with squeezing the spatial information by parallel Max and Average Pooling. The authors argue that simultaneously using the two pooling operations leads to a more complete representation when compared with using only one of the two \citep{Woo2018}. Afterwards, each vector is fed to a shared \ac{MLP} with one hidden layer and the outputs are summed element-wise to obtain channel attention scores. The full channel attention, 
$A_c$, operation is formally described by equation \ref{eq:chan_att}:

\begin{equation}
    \label{eq:chan_att}
    A_{c}(F) = \sigma(\psi (f^c_{avg}(F)) + \psi (f^c_{max}(F)))
\end{equation}

\noindent with $\psi$ representing a \ac{MLP}, $f_x$ the pooling functions, and $\sigma$ a sigmoid non-linearity. In turn, the spatial attention module starts by squeezing the channel information by Max and Average Pooling operations along the channel dimension. The pooled vectors are then concatenated, followed by a $7\times7$ convolution and sigmoid non-linearity, resulting in spatial attention scores. Equation \ref{eq:spat_att} is used to formally describe the spatial attention, $A_s$

\begin{equation}
    \label{eq:spat_att}
    A_{s}(F) = \sigma(\phi ([f^s_{avg}(F); f^s_{max}(F)]))
\end{equation}

\noindent with $\phi$ describing the $7\times7$ convolution, $[\;]$ the concatenation, and $f^s_x$ the channel-wide pooling operators. Although the authors suggest the modules could be applied in parallel or sequentially, they experimentally demonstrate that a sequential mechanism, with the channel attention module first, provides better results \citep{Woo2018}. The solution is tested with various standard classification networks and shows consistent improvements in model performance. Interestingly, they also enrich the backbones of various object detectors (e.g., Faster-RCNN \citep{Ren2015}) with \ac{CBAM} and demonstrate it could lead to better precision \citep{Woo2018}.

\subsubsection{Attention Feature Pyramid Network}

\citet{Ren2015} introduce Faster-RCNN, one of the most widely adopted object detection algorithms, aside from the YOLO \citep{Redmon2015} architecture. In turn, \cite{Lin2017} propose the \ac{FPN} that, when integrated with object detectors (e.g., Faster -RCNN), enriches feature representation in a hierarchical, multiscale approach. While it explores higher-level features for detecting large objects and lower-level features for enhanced smaller object detection, it explicitly models multiscale semantics resulting in effective multiscale object detectors. However, it does not work effectively for the problem of small dense object detection, which in many ways resembles the task of nuclei detection (see Figure \ref{fig:context_digipath}). More concretely, \citet{Min2022} have identified a few challenges. Lower-level features present noise that “masks” the representation of small objects \citep{Min2022}. At the same time, although high-level features encode mostly larger objects, they also present some information for small objects, which sometimes leads to ambiguous semantics for small and large object representations \citep{Min2022}. Third, as small objects contain only a few pixels, it becomes challenging to distinguish foreground from background features in complex contexts with objects of similar shape and appearance \citep{Min2022}. They thus suggest the \ac{AFPN} \citep{Min2022} for small object detection while arguing it could alleviate these issues. To enrich small object representation of lower-level features, they propose the  \ac{DTA} module \citep{Min2022}. The solution aims to amplify texture information, which simultaneously emphasises small object details at lower-level layers and highlights features of larger objects at higher layers \citep{Min2022}. In turn, the proposed \ac{FACA} \citep{Min2022} module aims to tackle the challenge of enhancing small object foreground features and suppressing noisy backgrounds, while the \ac{DCA} module aims to alleviate the loss of contextual information due to \ac{RoI} pooling.

The results suggest a significant improvement in object detection, especially for small object instances \cite{Min2022}. The authors considered the Tsinghua-Tencent 100K \citep{Zhu2016}, Pascal VOC \citep{Everingham2010}, and MS COCO \citep{TsungYi2014} datasets, demonstrating that their suggested solution surpasses most state-of-the-art algorithms while working with different model backbones, namely ResNet-50 \citep{He2016}, ResNet-101 \citep{He2016}, and ResNeXt-101 \citep{Xie2017}. 

\subsubsection{Relation Networks}

\begin{figure}[h]
  \centering
 \includegraphics[scale=0.625]{"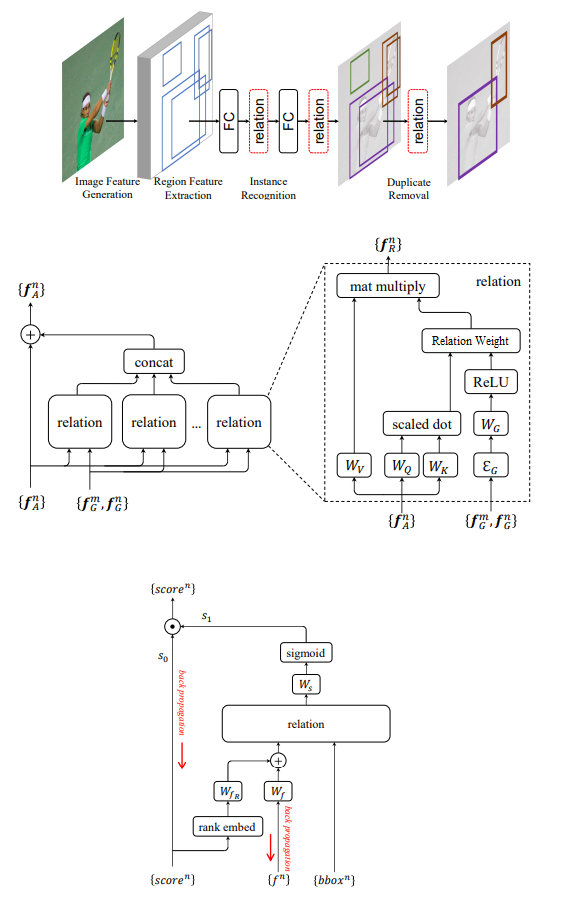"}
  \caption{Top: Overview of an Object Relation Network. Middle: Object Relation Module: Bottom: Duplicate Removal Module. Reproduced from \cite{Hu2018}, Doi: \href{https://doi.org/10.1109/CVPR.2018.00378}{10.1109/CVPR.2018.00378}}
  \label{fig:relation_networks}
\end{figure}

In most cases, object instances in a visual scene are not independent and identically distributed, i.e., their context and interdependencies play an important role in recognition (e.g., a green traffic light often co-occurs with pedestrians at specific image locations). This reasoning is very appealing for cell nuclei detection tasks, as tissue samples observed from \ac{HE}-stained \acp{WSI} are highly structured and semantically rich. More precisely, it is expected for histological concepts (e.g. cells, nuclei, and tissue regions) to be related at multiple levels \citep{Pati2022}.  In turn, typical object detection algorithms treat each instance individually but, inspired by the success transformers demonstrate on \ac{NLP} tasks, \cite{Hu2018} propose an end-to-end object detection strategy that incorporates a self-attention mechanism to capture inter-object relationships. The authors suggest the proposed Object Relation Module \citep{Hu2018} is effective, as it could capture correlations between instances without making excessive assumptions between their locations and features. By aggregating the original representation and relation features (computed from a set of appearance and geometric features), the object relation module refines the original appearance features \citep{Hu2018}. What we consider distinguishes this method is that it explores the relative geometry of objects, an interesting feature for cell nuclei detection in \ac{HE}-stained \acp{WSI}. The Object Relation Module is then stacked on top of multiple object detectors, including a Faster-RCNN \citep{Ren2015} with \ac{FPN} \citep{Lin2017}. Fig. \ref{fig:relation_networks} gives an overview of the proposed Object Relation Networks.

Another contribution of the method regards duplicate removal \citep{Hu2018}, which could benefit from exploring the relations between objects. To achieve the desired result, the authors define duplicate removal as a two-class classification problem. For each ground truth object, only one of the proposals is assumed to be correct. The Duplicate Removal Module  \citep{Hu2018} considers as input a set of bounding box proposal features and classification scores ($s_0$). The features are first fused with the classification scores and then passed through an object relation module before a linear layer for estimating the correctness of the bounding box proposal ($s_1$). The final classification score is established as the product between the bounding box classification score and correctness ($s_0s_1$). Overall, although it is still unclear what is learnt in the Object Relation Module, the authors claim it allows the network to learn object inter-dependencies that are not captured with regular object detectors \citep{Hu2018}. 

\subsubsection{Vision Transformers}

Global context provides important information regarding dependencies of a particular instance with the scene in which it is inserted. Currently, the most powerful tool in \ac{CV} to learn global contextual information is the global self-attention mechanism typical of \acp{ViT}. Unlike \acp{ConvNet}, which use localized receptive fields, \acp{ViT}' global self-attention allows each token to attend to all other image tokens, thus enabling a more comprehensive understanding of the entire image context. While self-attention excels in capturing semantic and spatial relationships among distant image regions, its inherent flexibility also empowers the encoding of local context surrounding an instance. Consequently, \acp{ViT} stand out as an immensely flexible and adaptable family of \acp{ANN}, adept at capturing local, long-range, and global dependencies, thereby acquiring representations rich in contextual information.

Unlike previous works that combine self-attention with convolutions, \ac{ViT} \citep{Dosovitskiy2021} is the first successful attempt at applying transformers directly to images. To deal with the high dimensionality of images, the authors propose to represent each input as a flattened sequence of 2D patches. The patches (or tokens) are then projected (resorting to a trainable linear projection) to a vector of dimension, $D$, referred to as the patch embeddings and combined with position embeddings to retain positional information. These embeddings then serve as input to a transformer encoder layer consisting of alternating layers of multi-head self-attention and \ac{MLP} blocks. The \ac{MLP} contains two layers and GELU \citep{Hendrycks2016} non-linearity. But although \ac{ViT} is a very flexible neural network capable of attending to global information and long-range context, a limitation of this model is the weak image-specific inductive bias. This is very disadvantageous for small datasets where the \acp{ConvNet} strong intrinsic inductive bias (locality, translation equivariance, etc.) becomes very useful. On the contrary, \ac{ViT} demands large amounts of training data. Nonetheless, the authors demonstrate that some transformer layers attend to the entire image, whereas others are more local. Overall, the network attends to image regions that are semantically relevant to the classification task.

Although there are several systematic reviews of vision transformers \citep{Khan2022, LinXiPeng2022, Han2022}, we also review a few publications that build on top of \ac{ViT} and that we consider being relevant for the discussion on context and attention mechanisms in  cell nuclei instance segmentation and classification. We give a high-level overview of the field and present works that could be fine-tuned for medical imaging applications.

\subsubsubsection{Data Efficient Image Transformer}

The fact that \ac{ViT} requires large volumes of annotated data is a major limitation, especially considering real-world applications, where often data acquisition and annotation are expensive (e.g., in digital pathology) or the object instances of interest are naturally underrepresented in the dataset (e.g., defective products in manufacturing). As such, if transformers are to compete with, or even surpass \acp{ConvNet}, proper strategies should be defined to attain data-efficient results. 

One of the first successful attempts to efficiently train \ac{ViT} is the \ac{DeiT} proposed by \cite{Touvron2021}, which provides a detailed set of training mechanisms to increase data efficiency. Given the relative sensitivity of \ac{ViT} to initialization, the authors demonstrate that initialization with a truncated normal distribution leads to better results \cite{Touvron2021}. Furthermore, they compensate for the datasets' limited size (from a transformer perspective) with extensive data augmentations, namely RandAugment \citep{Cubuk2020}. At the same time, they find that AdamW \citep{Loshchilov2017} optimizer, with stochastic depth regularization during training, is also beneficial. Finally, their research suggests batch augmentation is central to improving model performance. Likewise, \cite{Steiner2021} extensively study cost-effectiveness in vision transformers and find that carefully tuned augmentations and regularizations allow, in general, to reduce by 10x the amount of training data while maintaining accuracy. Moreover, the authors present empirical evidence of the common belief that pretraining vision transformers on large-scale datasets followed by fine-tuning to downstream tasks is “simultaneously cheaper and gives better results” \citep{Steiner2021}.

\subsubsubsection{Swin Transformer}

By merging patches in deeper layers, Swin Transformer introduces the hierarchical structure to \acp{ViT} \citep{Liu2021Swin} and closes the gap between \acp{ViT} and \acp{ConvNet}. Indeed, the hierarchical nature of \acp{ConvNet} makes them suitable to represent objects of varying scale and makes these networks capable of intrinsically encoding multiscale contextual information. Moreover, the hierarchical feature maps of Swin Transformer \citep{Liu2021Swin}, in alliance with the self-attention mechanism, make it a strong backbone for state-of-the-art \ac{CV} models (e.g., \acp{FPN} \citep{Lin2017} or U-Net \citep{Ronneberger2015}) to solve dense prediction tasks. Besides, the method introduces a shifted window approach to compute self-attention, which reduces the computational complexity and latency but still enables the encoding of global and long-range context. Notably, the focus on efficiency of the Swin Transformer \citep{Liu2021Swin} potentiates the scaling of \ac{ViT} to high-resolution images \citep{Liu2021Swin}.

\subsubsubsection{Convolution Vision Transformers}

Although frequent claims are made about transformers' superiority, provided larger training datasets are given, there is evidence that \acp{ConvNet} can achieve similar performance under proper design choices and training recipes \citep{Liu2022}. Furthermore, transformers most significant performance gap with \acp{ConvNet} has been on classification, whereas \ac{CV} entails other paradigms (e.g. object detection, instance segmentation), which benefit from \acp{ConvNet} properties (e.g., translation equivariance). Despite this, efforts have been made towards transformers for object detection and segmentation  \citep{Carion2020, FangWenyu2021}, although many still rely on convolutional feature extractors \citep{Carion2020, Cheng2022}. On the other hand, \cite{Raghu2021} compare transformers visual representations with \acp{ConvNet} and demonstrate that while transformers better preserve localization, attend to global information, and possess more uniform features across the neural network layers, local information is still relevant in lower layers and larger datasets become necessary to learn this property despite the weaker inductive bias. In essence, there is no clear argument to choose one family of neural networks over the other, particularly given both paradigms possess unique strengths and can suffer from similar limitations \citep{Pinto2022, Goncalves2022}, i.e., the tendency to learn spurious correlations, simplicity bias, and poor \ac{OOD} generalization. Ultimately, the choice of neural network family is conditioned by the application and available resources. But a promising and recurrent practice is a combination of the two paradigms to exploit their strengths and mitigate individual weaknesses \citep{Wu2021, Guo2022, Xiao2021}, which could allow the introduction of image domain-specific inductive bias into the transformers neural network, or promote self-attention and global context with \acp{ConvNet}. For instance, \cite{Xiao2021} demonstrate replacing the ``patchify" stem with a convolutional one improves training stability while increasing performance. \cite{Chu2021} introduce a convolutional operation in the transformer architecture to remove the positional embedding, whereas \cite{Wu2021} also consider a hierarchy of visual representations. Similarly, \cite{Guo2022b} propose a convolutional stem, depthwise separable convolutions, and an inverted residual feed-forward module with a transformer neural network, which maintains performance and increases efficiency. PatchConvNet \citep{Touvron2021a}, on the contrary, is a convolutional architecture that builds on top of the central idea of ResNets, i.e., the residual connection, but that adopts a modified convolutional stem (like \citet{Xiao2021}), and backbone with constant width and no stride/pooling (the so-called ``column''). Besides, the authors propose to replace the final average pooling layer with a transformer-like cross-attention module. This translates to a model that learns a weight for each patch, depending on its similarity with a trainable class vector (analogous to \ac{ViT} \citep{Dosovitskiy2021} classification token). In essence, after aggregation and reshaping of the attended features, the attention pooling module allows visualizing attention weights (per class) according to the importance of the patch to the classification, making this model more interpretable. The results suggest PatchConvNet achieves superior performance to a ResNet-50 \citep{He2016}, and the authors demonstrate it can easily be extended to instance segmentation and classification, for example, by replacing the Mask-RCNN \citep{He2017}  backbone (without Feature Pyramid Network \citep{Lin2017}) with a PatchConvNet.

\subsubsubsection{Detection Transformer}

\begin{figure*}[t]
  \centering
 \includegraphics[scale=0.7]{"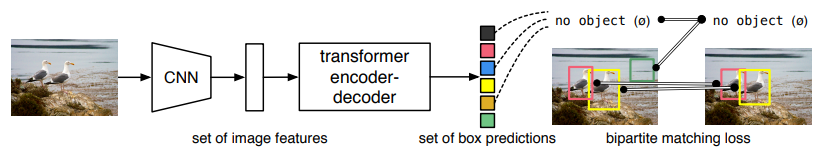"}
  \caption{Overview of the \ac{DETR}. Reproduced from \cite{Carion2020}, Doi: \href{https://doi.org/10.1007/978-3-030-58452-8\_13}{10.1007/978-3-030-58452-8\_13}}
  \label{fig:detr}
\end{figure*}

The \ac{DETR} is introduced by \cite{Carion2020} with the intent of simplifying modern object detectors while maintaining competitiveness. The strategy directly predicts the bounding boxes without needing post-processing and other manually tuned strategies (e.g., \ac{NMS}) through reasoning about the global image context and object relations. Despite consisting of a simple architecture, namely, a \ac{ConvNet} feature extractor and an encoder-decoder module (see Fig. \ref{fig:detr}), the algorithm enables an end-to-end training strategy through a bipartite matching loss and parallel decoding. 

Since the decoder is permutation invariant, to allow bounding box regression and classification, each $N$ object query is embedded with a learnable positional encoding at each attention layer. The decoder thus outputs $N$ normalized object queries that serve as input to a \ac{MLP} to predict bounding box positions, defined by the centre coordinate, height, and width, relative to the input image, as well as classification scores \citep{Carion2020}. Like other modern object detectors, a big challenge while training these networks is correctly corresponding the predicted bounding boxes with the ground truth. Given that the $N$ predictions are larger than the number of ground truth objects, the target vector is usually padded with $\emptyset$ (no object). Adopting a fixed order matching raises a few challenges, as the loss becomes too sensitive to false negatives or positives. To avoid such cumbersome training, the authors adopt an optimum matching strategy, the Hungarian algorithm \citep{Kuhn1955}, that efficiently enforces permutation-invariance, and ensures a single predicted bounding box correspondence for each target. To compute the pair-wise matching costs, the algorithm considers not only the error between the predicted bounding box and ground truth but also the classification scores.

\subsubsubsection{Masked Attention Mask Transformer}

The Masked-attention Mask Transformer (Mask2Former) \citep{Cheng2022} extends the MaskFormer  \citep{Cheng2021} architecture and is an attempt to replace specialized architectures for segmentation tasks by a universal architecture, while being easy and efficient to train. It comprises a backbone feature extractor, a pixel decoder, and a transformer decoder. The feature extraction module generates a low-resolution image feature map $F \in R^{\frac{H}{S} \times \frac{W}{S} \times C_F}$, with $C_F$ the dimension of the feature map, and $S$ the stride (here, $S=32$). In turn, the pixel decoder progressively up samples the feature maps to obtain per-pixel embeddings $\epsilon \in R^{H \times W \times C_\epsilon}$. The transformer decoder module takes as input the backbone image features, $F$, and $N$ learnable positional embeddings (i.e., queries), and outputs $N$ per segment embeddings, $Q$. A Linear Classifier and \ac{MLP} with 2 hidden layers then predict, respectively, $N$ object class probabilities (including no object $\emptyset$), and mask embeddings, $\epsilon_{mask}$, from the transformer decoder outputs. To predict the binary masks, a sigmoid non-linearity is applied after the dot product between the mask embeddings, and pixel-level embeddings. 

\begin{figure}[h]
  \centering
 \includegraphics[scale=0.5]{"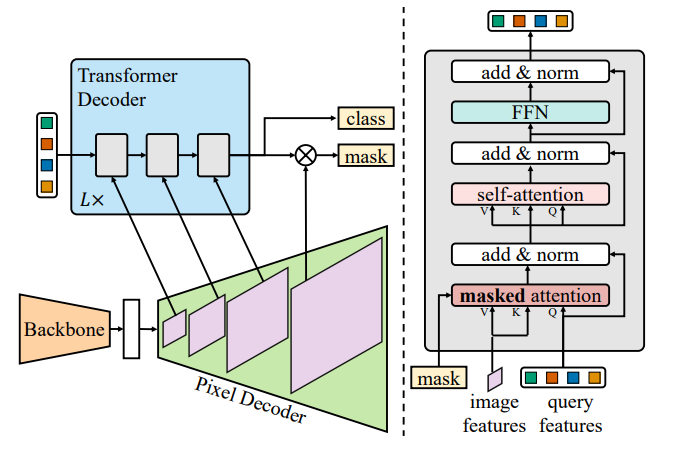"}
  \caption{Overview of Mask2Former. Reproduced from \cite{Cheng2022}, Doi: \href{https://doi.org/10.1109/CVPR52688.2022.00135}{10.1109/CVPR52688.2022.00135}}
  \label{fig:mask2former}
\end{figure}

Like \ac{DETR}, MaskFormer \citep{Cheng2021} uses a bipartite matching loss but instead of using bounding boxes to estimate the costs between ground truth and prediction, it directly uses class and mask predictions. Building on top of MaskFormer \citep{Cheng2021}, Mask2Former similarly leverages the concept of transformers' self- and cross-attention; however, instead of each token attending to all other tokens, it restricts the attention to localized tokens around the predicted segments. Contrary to the standard cross-attention, the proposed masked-attention considers the resized binarized output of the previous $l^{th}-1$ layer. This allows faster convergence and better performance. Besides, to improve small object detection, Mask2Former considers multi-scale high-resolution features. More precisely, it adopts a \ac{FPN} strategy where each feature scale proposed by the pixel-decoder is fed to the corresponding level at the transformer decoder, after positional and scale encoding. The transformer decoder layer thus possesses three levels, one for each feature resolution (1/8, 1/16, and 1/32). This structure is further repeated $L$ times. Fig. \ref{fig:mask2former} illustrates the Mask2Former architecture. 

To optimize the results, the authors propose further improvements such as switching the order of self and cross attention, learnable queries that work like a \ac{RPN}, and remove dropout from the transformer decoder. Finally, to increase training efficiency, the authors suggest the use of $K$ randomly sampled points for computing both matching and final losses, which reduces memory usage by three times. More precisely, for the matching loss they always use the same $K$ uniformly sampled points, whereas for the final loss they use different prediction/ground truth pairs according to importance sampling.

\subsection{Methods for Dense Prediction Tasks in Medical Imaging}
\label{subsec:context_medical_imaging}

Cell nuclei instance segmentation and classification is not an isolated area of knowledge. Indeed, \ac{CV} researchers, in general, have many times been inspired by or benefited from scientific findings in other fields, like \ac{NLP} \citep{Dosovitskiy2021}, geometric \ac{DL} \citep{Bronstein2021}, or neuroscience \citep{Maass1997}. Besides, many solutions in medical imaging are built on top-of high-impact contributions designed to achieve top performance (e.g., accuracy in classification problems, or panoptic quality in instance segmentation) in big and challenging, general \ac{CV}, benchmark datasets. This motivates us to the idea that context- and attention-based methods suggested in medical applications other than  cell nuclei instance segmentation and classification are also worthy of careful review, as some of these contributions could inspire and guide future research in  cell nuclei instance segmentation and classification from \ac{HE}-stained brightfield microscopy images. 

\subsubsection{Dermatology}

\begin{figure}[h]
  \centering
 \includegraphics[scale=0.85]{"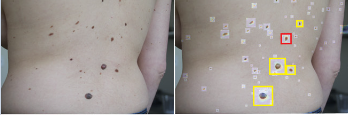"}
  \caption{Left: Example of a wide field image with multiple pigmented lesions on the back of a female subject. Right: 2-class \ac{DL}-based detection of skin lesions. Non-Suspicious Pigmented Lesions of Medium Priority (NSPL-B) are signalled in yellow, whereas Suspicious Pigmented Lesions (SPL) are marked with red. It is also observable that the task of skin lesion detection and classification possesses several similarities with the task of cell nuclei detection and classification, for example, oval foreground objects, dense small objects, RGB images, and high intra-subject variability. Reproduced, with permission, from \cite{Soenksen2021}, Doi: \url{https://doi.org/10.1126/scitranslmed.abb3652}}
  \label{fig:skin_lesions}
\end{figure}

Like cell nuclei, skin lesions possess very heterogeneous sizes, shapes, and appearances, often with irregular fuzzy boundaries, even considering lesions from the same subject. Furthermore, skin lesions' data are usually 2D RGB images with high variability (e.g., due to scanner differences), subject to occlusion (e.g., due to hairs), and sometimes with low contrast between foreground lesions and remaining healthy tissue. Other similarities with cell nuclei detection and classification include skin lesions being, to some extent, rounded, and frequently corresponding to dense, small objects in an image. Fig. \ref{fig:skin_lesions} gives an example of a skin lesion 2D RGB  image that illustrates several of these claims. 

Skin lesion detection, segmentation, and classification are thus difficult tasks. Nonetheless, several works propose the use of \ac{DL} with context- and attention-based mechanisms. \cite{Wang2019} resort to an encoder-decoder like neural network with attention \ac{FPN} to enable a richer multiscale contextual representation useful for downstream skin lesion segmentation.  Additionally, to improve the performance for small object instances, the authors suggest a scale-attention loss function, that weights the loss for objects at different scales, effectively leading to the network paying more attention to small instances. Similarly, \cite{Feng2020} introduce \acp{CPFNet} for medical image segmentation tasks, which combine two pyramidal modules with a U-Net \citep{Ronneberger2015} architecture to aggregate global and multi-scale contextual information. But as attention-based mechanisms lead to improved model performance, some works argue it could also help to achieve explainability in medical imaging tasks \citep{vanderVelden2022}. For instance, in medical imaging segmentation, \cite{Gu2021} discussed \acp{CANet} that adopt attention at multiple levels, namely, channel, spatial, and scale-wide attention, which they argue simultaneously improves accuracy and explainability. \cite{Kaul2021} also build on top of the U-Net \citep{Ronneberger2015} architecture for skin lesion and cell nuclei segmentation. This work, however, proposes \acp{ResGANet} that couples a  \ac{SE}-block \citep{Hu2018} with group convolutions for enhanced feature representations. In turn, to enhance visual representations, \cite{Xu2021} consider a neural network architecture capable of capturing global and multiscale context. The \ac{DCNet} combines hierarchical visual representations from a \ac{ConvNet} encoder with the transformers' self-attention to capture long-range contextual information. More specifically, the output features from each layer of the U-Net \citep{Ronneberger2015} encoder (i.e., feature maps from a ResNet-34 \citep{He2016} feature extractor) are linearly projected, along with flattened input image patches, and positional embeddings. The resulting feature sequence inputs a transformer encoder module to generate enhanced feature representations that encode long-range dependencies. The second mechanism, the \ac{ACF} Module \citep{Kaul2021}, combines multiscale contextual information from each layer of the U-Net \citep{Ronneberger2015} decoder, supported on spatial and channel attention along with dilated convolutions, to decode the feature representations into a 2D segmentation output. But different from previous works that consider only \acp{ConvNet} or hybrid \ac{ConvNet}/transformer networks, \cite{HeLei2022} propose the \ac{FTN} for skin lesion segmentation, which mimics the encoder-decoder topology of widely used \acp{ConvNet} (e.g., U-Net \citep{Ronneberger2015}), while introducing a hierarchy of features into the transformer architecture and skip connections from the encoder to the decoder modules to compensate for loss of spatial resolution. Notably, this design enables local, long-range, global, and multiscale contextual information, thus being a very flexible neural network.

\subsubsection{Radiology}

Although radiology data often corresponds to grayscale images (reflecting physical attributes of the underlying tissues), it is reasonable to state that certain aspects of detecting, and segmenting \acp{RoI} from these data approximate the task of nuclei detection and segmentation. A few similarities support this argument. For instance, both sorts of medical imaging data possess rich contexts, often with small object instances, and high intra-, and inter-subject variability. Moreover, although radiology data can be volumetric, it is often decomposed into 2D slices to ease downstream analysis. Therefore, we hypothesise contextual visual representations and attention-based algorithms previously popularized for radiology could be adapted for cell nuclei detection, and segmentation. For example, \cite{WeiBai2022} successfully apply an attention-based contrastive learning approach to detect pneumonia lesions. \cite{WangQuan2021} propose a neural network with global and local edge attention modules for the segmentation of global and finer details in CT images. In another application, \cite{Deng2021} successfully apply a hybrid transformer to segment the left ventricle in echocardiography. But there are other directions in medical imaging, including radiology, that are worth discussing as viable alternatives to address  cell nuclei instance segmentation and classification. These include multi-task learning, annotation-efficient \ac{DL} algorithms, and other active areas of research like explainability and interpretability. 

\cite{Chen2019MTLSeg} show that context- and attention-based multi-task learning is effective in semi-supervised brain lesion segmentation. Indeed, multi-task learning can work as an inductive bias that helps \ac{ML} models generalize and could work as regularization in semi-supervised settings. \cite{Imran2019} adopt a multi-task \ac{GAN} approach for simultaneous segmentation and classification of chest radiographs. Besides, the authors consider an attention-gated block to tackle possible loss of spatial resolution in masks generated at different levels of a pyramidal encoder, which allows suppression of noisy features, leading to the neural network encoding more effectively the useful multiscale contextual information. \cite{Ullah2023} illustrate that extending a chest radiographs dataset for COVID-19 detection with additional related data in an adversarial multi-task learning paradigm enables data-efficient generalization. Although their methodology does not contemplate explicit context and attention-based mechanisms, a possible improvement could be incorporating attention mechanisms into the neural network layers, or even adapting the adversarial autoencoder to context reconstruction. \cite{Le2021mtbrain}, on the other hand, rely on a contextual multi-task network with atrous convolutions for tumour detection and segmentation, which works as an attention mechanism to enhance contextual information surrounding the tumour region.

Other works demonstrate that attention mechanisms simultaneously increase model performance and explainability. For instance, \cite{Karri2021} propose a multi-module semantic guided attention network (MSGA-Net). The self-attention and multiscale attention modules are introduced to capture global context, while the channel, edge, and location attention modules ease the aggregation of local and global representations. The model is subsequently evaluated on several standard medical imaging datasets, including multi-organ segmentation from \ac{CT} imaging, and brain tumour segmentation from \ac{MRI} data. Whereas the results demonstrate higher segmentation quality, the authors overlay the attention weights each attention module provides onto the original images and demonstrate a more explainable model than other state-of-the-art alternatives.

\subsubsection{Cell/Nuclei Instance Segmentation and Classification in Microscopy Imaging}

\begin{figure*}[t]
  \centering
 \includegraphics[scale=0.7]{"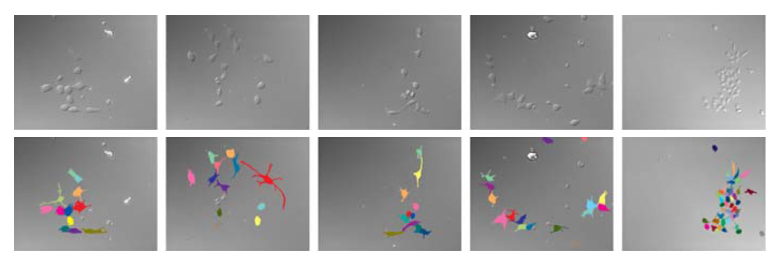"}
  \caption{Top: Sample images from a Central Nervous System (CNS) stem cells dataset \citep{Ravin2008}. Bottom: Ground truth cell instance segmentation maps. This image is illustrative of some similarities between cell instance segmentation and classification in stainless brightfield microscopy imaging and \ac{HE}-stained brightfield microscopy images. We observe morphological variability, air bubble artefacts, dense and overlapped instances, as well as low contrast between foreground instances and background. Reproduced, with permission, from \cite{Yi2018att}, Doi: \href{https://doi.org/10.1016/j.media.2019.05.004}{10.1016/j.media.2019.05.004}}
  \label{fig:cell_instance_microscopy}
\end{figure*}

Microscopy imaging allows the magnification and visualization of organelles, cells, and tissues by exploiting the optical properties of tissues and lens. Although the core of our work regards \ac{HE}-stained brightfield microscopy images, there exist other techniques in the microscopy imaging domain that provide different insights and distinct levels of detail depending on the staining or imaging modality. In this subsection, we delve into some works more tailored to microscopy imaging modalities excluding \ac{HE}-stained brightfield microscopy. We consider this to be relevant as, regardless of the imaging technique, some domain-specific attributes make solutions tailored to other imaging modalities of potential value to \ac{HE}-stained brightfield microscopy imaging. Some of the challenges in microscopy imaging orthogonal to the specificities of each technique are nuclei/cell morphology variability, class imbalance, overlapped, touching, and cluttered instances, noisy/irrelevant background information, and even artefacts, such as out-of-focus images, digitization artefacts, cracks, and air bubbles. Fig.~\ref{fig:cell_instance_microscopy} illustrates these claims. Methodologies suggested for some specific microscopy imaging modality(ies) could thus be worth considering in cell nuclei instance segmentation and classification from \ac{HE}-stained \acp{WSI} (and tiles). A successful illustration of this argument is the StarDist method \citep{Schmidt2018} that is first suggested for cell detection in fluorescence microscopy images, but that later wins the CoNIC Challenge \citep{Weigert2022}, showing its ability to generalize to \ac{HE}-stained brightfield microscopy images as well.

\cite{WanJackson2023} suggest CellT-Net, based on a Swin-transformer \citep{Liu2021Swin} backbone with multiscale contextual features to segment cells of different size (including small instances) in light microscopy imaging. The attention-based model enhances the semantic information of foreground regions and reduces background noise. Besides, the authors consider the adoption of deformable convolutions, which increase model robustness to cell morphology variability. \cite{Sugimoto2022} explore the context surrounding cells via a modified self-attention layer. \cite{Krug2022}, on the other hand, suggest a self-supervised learning paradigm for cell segmentation in fluorescence microscopy images based on a generated attention map and supported by an auxiliary task of predicting the scale of the image (prior knowledge) using as input the image multiplied by the attention map. The method thus makes use of attention mechanisms to learn features of the most relevant regions/concepts in the input image. \cite{WagnerRohr2022} propose CellCentroidFormer, a hybrid \ac{ViT} model for cell detection in microscopy imaging. The strategy suggests an EfficientNet \citep{Tan2019} backbone for feature extraction, and a MobileViT \citep{Mehta2021} with \ac{ConvNet} layers neck for subsequent cell centroid, width, and height regression. The suggested strategy thus combines global transformer-like context and attention-based processing, which promote the model to focus on cell shapes, with \acp{ConvNet} local attention to cell centroids, thus achieving enhanced performance when compared with state-of-the-art \acp{ConvNet}. 
\cite{Chen2022detection} adopt a Mask-RCNN \citep{He2017} for lung cell segmentation, followed by a Swin-transformer \citep{Liu2021Swin} for classification. \cite{Kakumani2022} adopt an attention-based U-Net \citep{Ronneberger2015} with encoder weights initialized from a pretrained autoencoder and attention-gate blocks in the decoder path for semantic segmentation of cells. CellDETR \citep{Prangemeier2020} is a modified version of \ac{DETR} for cell instance segmentation from microscopy images. It consists of a \ac{ConvNet} backbone feature encoder, and two branches: a transformer encoder-decoder for bounding box regression and classification, and a \ac{ConvNet} decoder for segmentation. The input to the segmentation decoder consists of multi-head attention-refined features from the transformer branch, aggregated with the shared \ac{ConvNet} encoder features. The authors also consider skip connections from the encoder to the segmentation decoder module (similar to U-Net \citep{Ronneberger2015}). \cite{Zhou2019irnet}, on the other hand, focus on the precise segmentation of overlapping instances with a method, \ac{IRNet}, supported on inter-cell contextual information ( i.e. long-range context). The strategy suggests extending the Mask-RCNN \citep{He2017} detection and segmentation heads with, respectively, Duplicate Removal \citep{Hu2018} and self-attention-based Instance Relation modules \citep{Hu2018}, which they show to improve the quality of the predicted instances' boundary. \cite{Yi2018att} consider the use of attention mechanisms to address two different challenges in neural cell detection and segmentation. While the authors suggest self-attention in a detection module to refine feature maps, an attention mechanism is proposed in the skip connection of the segmentation encoder-decoder module to suppress noisy background features. 

\subsubsection{Discussion}

The discussed publications present evidence that context- and attention-based mechanisms improve the detection and segmentation performance of neural networks, although the benefits for instance segmentation and classification are not completely clear. A notable methodological aspect of these works is that the challenge of domain generalization is reasonably addressed; namely, nearly all the adopted datasets contain data from multiple centres, spanning several regions of the globe, with samples acquired under different conditions (e.g., differences in data acquisition protocol) and scanners.  This could improve \ac{DL} model generalization while reducing domain shift complications, along with data augmentation and model regularization, which most of the reported publications also contemplate.  Guidelines \citep{England2019, Mongan2020, Lekadira2021} to evaluate the models on standard, public access datasets have also been followed, illustrating the scientific rigour of the discussed medical image segmentation publications. Despite this, a common limitation resides in dataset size and model evaluation. Although we observe the authors are careful not to leak the same patient's data from the training subset to the validation, and test splits, the lack of assessment of generalization to multiple target domains limits the conclusions that can be attained from the observations. Besides, and due to intrinsically balanced datasets, most publications do not address the issue of data imbalance, a common hurdle in medical imaging analysis (e.g., in rare diseases).

Another comment we make on these works is the focus on benchmark performance. Although it still is an important contribution for the \ac{ML} field, \cite{Varoquaux2022} quantitatively demonstrate it could have diminishing returns. As the ultimate objective of \ac{DL} research in medical imaging would be to deploy the algorithms in clinical practice, we suggest researchers should combine these efforts with other clinically meaningful tasks like biomarker research, quality control, data/label efficiency, \ac{OOD} generalization, or explainability. Although there is evidence that context- and attention-based mechanisms could be useful for these topics \citep{Chen2022Mathias, Rao2021, Gu2021Shaoting}, we underline a tendency for publication of positive results in \ac{ML} research, meaning there could be a bias in the scientific community's understanding of the benefit of these algorithms, particularly due to limited chance to learn from negative findings to generate new hypothesis. Besides, other works \citep{Goncalves2022, Laleh2022} present less clear evidence of the advantages of context- and attention-based mechanisms. This question is thus a sensitive topic in \ac{ML} for medical imaging research that should be thoroughly addressed.

Most previously summarised mechanisms could be adapted for  cell nuclei instance segmentation and classification from \ac{HE}-stained brightfield microscopy imaging. In addition, nearly all of the discussed publications carefully address reproducibility, as they publish details on hyperparameters, data splits, or resort to standard publicly available datasets.  But considering the guidelines for \ac{AI} in medical imaging \citep{England2019, Mongan2020, Lekadira2021}, we suggest the results and conclusions should not be blindly interpreted. For example, although statistical measures are reported, they are performed only between different repetitions of the main experiments, meaning the effects of the interaction between the various modules added to the neural networks are not properly assessed, albeit the ablation studies have suggested performance improvements with each individual component. Therefore, despite accumulating evidence that context- and attention-based mechanisms could improve the quality of automated medical image segmentation, it is not possible to unequivocally conclude that these are the only factors responsible for increased capabilities, which hinders the interpretability of these methods. This is a common issue in \ac{ML} research, but we understand it is difficult to assess, as previously discussed \citep{Pineau2020}. 

To finalize, it is worth commenting that each medical imaging modality presents unique characteristics, meaning there is no straightforward success of the same neural network and training recipe in all tasks. For instance, there are differences in contrast and noise between foreground and background objects in skin lesion images or radiology data, and \ac{HE}-stained \acp{WSI}.  The latter have more complex semantic and spatial contexts, with many features of no interest for cell nuclei instance segmentation and classification but of similar morphological or chromatic appearance. This translates to noise that masks the presence of foreground instances, i.e., cell nuclei, in learned visual representations. Besides, there are also domain-specific features and artefacts that differently affect the performance of \ac{DL} models trained on different types of medical imaging data.

\section{Cell Nuclei Instance Segmentation and Classification in Brightfield Microscopy with H\&E Staining}
\label{sec:context_attention_conic}

Due to digitized \ac{HE}-stained brightfield microscopy images (mostly \ac{HE}-stained \acp{WSI}), being part of routine clinical practice, these data are the most widely available. Moreover, \ac{HE} staining is one of the brightfield microscopy imaging techniques that provide the most detailed information, allowing clear visualization of tissue structures like the cytoplasm, nucleus, organelles and extra-cellular components. Therefore, this modality offers the best cost-effectiveness when compared with other techniques, even holding the promise that imaging biomarkers available in these data could replace expensive laboratory tests like genetic or molecular profiling. Methods for \ac{HE}-stained image analysis have thus been the focus of \ac{CPath} research. For these reasons, we give special attention to this imaging technique. Namely, we do a systematic review of context and attention mechanisms for \ac{DL}-based  cell nuclei instance segmentation and classification from H\&E-stained brightfield microscopy images, including \acp{WSI} and tiles. We searched the Scopus database and limited the results to publications between 2019 and 2023. We then read the abstract of every entry returned and excluded publications with no mention of nuclei segmentation/detection, neither attention nor context. To further refine the results, we read the text of the returned publications and excluded papers that did not contemplate algorithms for \ac{HE}-stained \acp{WSI}. Also, we limited our analysis to papers published in Q1 journals (Scimago Journal \& Country  Rank) (e.g., \emph{Medical Image Analysis}, \emph{IEEE Transactions on Medical Imaging}, \emph{Computers in Biology and Medicine}, etc.), and peer-reviewed conferences (ranked C or higher as per the CORE conference ranking) (\emph{Medical Image Computing and Computer Assisted Intervention (MICCAI)}, \emph{International Conference of the IEEE Engineering in Medicine Biology Society (EMBC)}, \emph{International Conference on Medical Imaging with Deep Learning}, etc.). Table \ref{tab:summary_search_nuclei_seg} presents the queries used to search the database, as well as the number of entries returned, inclusion and exclusion criteria, and the final number of publications reviewed.

\begin{table*}[h]
\captionsetup{font=normalsize}
\begin{adjustbox}{width=\textwidth,center}
\begin{tabular}{ccccc}
\hline
 \multicolumn{1}{|c|}{\textbf{Search terms}}                & \multicolumn{1}{c|}{\textbf{Returned}} & \multicolumn{1}{c|}{\textbf{Inclusion criteria}}                                                                                                                                                                                      & \multicolumn{1}{c|}{\textbf{Exclusion criteria}} & \multicolumn{1}{c|}{\textbf{Reviewed}} \\ \hline
 \multicolumn{1}{|c|}{Attention Nuclei Segmentation}        & \multicolumn{1}{c|}{302}                         & \multicolumn{1}{c|}{\begin{tabular}[c]{@{}c@{}}- Reference to attention or context in the abstract/document\\ - Trained and evaluated on \ac{HE}-stained \acp{WSI}\\ - Q1 journal or peer-reviewed conference \end{tabular}} & \multicolumn{1}{c|}{\begin{tabular}[c]{@{}c@{}} - No reference to attention or context in the abstract/document\\ - Imaging modalities other than \ac{HE}-stained \ac{WSI}\\ - Cell segmentation algorithms \\ -Duplicates \end{tabular}}     & \multicolumn{1}{c|}{24}                         \\ \hline
\multicolumn{1}{|c|}{Context Nuclei Segmentation}          & \multicolumn{1}{c|}{179}                          & \multicolumn{1}{c|}{\begin{tabular}[c]{@{}c@{}} - Reference to attention or context in the abstract/document\\ - Trained and evaluated on \ac{HE}-stained \acp{WSI}\\ - Q1 journal or peer-reviewed conference \end{tabular}} & \multicolumn{1}{c|}{\begin{tabular}[c]{@{}c@{}} - No reference to attention or context in the abstract/document\\ - Imaging modalities other than \ac{HE}-stained \ac{WSI}\\ - Cell segmentation algorithms\\ - Duplicates \end{tabular}}                            & \multicolumn{1}{c|}{8}                         \\ \hline
 \multicolumn{1}{|c|}{Transformer Nuclei Segmentation}      & \multicolumn{1}{c|}{58}                          & \multicolumn{1}{c|}{\begin{tabular}[c]{@{}c@{}} -Vision Transformer or Hybrid Transformer/ConvNet \\ - Q1 journal or peer-reviewed conference \end{tabular} }  & \multicolumn{1}{c|}{\begin{tabular}[c]{@{}c@{}} - Imaging modalities other than \ac{HE}-stained \ac{WSI}\\ - Cell segmentation algorithms\\ - Duplicates \end{tabular}}   & \multicolumn{1}{c|}{10}                         \\ \hline
 \multicolumn{1}{|c|}{Transformer Nuclei Detection}       & \multicolumn{1}{c|}{32}                          & \multicolumn{1}{c|}{\begin{tabular}[c]{@{}c@{}} -Vision Transformer or Hybrid Transformer/ConvNet \\ - Q1 journal or peer-reviewed conference \end{tabular}} & \multicolumn{1}{c|}{\begin{tabular}[c]{@{}c@{}} - Imaging modalities other than \ac{HE}-stained \ac{WSI}\\ - Cell segmentation algorithms\\ - Duplicates \end{tabular}} & \multicolumn{1}{c|}{1}                        \\ \hline
 \multicolumn{1}{|c|}{Weakly Supervised Nuclei Detection} & \multicolumn{1}{c|}{38}                          & \multicolumn{1}{c|}{\begin{tabular}[c]{@{}c@{}}- Reference to attention or context in the abstract/document\\ - Trained and evaluated on \ac{HE}-stained \acp{WSI}\\ - Q1 journal or peer-reviewed conference \end{tabular}}   & \multicolumn{1}{c|}{\begin{tabular}[c]{@{}c@{}} - No reference to attention or context in the abstract/document\\ - Imaging modalities other than \ac{HE}-stained \ac{WSI}\\ - Cell segmentation algorithms\\ - Duplicates \end{tabular}}                            & \multicolumn{1}{c|}{1}                         \\ \hline
                         
\end{tabular}
\end{adjustbox}
\captionof{table}{Details of the systematic review on context and attention-based nuclei instance segmentation and classification. The search is performed in the scopus database.\label{tab:summary_search_nuclei_seg}}
\end{table*}

\begin{figure}[!h]
\begin{center}
        \includegraphics[scale=0.65]{"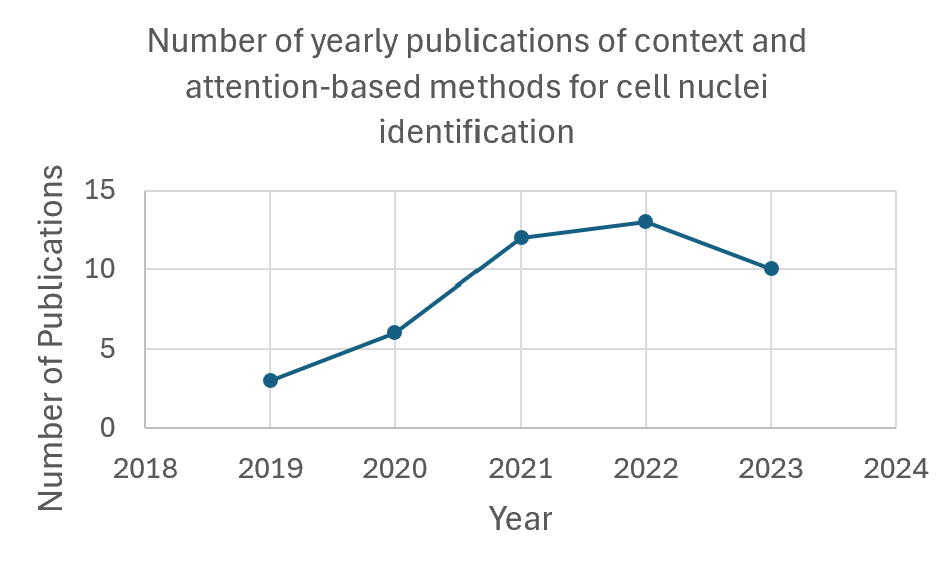"}
\caption{Number of publications reviewed, per year, on context and attention-based mechanisms for  cell nuclei instance segmentation and classification. We observe that these mechanisms are gaining momentum in \ac{CPath}, namely for nuclei instance segmentation and classification from \ac{HE}-stained \acp{WSI}. \label{fig:yearly_pubs}}
\end{center}
\end{figure}

In total, we reviewed $44$ publications on context and attention mechanisms for  cell nuclei instance segmentation and classification from \ac{HE}-stained microscopy images, including \acp{WSI} and tiles. Fig. \ref{fig:yearly_pubs} illustrates the number of reviewed publications, per year, from 2019 to 2022, on context and attention mechanisms for neural-network-based  cell nuclei instance segmentation and classification.

We next provide a comprehensive discussion of how the different context levels have been addressed in the literature for cell nuclei instance segmentation and classification from \ac{HE}-stained brightfield microscopy images, while illustrating the different attention mechanisms that have been explored and their intended use like enriching the estimated feature maps with contextual information or reducing sensitivity to irrelevant background information.

\subsection{Prior Knowledge}

Prior knowledge is very useful in representation learning, as it enables inductive biases useful for constraining the optimization search space. Indeed, having preference for some solutions is necessary for generalization \citep{Goyal2022} and enables learning task-discriminative features with unsupervised representation learning.

\ac{SSL} is a learning paradigm that transforms the unsupervised setting into supervised through the optimization of a loss function on a pretext task using as supervisory signal the data itself \citep{Hastie2009, Goodfellow2016}. As \ac{SSL} enables learning from high volumes of unlabelled data \citep{Campanella2023}, it is desirable in resource-constrained settings, where annotating large and diverse datasets is costly, or in rare target occurrence cases where the labels of some classes are naturally limited. Due to the prohibitive dimensionality of \acp{WSI} and the need for expert clinical annotators, this is especially relevant for \ac{CPath} and could accelerate the clinical implementation of \ac{DL}-based \ac{WSI} analysis tools, including cell nuclei instance segmentation and classification. Another advantage of \ac{SSL} is that it works as regularization, by constraining the search space of downstream fine-tuning optimization around the self-supervised weight initialization region in the parameter search space \citep{Erhan2010}. This regularization effect improves generalization, potentially to \ac{OOD} data as well. In the last years, unsupervised visual representation learning, largely supported by self-supervision with pretext tasks, contrastive learning \citep{Hjelm2018, Misra2020, Chen2020, He2020momentum, ChenHinton2020}, knowledge distillation \citep{Gidaris2021, Grill2021, Chen2021}, or information maximization \citep{Zbontar2021, Bardes2021, Ermolov2021}, has advanced such that it has been shown to compete with supervised learning in several applications \citep{Ciga2020, Bardes2021, Azizi2021, ChenRueckert2019}. Furthermore, self-supervision could translate to models that learn more generic and less spurious correlations, thus beneficial for domain adaptation and generalization \citep{Hendrycks2019}. For example, in \ac{CPath}, multi-task learning with self-supervised pretext tasks in conjunction with supervised classification allows the model to learn more discriminative and domain-invariant features under limited annotated data \citep{Koohbanani2021}. On the other hand, \citep{Srinidhi2021} provide evidence that while self-supervision methods result in more general visual representations, they tend to overfit the pretext tasks if the dataset is small or fail to generalize to \ac{OOD} data. Nonetheless, the authors suggest a \ac{CL} approach in downstream fine-tuning that they demonstrate to improve accuracy and generalization of patch-, and slide-level classification on benchmark histology datasets when compared with standard downstream transfer learning. In the scope of  cell nuclei instance segmentation and classification, contextual information, namely prior knowledge, has been used to learn more useful representations with self-supervision. For instance,  motivated by the annotation efficiency of \ac{SSL}, \cite{Sahasrabudhe2020} introduce an attention-based unsupervised nuclei segmentation algorithm. The authors propose a multilayer convolutional neural network, based on dilated convolutions, that increase the receptive field thus providing a larger context for the pixel classification task. This model outputs a coarse nuclei segmentation binary attention map.  The attention map is then aggregated, by multiplication, with the original image and the resulting nuclei-salient \ac{HE}-stained \ac{WSI} tile is used as input to a scale classification network (ResNet-34) \citep{He2016}. The networks are trained end-to-end where the total loss is a combination of the loss computed at the scale classification level, with equivariance and smoothness regularization losses at the attention mask prediction module.  The authors argue that given a \ac{HE}-stained \ac{WSI} tile, the level of magnification can be inferred by looking at the size and texture of the nuclei in the patch \citep{Sahasrabudhe2020}. They thus use an attention mechanism to exploit this domain-specific context, i.e. in the form of prior knowledge, to design a network capable of learning annotation-efficient and generalizable representations. The final segmentation map is obtained by post-processing the coarse segmentation image. The results demonstrate that, despite underperforming state-of-the-art supervised methods, the proposed model is competitive with other unsupervised learning methods \citep{Sahasrabudhe2020}. Another work exploring contextual information in the form of prior knowledge is that of \cite{Boserup2023}, which consider a contrastive loss at the pixel level for annotation-efficient nuclei segmentation. In this work, it is assumed that tiles containing the same nuclei types are likely to be similar. Therefore, positive pairs are defined as semantically similar patches that are likely to contain the same object(s), while negative pairs correspond to patches that are not likely to contain the same object(s), which could be interpreted as approximating a patch to that of its local context (i.e., its neighbours) when embedded in a low-dimensional manifold. To estimate patch similarity to compute the contrastive loss, the authors consider the mean squared error or mean cross entropy and devise a confidence prediction network which outputs, for each class, the likelihood of each pixel belonging to it. The input images are then scaled by the confidence maps (for each class), and the loss computed at the level of these confidence-weighted vectors. To complete the nuclei segmentation task, at inference time, the predicted confidence maps are thresholded at 0.5.

\citet{Hassan2022}, on the other hand, use the knowledge that cell nuclei are frequently organized into cellular "communities". Starting from a pretrained nuclei instance segmentation and classification method (Mask-RCNN \citep{He2017}), the authors resort to a messsage passing network to learn inter-instance relations and predict cellular communities, which they then use to refine the nuclei class predictions.

\subsection{Local Context}

Channel and spatial attention mechanisms are generally used to enhance foreground information while suppressing certain background features. Moreover, these are also used to refine coarse nuclei segmentation maps. Indeed, attention mechanisms are often used to suppress irrelevant contextual information allowing \ac{DL} models to focus on useful features like the local context of a pixel in a nucleus. Overall, local contextual information helps refine nuclei borders and allows distinguishing between nuclei types.

\cite{Kong2020} suggest two-stage stacked U-Nets, with attention, for nuclei segmentation in \ac{HE}-stained tiles. A stacked U-Net consists of four parallel U-Net backbones in which the hidden representations are merged with an attention module. This generates attention weights for each of the four U-Net outputs that are then aggregated for the final segmentation. However, in a two-stage scheme, the authors propose that the output of the first stacked U-Net be fused with the original input image for the second round of segmentation. This allows the suppression of background noise while making nuclei more salient, which facilitates the subsequent segmentation of overlapping nuclei. By leveraging attention in this way, the strategy effectively suppresses noisy or irrelevant contextual information (i.e., background information, like extracellular matrix or irrelevant cell organelles) while enhancing the relevant local context of the foreground pixels (i.e., nuclei). Similarly, \cite{Yoo2019} propose a \ac{WSL} strategy based on attention to enhance local context. Their method considers an auxiliary task to learn from point annotations. The main segmentation network tends to produce coarse segmentation maps; therefore, an auxiliary network is used to guide the main network to produce more delicate edges. The adjacent network is trained with pseudo-labels generated from a Sobel filter on the predicted segmentation masks from the main network, and an attention module is used to refine the edge predictions of the auxiliary model. 

CDNet \citep{He2021} is a multitask network for nuclei instance segmentation. The architecture contemplates a feature extraction module (encoder-decoder architecture) and three branches with residual units to predict the nuclei mask, centripetal direction feature map, and nuclei centers feature map. Each branch has as input, respectively, feature maps from the feature extraction module, feature maps from the mask branch, and feature maps from the centripetal direction prediction branch.  The spatial attention mechanism is then used to aggregate features from the different branches on the premise that the auxiliary tasks provide relevant contextual information to guide the instance segmentation task. Besides, at test time, a direction difference map computed from the centripetal direction feature map and nuclei centers is used to refine nuclei borders. The approach is suggested to effectively deal with ambiguities due to touching or overlapping nuclei \citep{He2021}.

The Mulvernet model \citep{Vo2023} is an extension of HoVer-Net \citep{Graham2019b}, where the skip connections are enriched with an attention gate and multiple filter units to enrich the representations with multiscale features. DRCA-Net, designed by \cite{Dogar2023}, is composed of modified ResNet-50 \citep{He2016} encoder and three decoder branches, with \ac{CBAM}-based \citep{Woo2018} dense attention blocks, for instance, segmentation and classification. Each branch is responsible for a single task: nuclei segmentation, regression of horizontal and vertical distance maps, and nuclei classification. A post-processing methodology, similar to HoVer-Net \citep{Graham2019b}, is then adopted to combine the three outputs into a single prediction. The authors demonstrate the proposed model results in finer segmentation maps, facilitates the separation of overlapping nuclei, and reduces false positives when compared with other state-of-the-art methods on three standard open-access datasets. ConDANet \citep{Imtiaz2023} is an encoder-decoder segmentation model where it is suggested to replace average or max pooling and unpooling with, respectively, the discrete and inverse discrete wavelet transforms, an operation that better preserves image details. Moreover, the authors propose the use of a contourlet filter \citep{Do2005} with a spatial attention mechanism to provide additional edge information, in essence, a strategy to enhance additional contextual information around the nuclei borders. Ultimately, the use of the attention mechanism allows for refining local contextual information, which translates to better nuclei instance segmentation and classification.

CPPNet \citep{Chen2023} is suggested to improve the segmentation of touching and overlapping nuclei. The method extends the StarDist \citep{Schmidt2018, Weigert2022} model by estimating the distance (along a set of predefined radial directions) from an additional set of points between the centroid and the initially predicted boundary. Furthermore, the approach contemplates a \ac{CEM} and a \ac{CWM} to refine the initial centroid to boundary distance relative to these points. Besides, a shape aware loss is suggested to penalize differences in shape features between the predicted and ground truth nucleus representations. Finally, the authors introduce a post-processing scheme to remove redundant predictions and obtain the the final segmentation map. Overall, by leveraging an additional set of points this method can capture more contextual information than StarDist, specially near the boundary.

\subsection{Long-range Context}

We have that \ac{CPath} data represent a rich tissue microenvironment that reflects multiple clinical variables like the diagnosis, the disease subtype, tumour stage and grade, lifestyle habits of the subject, etc. The imaging features encoding these variables are spread across the tissue, even in distant regions, but their relationships and co-occurence provide relevant anatomical and pathological information. Therefore, a representation rich in long-range context, encoding the co-occurrence and long-range dependencies of different concepts (e.g., cells, or nuclei) is more discriminative than a representation that only considers local context. Besides, such information could be especially useful in the presence of ambiguities (e.g., due to noisy or blurred data, occluded objects, small instances, etc.).

\cite{Ali2022} consider a U-Net-like encoder-decoder model with multi-scale attention through dense dilated convolutions at the level of the last hidden layer of the encoder, which given the large receptive field of the hidden units in this layer leads to learning a representation rich in long-range contextual information. Besides, they propose a multi-scale efficient channel attention module at each decoder layer, which in combination with the hierarchical structure of the decoder, increases the use of semantic and spatial context. A boundary refinement module is also proposed to refine nuclei borders. 

Attention mechanisms and long-range context have also been exploited in \ac{WSL}, a set of data-driven learning techniques that optimize predictive models with incomplete, inexact, or inaccurate supervision \citep{Zhou2018WSL}. Indeed, \ac{WSL} is a widely explored paradigm to tackle the challenges of learning from partially annotated data, label inconsistencies, noisy samples, and domain shifts. In object detection and segmentation, the goal of \ac{WSL} tasks is often to propagate image (bag-level) labels to individual object instances \citep{Zhang2021Hsuan}, or to generate segmentation masks from point annotations. \ac{WSL} is thus a cost-effective solution for \ac{CPath} as it could alleviate the burden of annotating the gigapixel \acp{WSI} at the pixel, or instance level. Notably, \ac{WSL} has demonstrated useful in several \ac{CPath} applications, like tissue grading~\citep{Campanella2019}, mitosis detection in breast tissue tiles~\citep{Li2019b}, or prediction of molecular alterations from \ac{HE}-stained \acp{WSI}~\citep{Kather2020}.  \citet{Thiagarajan2019}, introduce a \ac{KD} and \ac{MIL} framework for weakly supervised epithelial cell nuclei classification. To perform the patch-level classification task, namely to detect epithelial nuclei, each \ac{WSI} is subdivided into a set of patches and the attention-based \ac{MIL} method is adopted to optimize a bag-level classifier to predict cancerous tissue. First, patch-level feature extraction takes place followed by attention-based feature aggregation to learn a global representation for the \ac{WSI}. This attention-based \ac{MIL} harnesses long-range inter-patch context to develop a bag classifier. Notably, given the method's flexibility in accommodating a variable number of whole-slide patches, the optimized model can be employed for predicting instance-level labels by considering each patch as a collection of nuclei instances. This is motivated by the fact that chromatic and morphological alterations in epithelial cell nuclei are correlated with cancer mutations. Hence, to fine-tune the instance-level features, a student model, employing an identical architecture to that of the teacher, undergoes optimization with supervision from the teacher model. Besides, the authors adopt virtual adversarial training as regularization, which improves robustness to bag perturbations (e.g., noisy instances), thus reducing overfitting. Overall, this is a label-efficient strategy leveraging attention and long-range contextual information in \acp{WSI} to learn a label-effective tile-level epithelial cell nuclei classification network.

LIRNet~\citep{ZhouFei2020} is suggested as a framework for cell nuclei counting and localization. The architecture consisted of a lighter version of U-Net, with dot product attention in the bottom layer to capture inter-object relationships and long-range context. The network estimates pseudo-density maps whose ground truth is generated from point annotations, i.e., bounding box centroids. An advantage of the proposed algorithm is that it allows both \ac{SL} and \ac{WSL}. While in \ac{SL} the full pseudo-density map is used as ground truth, in \ac{WSL}, the image is divided into patches and each patch is assigned a label representing one of the following: no nuclei, one nucleus, and at least two nuclei. Afterwards, each learning paradigm adopts a specialized loss function, supported by the integral of local densities (i.e., foreground objects), under the assumption that if the integral of the whole density map is nuclei counts, then the integral of individual foreground objects should equal one. Nuclei locations are derived by post-processing the estimated pseudo-density maps.

\subsection{Global Context}

Global context refers to the relationship between each instance and the scene as a whole. In nuclei instance segmentation and classification this is achieved mostly through \acp{ViT} that allow modelling the relationship between a small tissue patch and all other patches.

Medical imaging instance segmentation and classification is an emerging application of \acp{ViT}, but the weak inductive bias and the lack of large annotated datasets slows progress and motivates the hybridization of transformers with convolutions, a more data-efficient alternative. Hybrid neural networks can potentially increase performance compared with a pure \ac{ViT} or \ac{ConvNet}, making them an attractive approach. \cite{LuoSelvan2022}, develop a U-Net-based neural network with parallel self-attention blocks connected by cross-attention blocks to the encoder module, which allows the exchange of complementary global and local information between the two encoding paths. \cite{An2022} define an efficient attention module, combined with dual-channel shifted \ac{MLP}, and transformers self-attention to enrich foreground representations and suppress background noise. \cite{Obeid2022} propose the use of DETR \citep{Carion2020} for nuclei detection but introduce a realistic data augmentation policy to increase the number of negative tiles (i.e., no nuclei) and a near-duplicate merging strategy to decrease the false positive rate. \cite{ChenZhao2022} enrich the HoVer-Net \citep{Graham2019b} trunk with a transformer attention module head and extend both the residual blocks and the decoding branches with SimAM \citep{Yang2021}. \cite{Valanarasu2021} define a two-branch encoder-decoder model with transformer-based gated axial attention. The global branch is shallower and reasons over the entire image to compute global attention, while the local branch processes image patches to capture finer details. At the final stage, the output of both branches is combined for the segmentation prediction. The gated axial transformer layers are a modified version of axial attention \citep{Ho2019} and are suggested to overcome the difficulty in learning positional encodings under limited data. Overall, by combining global and local self-attention mechanisms, the solution contemplated in this work increases data efficiency and performance in several medical imaging segmentation tasks, including cell nuclei segmentation.

\citet{Zhang2023} design CircleFormer, an extension of DETR that works with circular bounding boxes( $bbox = (c, r)$, with $c=(x, y)$ the coordinates of the bounding box center and $r$ the radius). Therefore, due to cell nuclei possessing a rounded shape, the strategy is more fitted for nuclei detection and segmentation tasks. Moreover, the work suggests circle modulated attention that takes into account the object radius, enabling the model to learn features of objects of various scales. This mechanism tailors the model to also work with smaller instances, a known shortcoming of state-of-the-art object detectors like DETR and Mask-RCNN \citep{Dubey2022, Min2022}.

Self-supervised learning has also been shown successful for pretraining vision transformers for nuclei segmentation. \cite{Haq2022} propose a triplet loss computed at the embedding space of a TransUNet \citep{chen2021transunet} to promote similarity between the representations of neighbouring (8-connected) foreground patches (i.e, with cell nuclei) while contrasting the representations of these patches with background (i.e., no nuclei). At the same time, the authors adopt the task of predicting the input image (i.e., H\&E-stained tile) scale (prior knowledge), with a dedicated model that receives as input the \ac{HE}-stained tile and the segmentation mask predicted by TransUNet \citep{chen2021transunet}. In a subsequent stage, the model is fine-tuned in a small annotated dataset. The authors also demonstrate the widely accepted idea that pretraining benefits downstream supervised fine-tuning and that tailored self-supervised tasks are more beneficial for the nuclei segmentation task than pretraining on regular RGB images (e.g., ImageNet) \citep{Deng2009}.

In turn, \citet{Pan2023} propose SMILE, a multitask architecture to simultaneously achieve nuclei instance and semantic segmentation, as well as nuclei classification. The design they introduce contemplates a tri-encoder and tri-decoder U-Net but with a shared \ac{RCCA} module \citep{Huang2019CCNet} at the bottom layer of the encoders. The criss-cross attention module resorts to the affinity mechanism that is similar to that of global self-attention, but where each pixel attends only to pixels in the same row and column, instead of to all other pixels. The \ac{RCCA} module consists of two stacked criss-cross attention modules. A second pass through this module produces a similar effect to that of global self-attention, thus being a computationally efficient way of capturing global context and long-range dependencies.

\subsection{Multiscale Context}

In \ac{CPath}, common forms of multiscale context include the relationship between nuclei and cells, and the way nuclei are arranged and grouped together to form cellular communities and tissue structures. Therefore, understanding this broader context may be useful for nuclei instance segmentation and classification tasks.

U-Net-like encoder-decoder neural networks enriched with attention mechanisms are widely adopted for nuclei instance segmentation and classification. Notably, through the use of attention mechanisms or dilated convolutions in conjunction with the hierarchical structure of encoder-decoder models, these strategies enable the representation of multiscale contextual information \citep{Li2023}. \cite{Vahadane2021} propose the use of a second encoder in a U-Net-like architecture that captures input-specific attention. The multi-resolution attention maps of the attention encoder are then aggregated with the U-Net \citep{Ronneberger2015} feature maps at an attention skip module. The dual-encoder U-Net then predicts nuclei and boundary probability maps that are post-processed to obtain the final segmentation output.  \cite{HeYongjie2021}, on the other hand, combine a nested U-Net, inspired by U-Net++ \citep{Zhou2018}, with a hybrid attention module (similar to \ac{CBAM} \citep{Woo2018}), and argue the resultant model can capture more multiscale context, which facilitates the segmentation of small and dense nuclei while being flexible to nuclei appearance variability. \cite{Wazir2022} combine a variant of the \ac{SE}-block \citep{Hu2018} with multiple losses in an encoder-decoder model to segment multiple structures, namely cells and nuclei. By adopting dilated (atrous) convolutions of multiple scales, the suggested design can cover multiple levels of context, including local, long-range, and multiscale context. Another strategy based on dilated convolutions \citep{LiYang2021} proposes to enrich a U-Net model with channel attention blocks, dilated convolutions, and a boundary smoothness loss, which promotes the separability of foreground boundaries and background representations, thus improving segmentation. FeedNet \citep{Deshmukh2022} suggests a mechanism (FE Block) to compensate for the loss of spatial information after the pooling layers in encoder-decoder neural networks. At the same time, the authors propose depth-wise separable convolutions and design an \ac{LSTM} \citep{Hochreiter1997} block to refine the encoder output features, which works like a channel attention mechanism. Interestingly, with fewer model parameters, the results demonstrate competitive performance on nuclei instance segmentation. ASPPU-Net \citep{WanQin2020} consists of a modified U-Net \citep{Ronneberger2015} with atrous spatial pyramid pooling to induce multiscale representations and capture nuclei contextual information. This model aims to improve the segmentation quality with overlapped instances. Therefore, the authors combine the benefits of contextual visual representations with a concave point detection algorithm to refine the segmentation of highly cluttered nuclei. Besides, the authors exploit the properties of the \ac{HE} staining; namely, they project the RGB image into the individual Haematoxylin and Eosin components on the argument that as Haematoxylin is more sensitive to nucleic acids (RNA/DNA), richly present in cell nuclei, rough endoplasmatic reticulum, and ribosomes, it provides a better channel to contrast nuclei borders with surrounding tissue, thus easing nuclei segmentation. \cite{Ahmad2023} introduce DAN-NucNet, an encoder-decoder segmentation model with spatial and channel attention gates. In DSCA-Net \citep{Shan2023}, the authors propose extending a U-Net \citep{Ronneberger2015} architecture with a channel and spatial attention module at each layer of the decoder, which, in conjunction with the hierarchical nature of the decoder, translates to a representation encoding context at multiple scales. Moreover, the method entails an edge attention module that aggregates multiscale features to refine nuclei borders.

Although \ac{FPN}-based dense prediction methods have shown successful, mainly driven by their inherent ability to aggregate multi-scale contextual information, they possess some limitations like difficulty in detecting small objects. \citet{Geng2021a} discuss other limitations including the inability to transfer some relevant semantic information from higher to lower levels and discuss the advantages of expanding \acp{FPN} \citep{Lin2017} with attention mechanisms to capture better long-range dependencies and enhance important spatial positions. Moreover, the authors identify that feature maps at the higher level of \acp{FPN} \citep{Lin2017} lack contextual information as they correspond to a single scale. Therefore, to enhance the expressiveness of contextual features at this level, the authors propose a context augmentation module based on dilated convolutions and the self-attention mechanism.

The architecture of the Swin transformer \citep{Liu2021Swin} enables the encoding of global, long-range, and multiscale contextual information. Therefore, it is a strong backbone for \ac{HE}-stained brightfield microscopy cell instance segmentation and classification models \citep{Zhou2023}. Nevertheless, the weak inductive bias that characterizes \acp{ViT} limits the capability of these models to encode spatial context \citep{Wang2023}. To address this challenge, \citet{Wang2023} consider an architecture similar to that of Mask-RCNN \citep{He2017}, but with Swin transformer \citep{Liu2021Swin} backbone and suggest dilated convolutions of multiple dilation rates at each stage of the transformer backbone, which increases the range of multiscale contextual information that is encoded. Besides, the method introduces a convolutional graph neural network-based feature fusion module that combines features from the top and bottom-layers of the Swin transformer \citep{Liu2021Swin}  backbone to obtain an affinity matrix, that captures the extent of the dependency between two nodes. Then, this matrix is used by graph convolutions to map the features from the Swin transformer bottom layer to a representation that encodes inter-instance relations. But another method focused on improving model performance in clustered nuclei instance segmentation is proposed by \citet{Ke2023}. Based on hybrid \acp{ViT}, the architecture extends the U-Net \citep{Ronneberger2015} encoder with transformer layers and modifies the decoder path to, besides nuclei masks, output individual nuclei edges, as well as the edges of clustered nuclei. Besides, the proposed model can be optimized only from partial edge annotations and a small portion of fully annotated masks, thus being annotation-effective. A similar approach \citep{He2023} based on a tri-decoder architecture replaces the hybrid convolutional encoder and decoders with Swin transformer \citep{Liu2021Swin} layers. Besides, to increase efficiency and capture correlations between the nuclei segmentation and edge prediction tasks, the design proposes a shared multi-head self-attention mechanism in the decoders. Overall, the broader contextual understanding of these methods helps the models to better differentiate the boundaries of clustered nuclei, that frequently contain low contrast and touching or overlapping instances.

\cite{Guo2021} posit a two-stage method for cell nuclei instance segmentation, bolstered with attention, that requires only nuclei centroid annotations. From these weak annotations, the authors then estimate two types of pseudo ground truth masks: point annotations + Voronoi labels, and superpixel pseudo ground truths generated by masking superpixels from a SLIC~\citep{Achanta2010} algorithm with Voronoi cells. These two types of ground truth labels are thus complementary. The main segmentation network is an encoder-decoder architecture, with slight modifications in the encoder, like dilated convolutions to incorporate multi-scale context, but with the insertion of a \ac{CBAM}-based feature aggregation module in the decoder. During stage I, the superpixel masks are used as the ground truth for updating the network weights. The authors propose an auxiliary network, the mask-guided attention network, to help refine the segmentation module predictions. The output of this network is then considered an attention weight to help denoising the nuclei segmentation predictions. A specific loss to the auxiliary task is proposed, which penalizes pixel-wise errors between the point annotations + Voronoi boundaries and the attention map. Stage II could then be considered a refinement step, where confident learning \citep{Northcutt2021} is used to update the pseudo ground truth labels before retraining the whole framework with the revised labels~\citep{Guo2021}.

\subsection{Discussion}


\begin{table*}[h]
\captionsetup{font=normalsize}
\begin{adjustbox}{width=\textwidth, center}
\begin{tabular}{cccccccccc}
Method & Year & Base Architecture & Attention & Context Type(s) & Context Level(s) & Learning & Dataset(s) & Reference \\ \hline
Self-sup. Scale Pred. & 2020 & ConvNet & spatial & spatial & local & \ac{SSL} & CoNSeP, KUMAR, TNBC &~\cite{Sahasrabudhe2020} \\
Patch Contrast. Learn & 2023 & ConvNet &  spatial & spatial & local & \ac{SSL} & CoNSeP, KUMAR &~\cite{Boserup2023} \\
MSAL-Net & 2022 & U-Net & channel & semantic & long-range & \ac{SL} &  KUMAR  &~\cite{Ali2022} \\
DEAU & 2021 & U-Net & channel, attention-prior, multiscale & semantic & long-range, multiscale & \ac{SL} & KUMAR, ConSeP, CPM-17 &~\cite{Vahadane2021}\\
SUNet & 2020 & stacked U-Net & spatial & spatial & local & \ac{SL} & KUMAR, TNBC &~\cite{Kong2020} \\
HistoSeg & 2022 & U-Net & channel, spatial & semantic, spatial & local, long-range & \ac{SL} & GlaS, KUMAR &~\cite{Wazir2022} \\
CAB-Net & 2021 & U-Net & channel & semantic & long-range & \ac{SL} & KUMAR & \cite{LiYang2021} \\
HAN-Net & 2021 & U-Net & channel, spatial & semantic, spatial & local, long-range & \ac{SL} & KUMAR & \cite{HeYongjie2021} \\
DRCA-Net & 2023 & U-Net & channel, spatial & semantic, spatial & local, long-range & \ac{SL} & ConSeP, KUMAR, CPM-17, PanNuke & \cite{Dogar2023} \\
HyLT & 2022 & Hybrid Transformer & self-/global & semantic, spatial &  global, long-range, multiscale & \ac{SL} &  GlaS, KUMAR & \cite{LuoSelvan2022} \\
HEA-Net & 2022 & Hybrid Transformer & channel, spatial, self-/global & semantic, spatial & global, long-range, multiscale &  \ac{SL} & GlaS, KUMAR & \cite{An2022} \\
TransNuSS & 2022 & Hybrid Transformer & self-/global & semantic & global, long-range & \ac{SSL} & TNBC, KUMAR & \cite{Haq2022} \\
NucDETR & 2022 & Hybrid Transformer & self-/global & semantic, spatial & global, long-range, multiscale & \ac{SL} &  ConSeP, PanNuke  & \cite{Obeid2022} \\
THSVNet & 2022 & Hybrid Transformer & self-/global, channel, spatial & semantic, spatial & global, long-range & \ac{SL} & ConSeP, PanNuke & \cite{ChenZhao2022} \\
MedT & 2021 & Hybrid Transformer &  self-/global  & semantic, spatial &  global, long-range, multiscale & \ac{SL} & GlaS, KUMAR  & \cite{Valanarasu2021}  \\
Mask Guided Attention & 2021 & Encoder-Decoder (ConvNet) & spatial, channel & semantic, spatial &local, long-range & \ac{WSL} &  KUMAR, TNBC & \cite{Guo2021} \\
Pseudo-edge Net & 2019 & Encoder-Decoder (ConvNet) & spatial & spatial & local & \ac{WSL} & KUMAR, TNBC & \cite{Yoo2019} \\
Distill-to-Label & 2019 & ConvNet & self-attention (gated) & semantic & local, long-range & \ac{WSL} &  CCa, BCa & \cite{Thiagarajan2019} \\
LIRNet & 2020 & U-Net & self-attention & semantic, spatial & local, long-range & \ac{WSL} & CA & \cite{ZhouFei2020} \\
DAN-NucNet & 2023 & U-Net & channel, spatial & semantic, spatial & local, long-range & \ac{SL}  & PanNuke, KUMAR, CPM-17 & \cite{Ahmad2023} \\ 
Mulvernet & 2023 & HoVer-Net & spatial & spatial & local & \ac{SL} & MoNuSAC, GlySAC, ConSeP &  \cite{Vo2023} \\
DSCA-Net & 2023 & U-Net & channel, spatial & semantic, spatial & local, multiscale, long-range & \ac{SL} & TNBC &  \cite{Shan2023} \\
REU-Net & 2022 & U-Net & branch, spatial & semantic, spatial & local, long-range & \ac{SL} & KUMAR, ConSeP, HUSTS, CPM-17 &  \cite{Qin2022} \\
AL-Net & 2022 & U-Net & spatial, channel & semantic, spatial & local, long-range & \ac{SL} & KUMAR, PanNuke, ConSeP &  \cite{ZhaoZuo2022} \\
RCSAU-Net & 2022 & U-Net & channel, spatial & semantic, spatial & local, long-range, global & \ac{SL} & KUMAR, PanNuke & \cite{WangLuo2022} \\
MCNS-UNet & 2022 & U-Net & channel, spatial & semantic, spatial & local, long-range & \ac{SL} & MoNuSAC &  \cite{Sheikh2022} \\
HBANet & 2021 & ConvNet & channel, spatial & semantic, spatial & local, long-range & \ac{SL} & KUMAR, CPM-17 &  \cite{ChengChen2021} \\
MASG-GAN & 2021 & Encoder-Decoder(ConvNet) & channel, spatial, branch & semantic, spatial & local, long-range & \ac{SL} & KUMAR, TNBC, ConSeP, KIRK, CPM-17 &  \cite{ZhangYuWang2021} \\
Feat. Att. Net & 2021 & Encoder-Decoder(ConvNet) & channel, spatial & semantic, spatial & local, long-range & \ac{SL} & KUMAR, ConSeP &  \cite{MurtazaDogar2021} \\
NucleiSegNet & 2021 & U-Net & channel & semantic & local, long-range  & \ac{SL} & KMC, KUMAR &  \cite{Lal2021} \\
ACDNN & 2020 & ConvNet & channel & semantic, spatial & local, long-range & \ac{SL} & - &  \cite{Yi2020} \\
NAS-SCAM & 2020 & ConvNet & channel, spatial & semantic, spatial & local, long-range  & \ac{SL} & MoNuSAC &  \cite{Liu2020} \\
RIC-UNet & 2019 & Encoder-Decoder (ConvNet) & channel & semantic & local, long-range & \ac{SL} & TCGA, CPM-17 & \cite{ZengLu2019} \\
MDC-Net & 2021 & Encoder-Decoder (ConvNet) & - & spatial, semantic & long-range, multiscale & \ac{SL} & TNBC, TCGA & \cite{LiuTang2021} \\
AA-Net & 2021 & Encoder-Decoder (ConvNet) & channel & semantic & long-range, multi-scale & \ac{SL} & KUMAR & \cite{Geng2021a} \\
ASPPU-Net & 2020 & U-Net & - & semantic, spatial & long-range, multi-scale & \ac{SL} & TNBC, TCGA & \cite{WanQin2020} \\
Color-space HoVer-Net & 2022 & ConvNeXt/HoVer-Net & channel, spatial & semantic, spatial & local, long-range, multi-scale &\ac{SL} & Lizard &\cite{Azzuni2022} \\
Message Passing Net. & 2022 & multiple & - & semantic, spatial & local, long-range &\ac{SL} & ConSeP, PanNuke, Lizard, CRCHisto. &\cite{Hassan2022} \\
CDNet & 2021 & Encoder-Decoder (ConvNet) & channel, semantic & semantic, spatial & local & \ac{SL} & KUMAR, CPM-17 & \citep{He2021} \\
ConDANet & 2023 & U-Net & channel, spatial & semantic, spatial & local & \ac{SL} & KUMAR, CPM-17 & \citep{Imtiaz2023} \\
CircleFormer & 2023 & Hybrid Transformer & selg, global & semantic, spatial & local, global, long-range & \ac{SL} & KUMAR & \cite{Zhang2023} \\
SMILE & 2023 & Hybrid Transformer & self, global & semantic/spatial & local, global, long-range & \ac{SL} & ConSEP, MoNuSAC & \citep{Pan2023} \\
ClusterSeg & 2023 & Hybrid Transformer & self, global & semantic, spatial & local, global, long-range, multiscale & \ac{SL} & KUMAR & \citep{Ke2023} \\
TransNuSeg & 2023 & Hybrid Transformer & self, global & semantic, spatial & local, global, long-range, multiscale & \ac{SL} & KUMAR & \citep{He2023} \\
\hline
\end{tabular}
\end{adjustbox}
\captionof{table}{\label{tab:resume_ctx_att_he_wsi}Summary of the reviewed, from 2019 to 2023, context- and attention-based  cell nuclei instance segmentation and classification methods. SSL: Self-supervised Learning; SL: Supervised Learning; WSL: Weakly Supervised Learning.}
\end{table*}

\begin{table*}[h]
\captionsetup{font=normalsize}
\begin{adjustbox}{width=0.8\textwidth, center}
\begin{tabular}{|ccccc|}
\hline
\textbf{Method}         & \textbf{Authors}                   & \textbf{Year} & \textbf{Public Repository}                                                                   & \multicolumn{1}{c|}{\textbf{Reference}} \\ \hline

TransNuSeg & He et al. & 2023 & \url{https://github.com/zhenqi-he/transnuseg} & \cite{He2023} \\
ClusterSeg & Ke et al. & 2023 & \url{https://github.com/lu-yizhou/ClusterSeg} & \cite{Ke2023} \\
SMILE & Pan et al. & 2023 & \url{https://github.com/panxipeng/nuclear_segandcls} & \cite{Pan2023} \\
CircleFormer & Zhang et al. & 2023 & \url{2023 https://github.com/zhanghx-iim-ahu/CircleFormer} & \cite{Zhang2023} \\
Patch. Contrast. Learn. & Boserup et al.                     & 2023          & \url{https://github.com/nickeopti/bach-contrastive-segmentation}      &   \cite{Boserup2023}                                      \\
DRCA-Net                & Dogar et al.                       & 2023          & \url{https://github.com/Ghulam111/DRCA_Net}                    &     \cite{Dogar2023} \\
DAN-NucNet              & Ahmad et al.                       & 2023          & \url{https://github.com/ibtihajahmadkhan/DAN-Nuc-Net}                 &  \cite{Ahmad2023} \\
MSAL-Net                & Ali et al.                         & 2022          & \url{https://github.com/haideralimughal/MSAL-Net}                    &    \cite{Ali2022}\\
HistoSeg                & Wazir and Fraz                     & 2022          & \url{https://github.com/saadwazir/HistoSeg}                           &  \cite{Wazir2022}                                      \\
HyLT                    & Luo et al.                         & 2022          & \url{https://github.com/Roypic/LTUNet}                                & \cite{LuoSelvan2022}                                    \\
RCSAU-Net               & Wang et al.                        & 2022          & \url{https://github.com/antifen/Nuclei-Segmentation}                  & \cite{WangLuo2022} \\
CDNet & He et al. & 2021 & https://github.com/honglianghe/CDNet & \cite{He2021} \\
MedT                    & Valanarasu et al.                  & 2021          & \url{https://github.com/jeya-maria-jose/Medical-Transformer}         &  \cite{Valanarasu2021} \\
Mask Guided Attention   & Guo et al.                         & 2021          & \url{https://github.com/RuoyuGuo/MaskGA_Net}                          & \cite{Guo2021} \\
MASG-GAN                & Zhang   et al. & 2021          & \url{https://github.com/wangpengyu0829/MASG-GAN}                     & \cite{ZhangYuWang2021}                                        \\
NucleiSegNet            & Lal et al.                         & 2021          & \url{https://github.com/shyamfec/NucleiSegNet}                        &  \cite{Lal2021}                                       \\
Self-sup. Scale Pred    & Sahasrabudhe et al.                & 2020          & \url{https://github.com/msahasrabudhe/miccai2020_self_sup_nuclei_seg} &    \cite{Sahasrabudhe2020}                                     \\
NAS-SCAM                & Liu et al.                         & 2020          & \url{https://github.com/ZuhaoLiu/NAS-SCAM}                           &  \cite{Liu2020}                                       \\ \hline
\end{tabular}
\end{adjustbox}
\captionof{table}{\label{tab:resume_source_code}Summary of the reviewed, from 2019 to 2023, context- and attention-based  cell nuclei instance segmentation and classification methods, and that provide public access to the source code.}
\end{table*}

\begin{figure}[!h]
\begin{center}
        \includegraphics[scale=0.6]{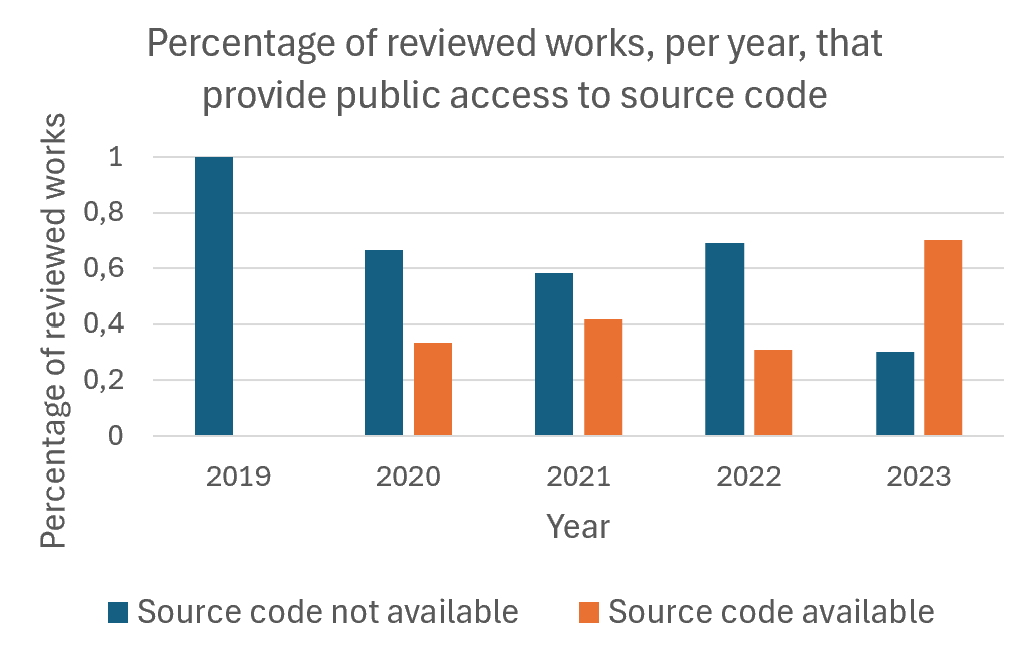}
\caption{Percentage of reviewed publications, per year, that provide public access to source code. We observe an increasing concern with reproducibility, although from 2019 to 2022 only less than 42.5 \% of published works addressed this critical aspect of \ac{ML} research.  \label{fig:source_code_per_year}}
\end{center}
\end{figure}

\begin{figure*}[t]
        \centering
        \includegraphics[scale=0.5]{"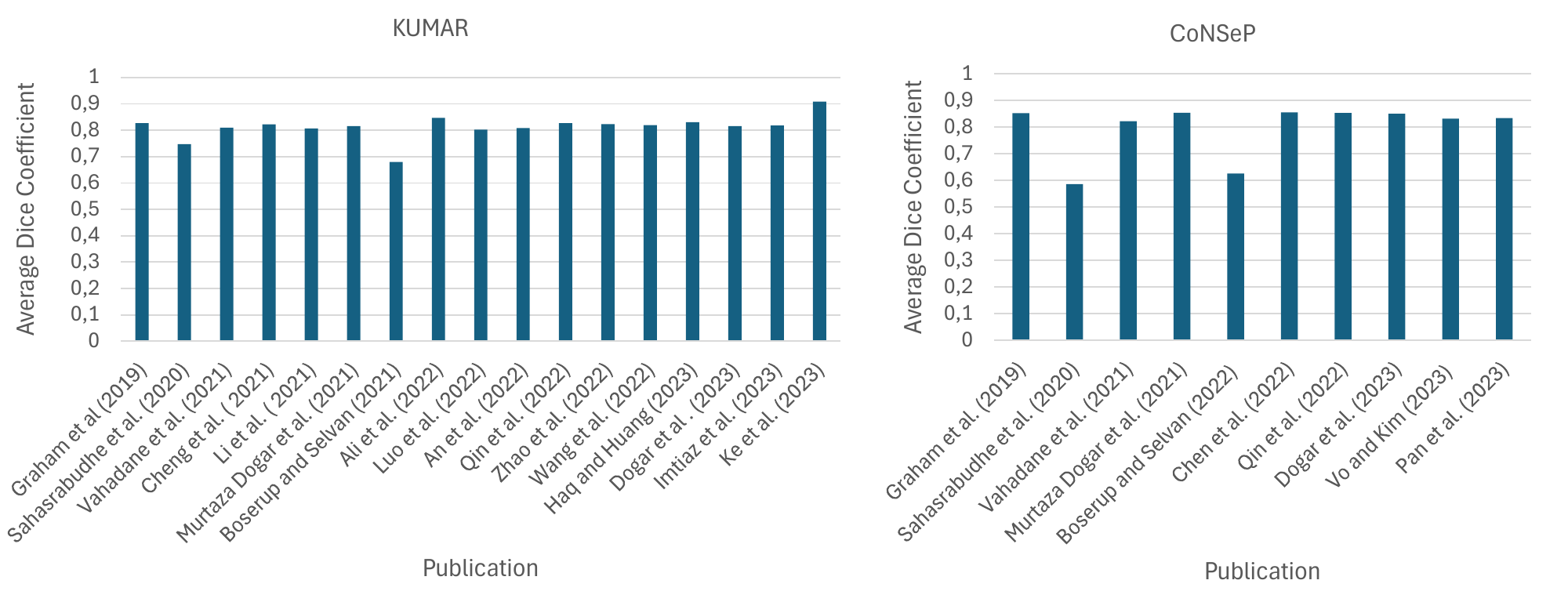"}
\caption{Comparison of \ac{DC} on the CoNSeP \citep{Graham2019b} and KUMAR \citep{Kumar2020} (test) datasets. \label{fig:sota_dice_KUMAR_consep}}
\end{figure*}

\begin{figure*}[t]
        \centering
        \includegraphics[scale=0.5]{"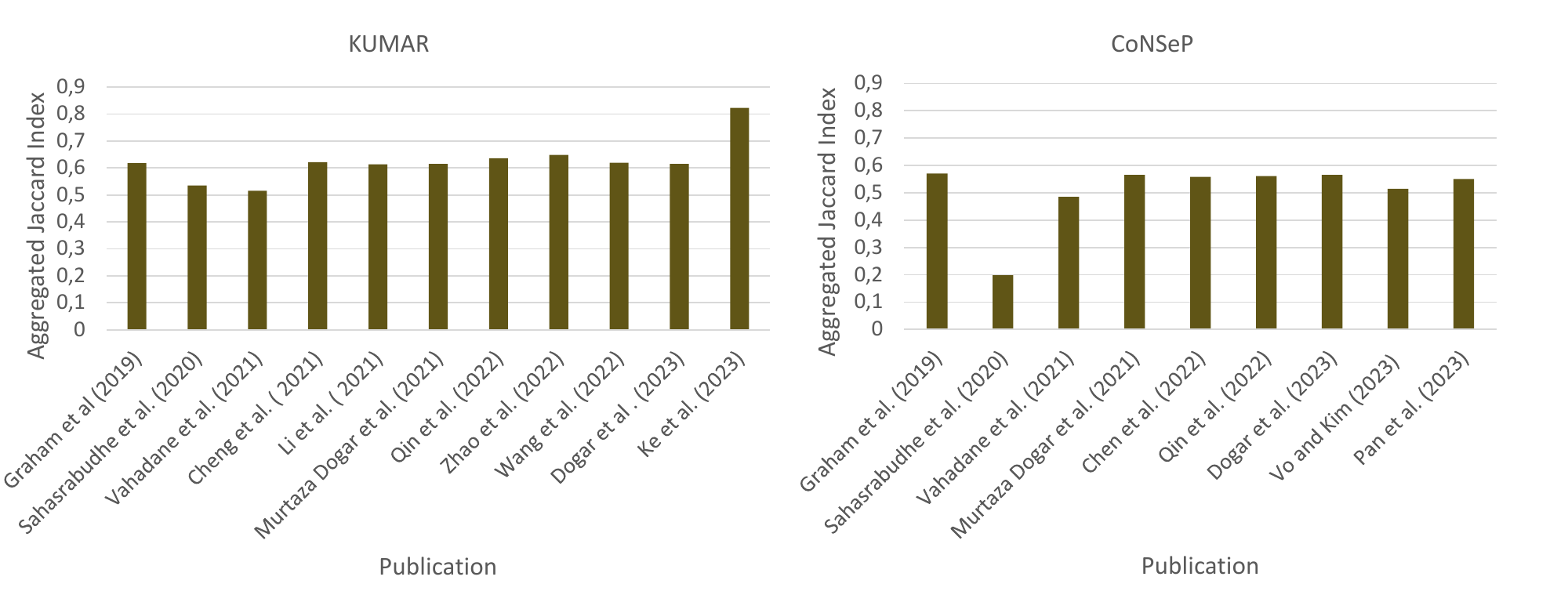"}
\caption{Comparison of \ac{AJI} on the CoNSeP \citep{Graham2019b} and KUMAR \citep{Kumar2020} (test) datasets. \label{fig:sota_aji_KUMAR_consep}}
\end{figure*}

\begin{figure*}[t]
        \centering
        \includegraphics[scale=0.5]{"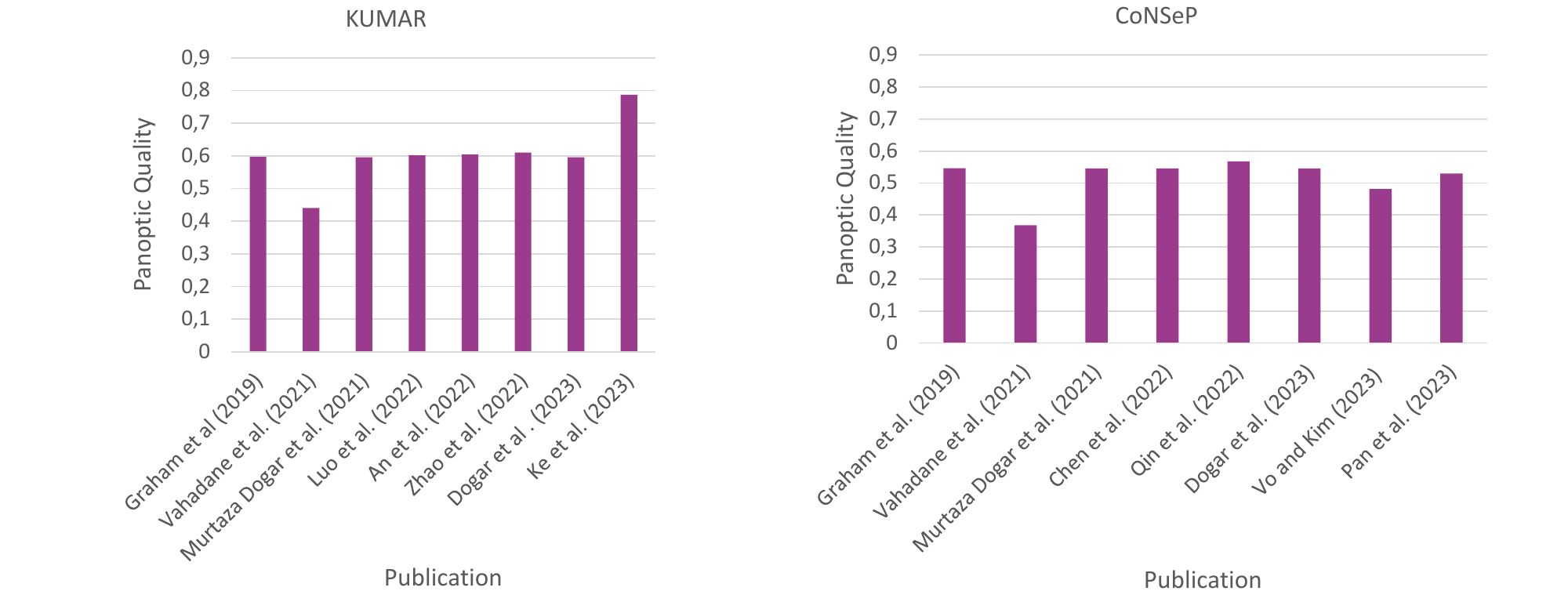"}
\caption{Comparison of \ac{PQ} on the CoNSeP \citep{Graham2019b} and KUMAR \citep{Kumar2020} (test) datasets. \label{fig:sota_pq_KUMAR_consep}}
\end{figure*}

\begin{table}[h]
\begin{adjustbox}{width=\columnwidth, center}
    \begin{tabular}{ccccccc}
\hline
\textbf{Reference}         & \textbf{Type of Segmentation} & \textbf{DC (avg)} & \textbf{F1} & \textbf{JI} & \textbf{AJI} & \textbf{PQ} \\ \hline
Graham et al. (2019) \cite{Graham2019b}    & Instance & 0.8260 & - & - & 0.6180     & 0.5970 \\
Zhou et al. (2019) \cite{ZhouBao2019}    & Instance & - & 0.8458 & - & 0.6306  & - \\ 
Qu et al. (2019) \cite{Qu2019}    & Instance & 0.8054 & 0.8639 & - & 0.6164  & - \\ 
Liu et al. (2022) \citep{Liu2022spn_ien} & Instance & 0.7394 & 0.7863 & 0.6494 & 0.5430 & - \\
Sahasrabudhe et al. (2020) \cite{Sahasrabudhe2020} & Semantic   & 0.7477 & - & - &  0.5354  & - \\
Vahadane et al. (2021) \cite{Vahadane2021}    & Instance  & 0.8100   & - & - & 0.5160   & 0.4400      \\
Cheng et al. (2021)  \cite{Cheng2021boundary} & Semantic  & 0.8227  & - & - &    0.6215    & -           \\
Lal et al. (2021)  \cite{Lal2021} & Semantic  & -   & 0.8136 &  0.6888 &   -    & -           \\
Li et al. (2021) \cite{LiYang2021}  & Semantic & 0.8061  & 0.8602 & -  &   0.6133    & -           \\
Murtaza Dogar et al. (2021) \cite{MurtazaDogar2021}    & Instance   & 0.8150 & - & -  &  0.6150  & 0.5960      \\
Boserup and Selvan (2022) \cite{Boserup2023}  & Semantic  & 0.6797  & - & -  &    -   & -   \\
Ali et al. (2022)  \cite{Ali2022} & Semantic  & 0.8470  & - &  0.7060 &   -    & -           \\
Luo et al. (2022)  \cite{LuoSelvan2022}      & Instance  & 0.8025 & - &  0.6711 &    -     & 0,6021           \\
An et al. (2022)  \cite{An2022}         & Instance  & 0.8080  & - & 0.6826 &    -   & 0.6040           \\
Qin et al. (2022)  \cite{Qin2022}         & Semantic  & 0.8260   & - & - &  0.6360   & -           \\
Zhao et al. (2022)  \cite{ZhaoZuo2022}         & Semantic   & 0.8230 & - &  - &   0.6490    &  0.6100          \\
Wang et al. (2022)  \cite{WangLuo2022}         & Semantic    & 0.8200  & 0.8670 & - & 0.6190     & -           \\
Haq and Huang (2023)  \cite{Haq2022}         & Semantic  & 0.8307   & - &  0.6872 &    -   & - \\
Dogar et al . (2023) \cite{Dogar2023}      & Instance     & 0.8150  & - & - &  0.6150   & 0.5960      \\ 
Ahmad et al. (2023)  \cite{Ahmad2023}         & Semantic  & - & 0.8414 & 0.7403 &  -     & - \\ 
Imtiaz et al. (2023) \citep{Imtiaz2023} & Semantic & 0.8181 & - & 0.7089 & - & -  \\ 
Ke et al. (2023) \citep{Ke2023} & Instance & 0.9080 & - & - & 0.8230 & 0.7870 \\
He et al. (2023) \citep{He2023} & Instance & 0.9081 & 0.8152 & 0.6949 & - & - \\
\hline
\end{tabular}
\end{adjustbox}
\captionof{table}{\label{tab:resume_KUMAR_eval}Summary of state-of-the-art results of context and attention-based neural networks (reviewed from 2019 to 2023) on the KUMAR \citep{Kumar2020} dataset (test). HoVer-Net \cite{Graham2019b}, CIA-Net \cite{ZhouBao2019}, and FullNet \cite{Qu2019} are included as a reference for comparison.}
\end{table}

\begin{table}[h]
\begin{adjustbox}{width=\columnwidth, center}
\begin{tabular}{ccccccc}
\hline
\textbf{Reference}         & \textbf{Type of Segmentation} & \textbf{DC (avg)}  & \textbf{AJI} & \textbf{PQ} \\ \hline
Graham et al. (2019) \cite{Graham2019b}     & Instance                      & 0.8530   & 0.5710      & 0.5470      \\
Doan et al. (2022) \cite{Doan2022}     & Instance                      & 0.8440   & 0.5860      & 0.5400      \\
Sahasrabudhe et al. (2020) \cite{Sahasrabudhe2020} & Semantic                      & 0.5870   & 0.1980   & -           \\
Vahadane et al. (2021) \cite{Vahadane2021}    & Instance                      & 0.8220    &  0.4850      & 0.3680      \\
Murtaza Dogar et al. (2021) \cite{MurtazaDogar2021}    & Instance                      & 0.8500   & 0.5650       & 0.5460      \\
Boserup and Selvan (2022)  \cite{Boserup2023} & Semantic                      & 0.6266   & -      & -           \\
Chen et al. (2022)  \cite{ChenZhao2022}       & Instance                      & 0.8560  &    0.5580    & 0.5460      \\
Qin et al. (2022)  \cite{Qin2022}         & Instance                     & 0.8540   &  0.5610 &  0.5680          \\
Dogar et al. (2023)   \cite{Dogar2023}     & Instance                      & 0.8500   &   0.5650   & 0.5460      \\ 
Vo and Kim (2023)   \cite{Vo2023}     &  Instance                     & 0.8330   &  0.5150   & 0.4820 \\ 
Pan et al. (2023) \citep{Pan2023} & Instance & 0.8350 & 0.5500 & 0.5300 \\

\hline
\end{tabular}
\end{adjustbox}
\captionof{table}{\label{tab:resume_consep_eval}Summary of state-of-the-art results of context and attention-based neural networks (reviewed from 2019 to 2023) on the CoNSeP \citep{Graham2019b} dataset (test). HoVer-Net \cite{Graham2019b} and SONNET \cite{Doan2022} are included as a reference for comparison.}
\end{table}

Overall, most algorithms for cell nuclei instance segmentation and classification revolve around encoder-decoder neural networks, demonstrating the approach's superiority over other alternatives. Table \ref{tab:resume_ctx_att_he_wsi} summarizes the main works (reviewed from 2019 to 2023) on cell nuclei instance segmentation and classification. Furthermore, table \ref{tab:resume_source_code} and Fig. \ref{fig:source_code_per_year} provide insights regarding the source code availability of the reviewed publications. We observe an increasing concern with reproducibility, although from 2019 to 2022 less than 37.5 \% of the reviewed papers address this critical part of \ac{ML} research. Figures \ref{fig:sota_dice_KUMAR_consep}, \ref{fig:sota_aji_KUMAR_consep}, and \ref{fig:sota_pq_KUMAR_consep}, on the other hand, report the performance of reviewed methods on the KUMAR \citep{Kumar2020} and CoNSeP \citep{Graham2019b} datasets. As different publications choose distinct evaluation metrics, not all works are directly comparable. However, most works resort to \ac{DC}, \ac{IoU} (JI), \ac{AJI}, or \ac{PQ}, making these suitable metrics for fair comparisons. We find that most strategies show significant improvements by enriching neural networks with context and attention-based mechanisms, but we observe that while it works satisfactorily well for binary and semantic segmentation tasks, performance is usually lower in instance segmentation and classification (quantified with the \ac{PQ} metric). Furthermore, although transformers are starting to prevail in \ac{CV} in general, \ac{CPath}, and  cell nuclei instance segmentation and classification are still dominated by \acp{ConvNet} (see tables \ref{tab:resume_KUMAR_eval}, and \ref{tab:resume_consep_eval}). A possible explanation could reside in transformers poor performance in small object instances \cite{Cheng2022, Carion2020} or in the limited availability of large-scale annotated datasets. 

In summary, there is a growing trend in context- and attention-based mechanisms for  cell nuclei instance segmentation and classification from \ac{HE}-stained brightfield microscopy images. In general, these methods were shown to increase overall performance, but there is no "one-size fits all" solution as different approaches tackle specific challenges. Contextual information at multiple levels in conjugation with tailored architectures and training recipes hold the promise to tackle the challenges of clustered nuclei with overlapping or touching instances. In turn, spatial and channel attention have been explored to enhance local context and relevant foreground information while suppressing noisy or irrelevant features. Other approaches consider context and attention modules to refine coarse nuclei segmentation maps, especially cell nuclei borders, whereas prior knowledge, like the image magnification level, is useful to learn from unlabelled data general but discriminative features that potentiate finetuning with scarce or incomplete annotations. Besides, there is evidence that context- and attention-based mechanisms enable more effective multi-task learning approaches, therefore allowing the encoding of more stable features simultaneously useful for instance segmentation, instance classification, and semantic segmentation. Notably, by reducing feature map noise, and by capturing inter-instance dependencies, context (e.g., long-range, and multiscale) and attention methods have been extensively used to improve the ability of \ac{DL}-based instance segmentation and classification methods in detecting small instances, a recurring challenge with both \acp{ConvNet} and \acp{ViT}, but an important one as cell nuclei instances are very likely to be small when compared with the total image size (in pixels).

The results of the present review also reveal some misleading practices. For example, comparing neural networks with large differences in the number of parameters or optimizing the hyperparameters of new methodologies while not that of the baseline \acp{ConvNet}. Another frequent practice that leads to misleading conclusions is comparing novel methodologies with works known to have inferior performance than the current top-performing solutions (e.g., comparing with UNet \citep{Ronneberger2015}, or UNet++ \citep{Zhou2018}, instead of with HoVer-Net \citep{Graham2019b}). From a scientific standpoint, this can be of interest and contribute to a better understanding of the different methods. But considering future clinical applications, algorithms with the best performance will always be preferred. Data availability is also a concern    in nuclei instance segmentation and classification. While a pool of datasets is available, many derive from the same source/set of sources. This poses a challenge for assessing model robustness as while algorithms may have similar performance when evaluated in \ac{iid} data, the generalization to \ac{OOD} settings may be undermined in some of the models. Other future directions worthy of careful consideration include data, and annotation efficient \ac{DL}. As a frequent hurdle in medical image analysis is data and annotation availability, improving data efficiency could not only increase the discriminative power of \ac{DL} models with smaller datasets but as well is a promising strategy to leverage some of the already existing large-scale, but unlabelled, datasets. This is where self-supervised, semi-supervised, and \ac{WSL} take a major role, and although the literature reports some significant contributions, we argue most works regard image classification paradigms, whereas dense prediction tasks are mostly open ground. We hold the next breakthroughs in dense prediction in \ac{CPath} could come from exploiting domain knowledge, and from introducing context- and attention-based mechanisms in algorithm design. These ideas can translate to more intrinsically interpretable models, increase model performance, and solve the challenge of generalizing to \ac{OOD} data, a requirement if the models are to be trusted and deployed in clinical practice.

\subsection{Future Research Directions}

Certain methods that potentiate the encoding of contextual information seem to translate to increased generalization (e.g., \acp{ViT} \citep{Ke2023}). However, we observe limited advantages with other approaches \citep{MurtazaDogar2021}. This suggests that although context and attention-based mechanisms hold the potential to increase performance and generalization, it is still not clear what context priors are more useful neither how to optimally explore contextual information in cell nuclei instance segmentation and classification. Furthermore, although domain knowledge is relevant to impose useful inductive biases, we observe contextual information has been disregarded during data collection and annotation which hinders downstream model optimization.  For instance, cell nuclei annotations could be accompanied by pixel-level annotations of tissues or tissue concepts (e.g., glands), which could then be used to develop supervised methods that consider contextual features. In turn, debiasing the effects of context confounders is also relevant for domain generalization, therefore, providing along with the dataset and annotations metadata to help identify irrelevant context variables (e.g., stain manufacturer, scanner provider, patient age and gender, etc.) could aid in developing \ac{DL} models more robust to spurious correlations between instances and their surrounding tissue context. In general, there is the need to understand the most relevant context priors for cell nuclei instance segmentation and classification. Nonetheless, there a few promising research topics that could benefit from context and attention like causal representation learning, multimodal \ac{ML}, and annotation-efficient learning.

\subsubsection{Causal Representation Learning}

Current \ac{DL} methods are based on statistical learning, which relies on the assumption of identically distributed data. Indeed, although we frequently observe generalizability increases with the scale and diversity of training environments, the assumption of identical distribution can still be violated in the testing domains (e.g., due to sample selection bias, interventions, etc.). Furthermore, statistical models are incomplete descriptors of a system, as they rely on the distribution of observed random variables instead of the true generative factors (unobserved). In fact, statistical models could easily learn spurious correlations and thus catastrophically fail under covariate shift \citep{Arjovsky2019}. On the contrary, causal learning aims to uncover the causal structure and underlying causal variables from observations and outcomes, interventions, and distribution shifts \citep{Pearl2009, Peters2017, Scholkopf2021}. As an emerging research direction, causal representation learning could thus lead to more robust predictors, invariant to distribution shifts \citep{Peters2016}. For instance, \citet{Huang2021} suggest a \ac{SCM} for object detection tasks and consider that certain context factors are confounding variables for instances in a given observation. They thus suggest a causal intervention approach to deconfound the effect of context features that they demonstrate to improve generalization. In fact, suitable assumptions for a structural causal model representing the data-generating mechanism of tissue concepts and their corresponding context could lead to a representation that better approximates the ground truth (although unobserved and unknown) latent factors. \cite{Bashir2024} are among the first to suggest a regularization approach to learn context-invariant representations for tissues and cell nuclei semantic segmentation tasks. A causal perspective would allow us to argue that the learned representation is provably (at least partially) causal, as the causal features are the most likely to remain stable in varying tissue contexts. Likewise, preliminary work on causal segmentation in medical imaging as shown some degree of success. For example, a recurring inconvenience in medical imaging is that often training data originates from a single source. This adds to the previously mentioned challenges of \ac{DL} (e.g., batch effects; data imbalance, etc.) thus making single source domain generalization one of the most challenging objectives of \ac{DL}-based segmentation algorithms. But robust learning with causal data augmentation could simulate distribution shifts \citep{Ouyang2023} and deconfound the effects of spurious context correlations \citep{Ouyang2023}.

\begin{figure*}[h]
  \centering
 \includegraphics[scale=0.4]{"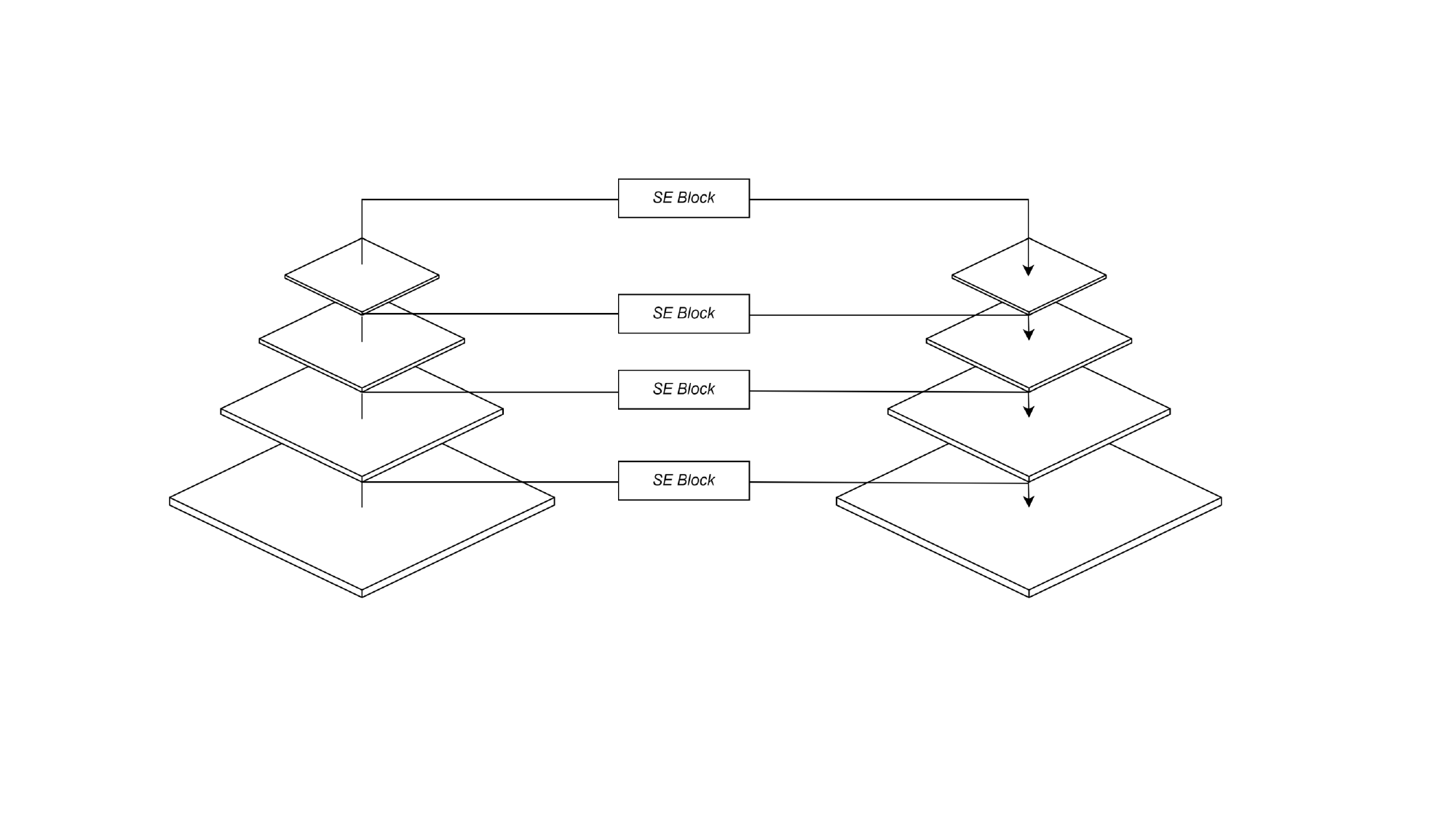"}
  \caption{Squeeze and Excitation-Feature Pyramid Network (SE-FPN). We add a SE-block before every lateral connection of the FPN module.}
  \label{fig:SEFPN}
\end{figure*}

\begin{figure*}[h]
  \centering
 \includegraphics[scale=0.4]{"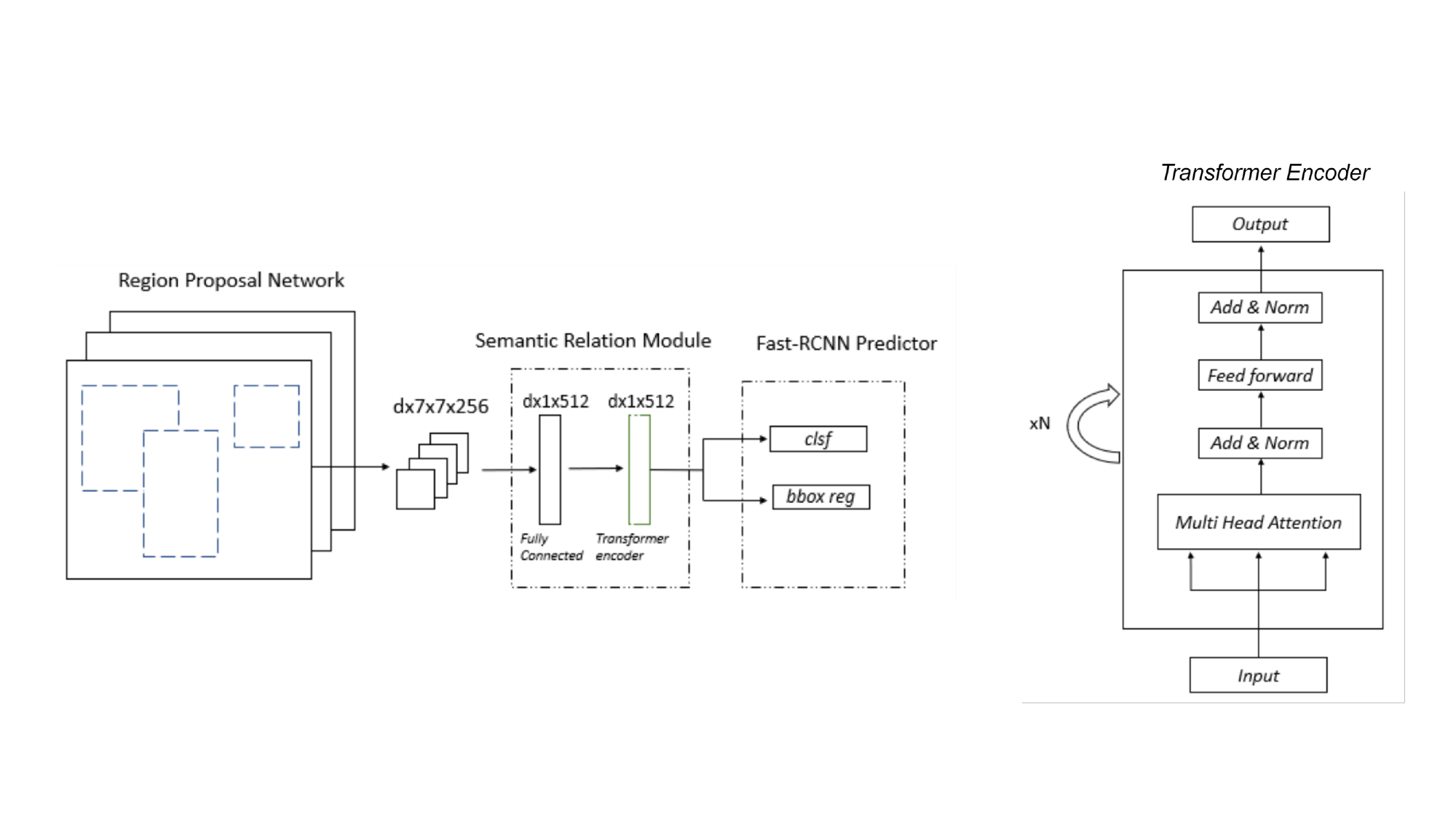"}
  \caption{Left: Semantic Relation Module. We replace the second fully connected layer of the \ac{MLPHead} with a transformer encoder. Right: Transformer Encoder.}
  \label{fig:SRM}
\end{figure*}

\subsubsection{Graph Neural Networks}

Not all learning tasks are defined in the Euclidean space (e.g., social network-based recommender systems, molecular structure prediction, etc.), meaning algorithms operating in grid-like data (e.g., \acp{ConvNet}) are not appropriate for these tasks. At its core, \ac{WSI} analysis and  cell nuclei instance segmentation and classification could also be defined in a non-Euclidean space as tissues and cells possess intricate relationships and interdependencies. With this view, \acp{ViT} are a more flexible alternative as they allow reasoning over long-range context, but \acp{ViT} are a special case of a more general and flexible family of parametric hypothesis, namely \acp{GNN} \citep{Velivckovic2023}. \acp{GNN} represent data in a graph structure defined as a set of node vectors (objects) connected by edges (modelling inter-object relationships).  Multiple tasks are thus allowed with \acp{GNN} such as node classification, link prediction, or node clustering. For instance, \citet{Hassan2022} use domain knowledge and, starting from a set of pre-detected and segmented nuclei, propose a \ac{GNN} (\ac{MPN}) where each node represents a nucleus and edges encode the relation from each nucleus to its neighbours. Through reasoning over spatial and semantic context, the suggested \ac{GNN} method allows the clustering of cell nuclei, which is used to refine nuclei classification on the premise that each nucleus in a cluster should be of the same cell type. 

Due to the ability of \acp{GNN} to learn better object level features \citep{Xu2022} and inter-instance relations \citep{Yang2020}, these enable a more accurate instace segmentation in challenging settings like touching and overlapping instances \citep{Xu2022, Yang2020}. Likewise, by encoding inter-instance relations (context inductive biases), GNNs demonstrate potential to refine instance masks in cluttered scenarious \citep{Xie2022gnn}, a commonly found condition in \ac{HE}-stained microscopy images. But these are still largely underexplored in \ac{CPath}, leaving room for further research in this direction to design algorithms that can effectively encode contextual priors for reliable cell nuclei instance segmentation and classification.

\subsubsection{Multimodal Machine Learning} 

Until recently, most \ac{ML} models are optimized to achieve a decision using data from a single modality. However, when making decisions, clinical experts commonly consider data from multiple streams of information (e.g., imaging data, clinical reports, molecular profiling data, etc.). Therefore, to close the gap between artificial and biological intelligence, \ac{ML} algorithms should also consider integrating data from multiple modalities. The complementarity of multiple data types thus gives more context (i.e., semantic context/ prior-knowledge) to the learned representations, which could increase generalization. Nonetheless, there are a few challenges, namely regarding \emph{what} types of data to combine, \emph{when} to use multimodality vs single modality and \emph{how} to best fuse the multiple information streams. In a recent example, \cite{Li2024} design an attention-based multi-modal approach for unsupervised medical imaging segmentation tasks, where a transformer-based text-guided model is shown to improve the quality of generated pseudo-labels for unlabeled medical images. Interestingly, the authors also show the effectiveness of the approach in a nuclei segmentation task. 

Nonetheless, a challenge with multimodal \ac{ML} is the availability of datasets with paired samples of multiple data types. For instance, representation learning with language and imaging data \citep{Huang2023, Lu2023, Lai2023, Naseem2023} is an emerging direction in \ac{CPath}. But if we consider other modalities, like radiology-histopathology \citep{Hamidinekoo2021}, genomics-histopathology \citep{Chen2022mml}, or radiology-genomics-histopathology \citep{Chen2021MML} data integration, literature and available datasets are only slowly gaining interest \citep{Vanguri2022}. But to advance future efforts, experts should start by collecting, curating, and annotating large-scale multimodal datasets that could then be used to devise more robust algorithms, not only for nuclei instance segmentation and classification, but also for downstream applications that integrate nuclear features with other data modalities. Depending on the available data, (e.g., radiology-histopathology), this could even correspond to learning a representation expressing contextual information like global, hierarchical and prior context . Moreover, aside from integrated language and vision data \citep{Wu2023}, we observe there is no current literature reporting the integration of multiple data modalities to improve nuclei instance segmentation and classification tasks. Therefore, although multimodal approaches are believed to improve model performance and \ac{OOD} generalization, there is no consensus if such an approach would benefit nuclei instance segmentation and classification tasks. If we bring context and attention to the discussion, further questions arise concerning primarily if and how these mechanisms could be used to better combine multiple data types.

\subsubsection{Self-supervised Learning}

Self-supervised pretraining on large \ac{CV} datasets or on large volumes of unlabeled histopathology data followed by task-specific fine-tuning has been shown to improve model performance \citep{Ciga2022, KangPereira2022}. Indeed, self-supervised pathology-tailored foundation models are effective in solving multiple tasks, even generalizing to distributions beyond those seen during training \citep{Filiot2023, Huang2023, Wang2022ssl}. But while \ac{SSL} methods were shown to generalize better than \ac{SL} on \ac{OOD} data \citep{Shi2023}, these are still prone to learning spurious correlations \citep{Hamidieh2022, Chhipa2023}, particularly on subgroups least represented in the dataset \citep{Hamidieh2022}. However, as \ac{HE}-stained brightfield microscopy images are rich in  contextual information, a viable solution to improve generalization could  stem from using context priors and attention mechanisms to  impose inductive biases that enable \ac{SSL} algorithms more robust to distribution shifts. For instance, the self-attention mechanism in \acp{ViT} has been shown to decrease model bias towards spurious correlations \citep{Paul2022, Ghosal2023}, especially considering larger models and large-scale pretraining \citep{Zhang2022ViTGen, Ghosal2023}, which could be attributed to learning better contextual dependencies \citep{Paul2022}.

In turn, some \ac{SSL} paradigms promote invariance to synthetic data transformations, which increases robustness to real world  transforms (i.e., distribution shifts). The effectiveness of these methods is higher the better the chosen transforms simulate expected spurious factors of variation, but as the challenge of generalizing cell nuclei instance segmentation and classification algorithms is still unsolved, more effective (domain specific) data augmentation strategies need to be devised. Here, context priors could also play a major role as certain transformations could corrupt the semantics of the data. In fact, there is work showing the effectiveness of context-driven data augmentation not only for dense prediction tasks in \ac{CV}\citep{Dvornik2021}, but also for cell nuclei instance segmentation \citep{Lin2022}, where a \ac{GAN}-based approach is used to achieve context consistency after mixing the instances of two images.

\subsubsection{Weakly Supervised Learning}

One of the key strengths of \ac{WSL} is the ability to harness large-scale datasets with minimal manual labeling efforts, easing the need for exhaustive annotation. This is particularly valuable in \ac{CPath} where collecting fully labeled data is impractical. Several strategies can be adopted for weakly-supervised cell nuclei instance segmentation and classification. For example, generative models show promise in solving several segmentation tasks \citep{Li2021, Lei2020, Nema2020, Majurski2019}, but could be adapted to \ac{WSL}. DetectGAN is suggested for data-efficient object detection in medical imaging \citep{LiuJia2019}. Generative adversarial learning and template matching allow localizing objects in an image with weak supervision \citep{Diba2019}, and another work shows the potential of \acp{GAN} for weakly supervised brain tumour segmentation \citep{Yoo2022}. An idea we suggest is \ac{VAE} regularization, which has been successful in \ac{SL} brain tumour segmentation \citep{Myronenko2019}. The design could be extended with context encoding \citep{Zimmerer2018} and adapted to weak supervision (e.g., point annotations). This could increase model robustness to input perturbations. Moreover, besides promoting \ac{OOD} generalization, generative models also have the advantage of producing disentangled and interpretable representations, especially if jointly optimized for reconstruction fidelity and interpretability \citep{Tameem2018}. Another viable future direction in weakly supervised nuclei instance segmentation and classification is prior knowledge to guide the learning task \citep{Zhang2021Hsuan}. For instance, predicting spatial semantic context, like cellular communities' orientation along an image (or patch in \ac{MIL}), could be exploited in a multi-task setting as a regularization strategy.

In turn, \ac{SAM} \citep{Kirillov2023} is introduced as a foundation model for image segmentation. But \ac{SAM} performance is lower when directly applied to histopathology images \citep{Zhang2023SamPath, Huang2024}, as the covariate and concept shifts limit the zero-shot performance in histopathology segmentation tasks. Notably, this is exacerbated in cell nuclei segmentation \citep{Deng2023}, underscoring the effectiveness of domain-specific models and domain specific pretraining as the prevailing paradigms. Yet, there is evidence that generalization increases with the number of input prompts \citep{Deng2023, Xu2024}, hinting that \ac{SAM}  and equivalent foundation models could alleviate the annotation burden by predicting cell nuclei pseudo-masks from point or bounding box annotations for \ac{WSL} methods. This unveils a particularly interesting application of \ac{SAM} and equivalent foundation models: predicting relevant context priors in histopathology images from only a few input prompts, like cellular communities, tumours, etc. These priors could then be used to optimize cell nuclei instance segmentation and classification models, for example, as pseudo-labels or as additional input variables (i.e., attention maps). Another promising use case for these foundation models is data augmentation. For instance, using the U-Net \citep{Ronneberger2015} architecture in a cell nuclei segmentation task, \cite{Zhang2023SAMAug} augment \ac{HE}-stained microscopy images with boundary and segmentation prior attention maps generated from \ac{SAM}, which lead to improved performance.

Overall, despite the promises of \ac{WSL}, a major limitation is the generation of noisy/incorrect pseudo-masks. This could be attributed to confounding contextual information, but recent work demonstrates the potential of causal interventions to enable more effective \ac{WSL} as this approach leads to more reliable pseudo-masks \citep{Chen2022c, Zhang2020Causal}.

\section{Colon Nuclei Instance Segmentation and Classification: A Multi-centre Comparative Study}
\label{sec:comp_study}

In this section, we consider a case study with a comparative analysis to assess how extending an instance segmentation and classification model, namely Mask-RCNN \citep{He2016}, with context and attention-based mechanisms impacts model performance and generalization. We aim to assess if simple mechanisms offer any advantages or if more tailored solutions would be required. We thus extend a baseline Mask-RCNN \citep{He2016} with \ac{SE}-blocks \citep{Hu2018} as well as transformer layers.  

\subsection{Datasets}

This work makes use of the Lizard dataset \citep{Graham2021a} consisting of \ac{HE}-stained \ac{WSI} tiles, from six different sources at 20x objective magnification. The data contains annotations for six nuclei types, namely, epithelial cells, connective tissue, plasma, lymphocytes, neutrophils, and eosinophils. This is one of the biggest fully annotated public datasets for cell nuclei segmentation, with a total of 431,913 annotated nuclei. Despite this, the data are highly imbalanced. For example, the lymphocytes category contains 210,372 annotated nuclei, whereas the eosinophils class represents less than $1\%$ of all instances. For training and validation, we consider data from the 5 centres on the training set of the CoNIC challenge \cite{Graham2021b}. The experiments in this work thus include 5 data sources: GLaS \citep{Sirinukunwattana2017}, CRAG \citep{Graham2019a}, CoNSeP \citep{Graham2019b}, DigestPath \citep{Da2022}, and PanNuke \citep{Gamper2020}. To evaluate the generalizability of the context- and attention-based mechanisms developed through this work, we resort to \textit{leave-one-domain-out} validation \citep{Zhou2022}, where each individual source is treated as a domain. 

\begin{table*}[h]
\captionsetup{font=scriptsize}
\begin{adjustbox}{width=1\textwidth}
\begin{tabular}{cccccccccccccc}
    \multicolumn{14}{c}{\textbf{Comparison of Mask-RCNN-ResNet50-FPN with Mask-RCNN-ResNet50-SE-FPN, and Mask-RCNN-ResNet50-FPN-SRM}} \\ \hline
\textbf{model} &
  \textbf{val set} &
  $\mathbf{map_{50}}$ \textbf{bbox} &
  $\mathbf{map_{50}}$ \textbf{segm} &
  \multicolumn{1}{c}{\textbf{neu}} &
  \multicolumn{1}{c}{\textbf{epi}} &
  \multicolumn{1}{c}{\textbf{lym}} &
  \multicolumn{1}{c}{\textbf{pls}} &
  \multicolumn{1}{c}{\textbf{eos}} &
  \multicolumn{1}{c}{\textbf{con}} &
  \multicolumn{1}{c}{\textbf{pq}} &
  \multicolumn{1}{c}{\textbf{pq det}} &
  \multicolumn{1}{c}{\textbf{pq seg}} &
  \multicolumn{1}{c}{\textbf{multi pq+}} \\ \hline
\multirow{5}{*}{\textbf{mask-rcnn-resnet50-fpn}} &
  glas &
  0.7348 &
  0.7504 &
  0.1464 &
  0.5107 &
  0.5439 &
  0.3283 &
  0.2935 &
  0.4510 &
  0.5309 &
  0.6868 &
  0.7658 &
  0.3790 \\
 &
  dpath &
  0.5826 &
  0.5848 &
  0.0200 &
  0.5228 &
  0.5307 &
  0.3093 &
  0.2142 &
  0.4610 &
  0.5792 &
  0.7480 &
  0.7685 &
  0.3430 \\
 &
  cons &
  0.8707 &
  0.9642 &
  0.3645 &
  0.4998 &
  0.5457 &
  0.4333 &
  0.4947 &
  0.5691 &
  0.5654 &
  0.7414 &
  0.7620 &
  0.4845 \\
 &
  pann &
  0.4819 &
  0.4831 &
  0.4612 &
  0.5254 &
  0.5186 &
  0.3184 &
  0.3241 &
  0.5378 &
  0.5689 &
  0.7344 &
  0.7450 &
  0.4476 \\
 &
  crag &
  0.8091 &
  0.8017 &
  0.0808 &
  0.5237 &
  0.5152 &
  0.3539 &
  0.3010 &
  0.5004 &
  0.5197 &
  0.6710 &
  0.7247 &
  0.3792 \\
  &
  overall &
   0.6958 $\pm$ 0.1439 &
   0.7168 $\pm$ 0.1683 &
   0.2146 $\pm$ 0.1695 &
   0.5165 $\pm$ 0.0098 &
   0.5308 $\pm$ 0.0125 &
   0.3486 $\pm$ 0.0449 &
   0.3255 $\pm$ 0.0923 &
   0.5039 $\pm$ 0.0448 &
   0.5528 $\pm$ 0.0232 &
   0.7163 $\pm$ 0.0313 &
   0.7532 $\pm$ 0.0164 &
   0.4066 $\pm$ 0.0516 \\
  \hline 
\multirow{5}{*}{\textbf{mask-rcnn-resnet50-se-fpn}} &
  glas &
  0.7496 &
  \multicolumn{1}{c}{0.7534} &
  \multicolumn{1}{c}{0.1642} &
  \multicolumn{1}{c}{0.5199} &
  \multicolumn{1}{c}{0.5419} &
  \multicolumn{1}{c}{0.3471} &
  \multicolumn{1}{c}{0.3114} &
  \multicolumn{1}{c}{0.4459} &
  \multicolumn{1}{c}{0.5317} &
  \multicolumn{1}{c}{0.6886} &
  \multicolumn{1}{c}{0.7661} &
  \multicolumn{1}{c}{0.3884} \\
 &
  dpath &
  0.5746 &
  0.5408 &
  \multicolumn{1}{c}{0.0172} &
  \multicolumn{1}{c}{0.5285} &
  \multicolumn{1}{c}{0.5586} &
  \multicolumn{1}{c}{0.3249} &
  \multicolumn{1}{c}{0.2168} &
  \multicolumn{1}{c}{0.4709} &
  \multicolumn{1}{c}{0.5859} &
  \multicolumn{1}{c}{0.7556} &
  \multicolumn{1}{c}{0.7692} &
  \multicolumn{1}{c}{0.3528} \\
 &
  cons &
  0.8912 &
  0.9355 &
  \multicolumn{1}{c}{0.3463} &
  \multicolumn{1}{c}{0.4951} &
  \multicolumn{1}{c}{0.5453} &
  \multicolumn{1}{c}{0.4464} &
  \multicolumn{1}{c}{0.4779} &
  \multicolumn{1}{c}{0.5731} &
  \multicolumn{1}{c}{0.5629} &
  \multicolumn{1}{c}{0.7389} &
  \multicolumn{1}{c}{0.7607} &
  \multicolumn{1}{c}{0.4807} \\
 &
  pann &
  0.6500 &
  0.6500 &
  \multicolumn{1}{c}{0.4400} &
  \multicolumn{1}{c}{0.5300} &
  \multicolumn{1}{c}{0.5200} &
  \multicolumn{1}{c}{0.3200} &
  \multicolumn{1}{c}{0.3400} &
  \multicolumn{1}{c}{0.5400} &
  \multicolumn{1}{c}{0.5700} &
  \multicolumn{1}{c}{0.7300} &
  \multicolumn{1}{c}{0.7500} &
  \multicolumn{1}{c}{0.4500} \\
 &
  crag &
  0.7542 &
  0.7630 &
  \multicolumn{1}{c}{0.0661} &
  \multicolumn{1}{c}{0.5304} &
  \multicolumn{1}{c}{0.5245} &
  \multicolumn{1}{c}{0.3820} &
  \multicolumn{1}{c}{0.2778} &
  \multicolumn{1}{c}{0.5100} &
  \multicolumn{1}{c}{0.5235} &
  \multicolumn{1}{c}{0.6747} &
  \multicolumn{1}{c}{0.7271} &
  \multicolumn{1}{c}{0.3818} \\
&
    overall &
    0.7239 $\pm$ 0.0977 &
    0.7285 $\pm$ 0.1198 &
    0.2068 $\pm$ 0.1480 &
    0.5208 $\pm$ 0.0122 &
    0.5381 $\pm$ 0.0129 &
    0.3641 $\pm$ 0.0426 &
    0.3248 $\pm$ 0.0793 &
    0.5080 $\pm$ 0.0418 &
    0.5548 $\pm$ 0.0215 &
    0.7176 $\pm$ 0.0281 &
    0.7546 $\pm$ 0.0139 &
    \textbf{0.4106 $\pm$ 0.0432}
     \\
  \hline
  \multirow{5}{*}{\textbf{mask-rcnn-resnet50-fpn-srm}} &
  glas &
  0.6427 & 												

  \multicolumn{1}{c}{0.3052} &
  \multicolumn{1}{c}{0.0791} &
  \multicolumn{1}{c}{0.4520} &
  \multicolumn{1}{c}{0.3735} &
  \multicolumn{1}{c}{0.2445} &
  \multicolumn{1}{c}{0.0631} &
  \multicolumn{1}{c}{0.2681} &
  \multicolumn{1}{c}{0.4317} &
  \multicolumn{1}{c}{0.5694} &
  \multicolumn{1}{c}{0.7469} &
  \multicolumn{1}{c}{0.2467} \\
 & 												

  dpath &

  \multicolumn{1}{c}{0.5498} &
  \multicolumn{1}{c}{0.5498} &
  \multicolumn{1}{c}{0.0205} &
  \multicolumn{1}{c}{0,5507} &
  \multicolumn{1}{c}{0.4668} &
  \multicolumn{1}{c}{0.2251} &
  \multicolumn{1}{c}{0.1005} &
  \multicolumn{1}{c}{0.3751} &
  \multicolumn{1}{c}{0.5477} &
  \multicolumn{1}{c}{0.6971} &
  \multicolumn{1}{c}{0.7782} &
  \multicolumn{1}{c}{0.2898} \\
 &
  cons & 										

  0.8575 &
  0.8908 &
  \multicolumn{1}{c}{0.2656} &
  \multicolumn{1}{c}{0.4589} &
  \multicolumn{1}{c}{0.5008} &
  \multicolumn{1}{c}{0.3413} &
  \multicolumn{1}{c}{0.4366} &
  \multicolumn{1}{c}{0.5222} &
  \multicolumn{1}{c}{0.5314} &
  \multicolumn{1}{c}{0.6996} &
  \multicolumn{1}{c}{0.7580} &
  \multicolumn{1}{c}{0.4209} \\
 &
  pann & 													

  0.8574 &
  0.8676 &
  \multicolumn{1}{c}{0.3715} &
  \multicolumn{1}{c}{0.5032} &
  \multicolumn{1}{c}{0.4823} &
  \multicolumn{1}{c}{0.2685} &
  \multicolumn{1}{c}{0.2957} &
  \multicolumn{1}{c}{0.4891} &
  \multicolumn{1}{c}{0.5357} &
  \multicolumn{1}{c}{0.6935} &
  \multicolumn{1}{c}{0.7423} &
  \multicolumn{1}{c}{0.4017} \\
 &
  crag & 											
  0.7078 &
  0.7124 &
  \multicolumn{1}{c}{0.045} &
  \multicolumn{1}{c}{0.4992} &
  \multicolumn{1}{c}{0.4604} &
  \multicolumn{1}{c}{0.2979} &
  \multicolumn{1}{c}{0.1373} &
  \multicolumn{1}{c}{0.4530} &
  \multicolumn{1}{c}{0.4860} &
  \multicolumn{1}{c}{0.6342} &
  \multicolumn{1}{c}{0.7180} &
  \multicolumn{1}{c}{0.3155} \\
&
    overall &

    0.7230 $\pm$ 0.1102 &
    0.6652 $\pm$ 0.1987 &
    0.1563 $\pm$  0.1259 &
    0.4928 $\pm$ 0.0324 &
    0.4568 $\pm$ 0.0401 &
    0.2755 $\pm$ 0.0374 &
    0.2066 $\pm$ 0.1275 &
    0.4215 $\pm$ 0.0831 &
    0.5065 $\pm$ 0.0391 &
    0.6588 $\pm$ 0.0464 &
    0.7487 $\pm$ 0.0180 &
    0.3349 $\pm$ 0.0606  \\
  \hline
  \multirow{5}{*}{\textbf{mask-rcnn-se-resnet50-fpn}} &
glas &
0.6365 &
0.6617 & 
0.1266 & 
0.3689 & 
0.4779 & 
0.2797 & 
0.1089 & 
0.3923 & 
0.4652 & 
0.6047 & 
0.763  & 
0.2924 \\
&
dpath &
0.4506 & 
0.4655 &  
0.0124 & 
0.3476 & 
0.5284 & 
0.1402 & 
0.1272 & 
0.3426 & 
0.5019 & 
0.6538 & 
0.7613 & 
0.2497 \\
&
cons &
0.8886 & 
0.9510  & 
0.3999 & 
0.4556 & 
0.5315 & 
0.2688 & 
0.3237 & 
0.4972 & 
0.5632 & 
0.7357 & 
0.7649 & 
0.4128 \\
&
pann &
0.5010  & 
0.5180  & 
0.4633 & 
0.4860  & 
0.4814 & 
0.2975 & 
0.3035 & 
0.5149 & 
0.5574 & 
0.7202 & 
0.7441 & 
0.4244 \\
&
crag &
0.6654 &
0.6794 &
0.0308 &
0.3441 &
0.4553 &
0.3106 &
0.0325 &
0.3471 &
0.4151 &
0.537  &
0.7154 &
0.2534 \\
 &
    overall &
0.6284	$\pm$ 0.1530 &
0.6551	$\pm$ 0.1691 &
0.2066	$\pm$ 0.1888 &
0.4004	$\pm$ 0.0589 &
0.4949	$\pm$ 0.0300 &
0.2594	$\pm$ 0.0613 &
0.1792	$\pm$ 0.1145 &
0.4188	$\pm$ 0.0735 &
0.5006	$\pm$ 0.0561 &
0.6503	$\pm$ 0.0736 &
0.7497	$\pm$ 0.0187 &
0.3265 $\pm$ 0.0767 
     \\
\hline 

\multirow{5}{*}{\textbf{mask-rcnn-srm-resnet50-fpn}} &

glas & 0.6751 & 0.1529 & 0      & 0.455  & 0.5157 & 0.0851 & 0      & 0.311  & 0.4825 & 0.6304 & 0.7577 & 0.2278 \\ &
dpath & 0.4766 & 0.3293 & 0.1898 & 0.4739 & 0.4702 & 0.309  & 0.1219 & 0.4009 & 0.5484 & 0.7109 & 0.765  & 0.3276 \\ &
cons & 0.7075 & 0.5339 & 0      & 0.2673 & 0.4485 & 0.0502 & 0      & 0.2034 & 0.3507 & 0.4826 & 0.7217 & 0.1616 \\ &
pann & 0.8238 & 0.6291 & 0      & 0.3049 & 0.2915 & 0.1237 & 0      & 0.2049 & 0.3812 & 0.5041 & 0.74   & 0.1542 \\ &
crag & 0.6508 & 0.4806 & 0.1564 & 0.4633 & 0.5072 & 0.2515 & 0.0811 & 0.3975 & 0.4607 & 0.5885 & 0.7142 & 0.3095 \\ &
overall & 0.6654 $\pm$ 0.1121 & 0.5393 $\pm$ 0.1122 & 0.0308 $\pm$ 0.0855 & 0.3441 $\pm$ 0.0882 & 0.4553 $\pm$  0.0813 & 0.3106 $\pm$  0.0995 & 0.0325 $\pm$ 0.0514 & 0.3471 $\pm$  0.0873 & 0.4151 $\pm$ 0.0711 & 0.5370 $\pm$ 0.0836 & 0.7154 $\pm$ 0.0197 & 0.2534 $\pm$ 0.0722  \\
\hline    
\end{tabular}
\end{adjustbox}
\captionof{table}{ \label{tab:se_vs_baseline} Comparison of Mask-RCNN-ResNet50-FPN with Mask-RCNN-ResNet50-SE-FPN, Mask-RCNN-SE-ResNet50-FPN, Mask-RCNN-SRM-ResNet50-FPN, and Mask-RCNN-ResNet50-FPN-SRM in a colon nuclei instance segmentation and classification task. The initials ``neu'', ``epi'', ``lym'', ``pls'', ``eos'', and ``con'' indicate $m\mathcal{PQ}^+$ for each nucleus type. Results in bold indicate the best $m\mathcal{PQ}^+$.}
\end{table*}

\begin{table*}[h]
\captionsetup{font=scriptsize}
\begin{adjustbox}{width=1\textwidth}
\begin{tabular}{cccccccccccccccccccccc}
    \multicolumn{20}{c}{\textbf{Paired Samples T-Test for the Mask-RCNN Experiments}} \\ \hline
\multirow{2}{*}{\textbf{null hypothesis}} & \multirow{2}{*}{\textbf{val set}} & \multicolumn{2}{c}{\textbf{neu}} & \multicolumn{2}{c}{\textbf{epi}} & \multicolumn{2}{c}{\textbf{lym}} & \multicolumn{2}{c}{\textbf{pls}} & \multicolumn{2}{c}{\textbf{eos}} & \multicolumn{2}{c}{\textbf{con}} & \multicolumn{2}{c}{\textbf{pq\_det}} & \multicolumn{2}{c}{\textbf{pq\_seg}} & \multicolumn{2}{c}{\textbf{pq}} & \multicolumn{2}{c}{\textbf{mpq+}} \\ \cline{3-22} 
 &  & \textbf{t-stat} & \textbf{p-val} & \textbf{t-stat} & \textbf{p-val} & \textbf{t-stat} & \textbf{p-val} & \textbf{t-stat} & \textbf{p-val} & \textbf{t-stat} & \textbf{p-val} & \textbf{t-stat} & \textbf{p-val} & \textbf{t-stat} & \textbf{p-val} & \textbf{t-stat} & \textbf{p-val} & \textbf{t-stat} & \textbf{p-val} & \textbf{t-stat} & \textbf{p-val} \\ \hline
\multirow{5}{*}{\textbf{mask-rcnn-resnet50-se-fpn \textgreater{} mask-rcnn-resnet50-fpn}} & glas & 0.7320 & 0.4644 & 5.2427 & \textbf{\textless{}0.0001} & -0.2195 & 0.8263 & 1.5973 & 0.1106 & 2.8566 & \textbf{0.0044} & -1.4217 & 0.1555 & 3.6597 & \textbf{0.0003} & -1.0845 & 0.2785 & 3.4555 & \textbf{0.0006} & 3.1655 & \textbf{0.0016} \\
 & dpath & -0.9670 & 0.3337 & 6.2508 & \textbf{\textless{}0.0001} & 12.0822 & \textbf{\textless{}0.0001} & 2.6934 & \textbf{0.0071} & 0.8739 & 0.3281 & 5.8705 & \textbf{\textless{}0.0001} & 9.0679 & \textbf{\textless{}0.0001} & 1.8547 & 0.0638 & 10.9622 & \textbf{\textless{}0.0001} & 9.9595 & \textbf{\textless{}0.0001} \\
 & cons & 0.8869 & 0.3785 & -1.3743 & 0.1742 & -0.8178 & 0.4166 & 0.6014 & 0.5498 & -0.9826 & 0.3296 & 0.8242 & 0.4129 & -0.5589 & 0.5782 & -1.0243 & 0.3096 & -0.7573 & 0.4517 & 0.1941 & 0.8467 \\
 & pann & -0.6228 & 0.5347 & 1.8002 & 0.0745 & 2.1394 & \textbf{0.0346} & -1.3756 & 0.1717 & -0.5498 & 0.5836 & 0.9819 & 0.3283 & -0.4647 & 0.643 & 2.1738 & 0.0318 & 0.0467 & 0.9628 & 0.4549 & 0.6501 \\
 & crag & -3.0945 & \textbf{0.0020} & 5.5218 & \textbf{\textless{}0.0001} & 7.8009 & \textbf{\textless{}0.0001} & 5.2249 & \textbf{\textless{}0.0001} & -2.9030 & \textbf{0.0037} & 7.4053 & \textbf{\textless{}0.0001} & 3.8205 & \textbf{0.0001} & 3.6464 & \textbf{0.0003} & 5.3357 & \textbf{\textless{}0.0001} & 8.5734 & \textbf{\textless{}0.0001} \\ \hline
\multirow{5}{*}{\textbf{mask-rcnn-resnet50-se-fpn \textgreater{} mask-rcnn-resnet50-fpn-srm}} & glas & 6.8748 & \textbf{\textless{}0.0001} & 11.7056 & \textbf{\textless{}0.0001} & 22.5822 & \textbf{\textless{}0.0001} & 11.2663 & \textbf{\textless{}0.0001} & 12.8686 & \textbf{\textless{}0.0001} & 37.4220 & \textbf{\textless{}0.0001} & 27.3278 & \textbf{\textless{}0.0001} & 12.235 & \textbf{\textless{}0.0001} & 27.7169 & \textbf{\textless{}0.0001} & 31.26 & \textbf{\textless{}0.0001} \\
 & dpath & -1.0570 & 0.2906 & -7.0787 & \textbf{\textless{}0.0001} & 26.6229 & \textbf{\textless{}0.0001} & 17.5991 & \textbf{\textless{}0.0001} & 10.6613 & \textbf{\textless{}0.0001} & 30.3162 & \textbf{\textless{}0.0001} & 30.0842 & \textbf{\textless{}0.0001} & -14.3029 & \textbf{\textless{}0.0001} & 21.8959 & \textbf{\textless{}0.0001} & 13.921 & \textbf{\textless{}0.0001} \\
 & cons & 1.9370 & 0.0572 & 2.6935 & 0.7973 & 4.5987 & 0.0090 & 4.4060 & \textbf{\textless{}0.0001} & 0.2580 & \textbf{\textless{}0.0001} & 4.6524 & \textbf{\textless{}0.0001} & 4.0888 & \textbf{0.0001} & 0.9253 & 0.3584 & 4.6578 & \textbf{\textless{}0.0001} & 13.091 & \textbf{\textless{}0.0001} \\
 & pann & 1.4672 & 0.1451 & 2.1132 & \textbf{0.0368} & 3.8403 & \textbf{0.0002} & 4.0339 & \textbf{0.0001} & 1.2108 & 0.2285 & 7.9571 & \textbf{\textless{}0.0001} & 4.9897 & \textbf{\textless{}0.0001} & 1.6873 & 0.0943 & 5.3082 & \textbf{\textless{}0.0001} & 16.686 & \textbf{\textless{}0.0001} \\
 & crag & 0.8642 & 0.3876 & 5.0746 & \textless{}0.0001 & 18.4799 & \textbf{\textless{}0.0001} & 12.1518 & \textbf{\textless{}0.0001} & 8.0579 & \textbf{\textless{}0.0001} & 22.1590 & \textbf{\textless{}0.0001} & 27.0217 & \textbf{\textless{}0.0001} & 9.7992 & \textbf{\textless{}0.0001} & 32.5868 & \textbf{\textless{}0.0001} & 33.256 & \textbf{\textless{}0.0001} \\ \hline
\multirow{5}{*}{\textbf{mask-rcnn-resnet50-fpn  \textgreater{} mask-rcnn-resnet50-fpn-srm}} & glas & 6.8530 & \textbf{\textless{}0.0001} & 9.8997 & \textbf{\textless{}0.0001} & 23.0282 & \textbf{\textless{}0.0001} & 10.4970 & \textbf{\textless{}0.0001} & 12.1580 & \textbf{\textless{}0.0001} & 37.4541 & \textbf{\textless{}0.0001} & 26.2797 & \textbf{\textless{}0.0001} & 12.0827 & \textbf{\textless{}0.0001} & 26.6727 & \textbf{\textless{}0.0001} & 31.115 & \textbf{\textless{}0.0001} \\
 & dpath & -0.4312 & 0.6639 & -9.2711 & \textbf{\textless{}0.0001} & 19.4217 & \textbf{\textless{}0.0001} & 15.8981 & \textbf{\textless{}0.0001} & 10.7513 & \textbf{\textless{}0.0001} & 28.8324 & \textbf{\textless{}0.0001} & 26.9082 & \textbf{\textless{}0.0001} & -13.7927 & \textbf{\textless{}0.0001} & 18.5482 & \textbf{\textless{}0.0001} & 10.098 & \textbf{\textless{}0.0001} \\
 & cons & 1.5949 & 0.1157 & 2.9231 & 0.0048 & 4.2319 & \textbf{\textless{}0.0001} & 4.3483 & \textbf{\textless{}0.0001} & 0.4117 & 0.6820 & 4.2142 & \textbf{\textless{}0.0001} & 4.4034 & \textbf{\textless{}0.0001} & 1.3173 & 0.1925 & 5.0382 & \textbf{\textless{}0.0001} & 13.621 & \textbf{\textless{}0.0001} \\
 & pann & 2.0367 & \textbf{0.0441} & 1.8431 & \textbf{0.0680} & 2.4267 & \textbf{0.0168} & 4.8150 & \textbf{\textless{}0.0001} & 1.4769 & \textbf{0.1425} & 5.6564 & \textbf{\textless{}0.0001} & 5.5451 & \textbf{\textless{}0.0001} & 1.1294 & 0.2611 & 5.6305 & \textbf{\textless{}0.0001} & 16.707 & \textbf{\textless{}0.0001} \\
 & crag & 3.2289 & \textbf{0.0013} & 1.6125 & \textbf{0.1070} & 13.3727 & \textbf{\textless{}0.0001} & 8.6679 & \textbf{\textless{}0.0001} & 9.4167 & \textbf{\textless{}0.0001} & 16.9443 & \textbf{\textless{}0.0001} & 25.5689 & \textbf{\textless{}0.0001} & 9.1502 & \textbf{\textless{}0.0001} & 30.5842 & \textbf{\textless{}0.0001} & 27.634 & \textbf{\textless{}0.0001} \\ \hline
\end{tabular}
\end{adjustbox}
\captionof{table}{\label{tab:significance_mask_rcnn} Paired samples t-test for the Mask-RCNN experiments at the level of the \ac{FPN}. Values in bold indicate significant differences (p-value $<0.05$). The experiments for Mask-RCNN with context and attention mechanisms at the level of the ResNet-50 backbone always demonstrated significant differences (p-value $<0.0001$). The initials ``neu'', ``epi'', ``lym'', ``pls'', ``eos'', and ``con'' indicate $m\mathcal{PQ}^+$ for each nucleus type.}
\end{table*}

\subsection{Methods}

We first consider Mask-RCNN \citep{He2017} with a ResNet-50 \citep{He2016} backbone and \ac{FPN} \citep{Lin2017} as a baseline model (Mask-RCNN-ResNet50-FPN). We then propose two variants of Mask-RCNN \citep{He2017} with attention-based mechanisms. On the one hand, we add a \ac{SE}-block \citep{Hu2018} before every lateral connection of the \ac{FPN} module, which we call the \ac{SE}-\ac{FPN} (Mask-RCNN-ResNet50-SE-FPN). We hypothesised that this feature refinement step between the \ac{FPN} top-down and bottom-up paths would reduce the \ac{FPN} sensitivity to noise from non-foreground objects, especially in lower-resolution feature maps. Second, we consider an object relation network, without geometric features, that we term the \ac{SRM} (Mask-RCNN-ResNet50-FPN-SRM). The \ac{SRM}  considers a transformer-encoder to replace the second fully connected layer of the \ac{MLPHead}. The objective was to enrich the semantics of every object candidate with cross-attention and long range-dependencies, while also exploring the potential of hybridizing \acp{ConvNet} with transformer neural networks. Figs. \ref{fig:SEFPN} and \ref{fig:SRM} illustrate, respectively, the proposed \ac{SE}-\ac{FPN} and \ac{SRM}. In addition, not only do we experiment with replacing the ResNet-50 backbone of Mask-RCNN with \ac{SE}-blocks, the designated Mask-RCNN-SE-ResNet50-FPN, as also suggest extending the last convolutional block of ResNet-50 with a \ac{SRM}, the so-called Mask-RCNN-SRM-ResNet50-FPN.

But a known limitation of Mask-RCNN \citep{He2017} is the difficulty in detecting small instances and that are densely distributed in challenging contexts \citep{Min2022}. Therefore, we also extend the domain-specific HoVer-Net \cite{Graham2019b} with context and attention mechanisms to understand if it would compare favourably with the Mask-RCNN-based models and if incorporating attention and context would increase generalization. As before, we extend the HoVer-Net ResNet-50 backbone with the \ac{SE}-blocks at every block and with a \ac{SRM} at the lowest level, that we term, respectively, SE-HoVer-Net and SRM-HoVer-Net.

\begin{table*}[h]
\captionsetup{font=scriptsize}
\begin{adjustbox}{width=1\textwidth}
\begin{tabular}{cccccccccccc}
    \multicolumn{12}{c}{\textbf{Comparison of HoVer-Net with SE-HoVer-Net, and SRM-HoVer-Net}} \\ \hline
\textbf{model} &
  \textbf{val set} &

  \multicolumn{1}{c}{\textbf{neu}} &
  \multicolumn{1}{c}{\textbf{epi}} &
  \multicolumn{1}{c}{\textbf{lym}} &
  \multicolumn{1}{c}{\textbf{pls}} &
  \multicolumn{1}{c}{\textbf{eos}} &
  \multicolumn{1}{c}{\textbf{con}} &
  \multicolumn{1}{c}{\textbf{pq}} &
  \multicolumn{1}{c}{\textbf{pq det}} &
  \multicolumn{1}{c}{\textbf{pq seg}} &
  \multicolumn{1}{c}{\textbf{multi pq+}} \\ \hline
\multirow{5}{*}{\textbf{hover-net}} &
  glas &
  0.0491 &
  0.4799 &
  0.5861 &
  0.2924 &
  0.1686 &
  0.3439 &
  0.5423 &
  0.6952 &
  0.7686 &
  0.3197 \\
 &
  dpath &
    0.0175 &
    0.3600 & 
    0.5225 &
    0.2688 &
    0.0762 &
    0.3895 &
    0.5359 &
    0.6798 &
    0.7797 &
    0.2724 \\
 &
  cons &

    0.1556 &
    0.499 &
    0.5941 &
    0.4075 &
    0.2219 &
    0.4587 &
    0.5322 &
    0.6840 &
    0.7635 &
    0.3895 \\
 &
  pann &

    0.2394	&
    0.5095	&
    0.5368	&
    0.3227	&
    0.0642	&
    0.4438	&
    0.5581	&
    0.7115	&
    0.7541	&
    0.3528 \\

 &
  crag &

    0.0172	&
    0.4574	&
    0.5623	&
    0.3646	&
    0.1031	&
    0.4044	&
    0.4860	&
    0.6169	&
    0.7359	&
    0.3182 \\
  & 
  overall & 

    0.0342	$\pm$	0.0880 &
    0.4539	$\pm$	0.0536  &
    0.5586	$\pm$	0.0272  &
    0.3239	$\pm$	0.0498  &
    0.1024	$\pm$	0.0597 & 
    0.4038	$\pm$	0.0408  &
    0.5297	$\pm$	0.0241 &
    0.6759	$\pm$	0.0322 &
    0.7601	$\pm$	0.0148 &
    0.3259 $\pm$	0.0390 
    
    \\
\hline 
\multirow{5}{*}{\textbf{se-hover-net}} &
  glas &

    0.1041	&
    0.4853	&
    0.5848	&
    0.3239	&
    0.1963	&
    0.3860	&
    0.5448	&
    0.6985	&
    0.7681	&
    0.3467  \\
 &
  dpath &

    0.0461	&
    0.4095	&
    0.5229	&
    0.3069	&
    0.1116	&
    0.4321	&
    0.5657	&
    0.7085	&
    0.7905	&
    0.3049  \\
 &
  cons &

    0.2132	&
    0.5055	&
    0.5883	&
    0.4017	&
    0.3887	&
    0.4664	&
    0.5386	&
    0.6928	&
    0.7758	&
    0.4273  \\
 &
  pann &

    0.1078 &
    0.5057 &
    0.5559 &
    0.3125 &
    0.1382 &
    0.4420 &
    0.5591 &
    0.7120 &
    0.7613 &
    0.3437 \\

 &
  crag &

    0.0297	&
    0.4443	&
    0.5600	&
    0.3944	&
    0.1017	&
    0.3971	&
    0.4871	&
    0.6195	&
    0.7335	&
    0.3211	\\
&
    overall & 
    0.0633 $\pm$ 0.0644 &
    0.4669 $\pm$ 0.0377 &	
    0.5614 $\pm$ 0.0236 &	
    0.3432 $\pm$ 0.0414 &	
    0.1484 $\pm$ 0.1059 &	
    0.4227 $\pm$ 0.0295 &	
    0.5375 $\pm$ 0.0277 &	
    0.6844 $\pm$ 0.0341 &	
    0.7654 $\pm$ 0.0189 &	
    0.3442 $\pm$ 0.0422  \\
  \hline
  \multirow{5}{*}{\textbf{srm-hover-net}} &
  glas &
    0.0994	&
    0.4826	&
    0.5838	&
    0.355	&
    0.1981	&
    0.3244	&
    0.5407	&
    0.6918	&
    0.7698	&
    0.3406 \\
 & 												

  dpath &

    0.0448 &	
    0.4143 &
    0.5199 &
    0.3002 &	
    0.1073 &	
    0.4288 &
    0.5661 &
    0.7122 &
    0.7869 &
    0.3026 \\
 &
  cons & 										
    0.1822 &	
    0.5115 &	
    0.5870 &	
    0.4482 &	
    0.3413 &	
    0.4704 &
    0.5465 &	
    0.7037 &
    0.7736 &	
    0.4234 \\
 &
  pann & 													

    0.2435	&
    0.5249  &	
    0.5772	&
    0.3459	&
    0.166	&
    0.4607	&
    0.5671	&
    0.7206	&
    0.7638	&
    0.3864  \\
 &
  crag & 											
    0.0203	&
    0.4868	&
    0.5705	&	
    0.4402	&	
    0.1235	&	
    0.4391	&	
    0.5065	&
    0.6420	&
    0.7388	&
    0.3467	\\
&
    overall & 

    0.0548	$\pm$ 0.0838 &
    0.4808	$\pm$ 0.0382 &
    0.5666	$\pm$ 0.0246 &
    0.3692	$\pm$ 0.0573 &
    0.1591	$\pm$ 0.0834 &
    0.4171	$\pm$ 0.0523 &
    0.5445	$\pm$ 0.0221 &
    0.6929	$\pm$ 0.0277 &
    0.7662	$\pm$ 0.0158 &
    \textbf{0.3552	$\pm$ 0.0414}  \\

     \hline
\end{tabular}
\end{adjustbox}
\captionof{table}{ \label{tab:se_vs_baseline_hovernet} Comparison of HoVer-Net with SE-HoVer-Net, and SRM-HoVer-Net in a colon nuclei instance segmentation and classification task. The initials ``neu'', ``epi'', ``lym'', ``pls'', ``eos'', and ``con'' indicate $m\mathcal{PQ}^+$ for each nucleus type. Results in bold indicate the best $m\mathcal{PQ}^+$.}
\end{table*}

\begin{table*}[h]
\captionsetup{font=scriptsize}
\begin{adjustbox}{width=1\textwidth}
\begin{tabular}{cccccccccccccccccccccc}
    \multicolumn{20}{c}{\textbf{Paired Samples T-Test for the HoVer-Net Experiments}} \\ \hline
\multirow{2}{*}{\textbf{null hypothesis}} & \multirow{2}{*}{\textbf{val set}} & \multicolumn{2}{c}{\textbf{neu}} & \multicolumn{2}{c}{\textbf{epi}} & \multicolumn{2}{c}{\textbf{lym}} & \multicolumn{2}{c}{\textbf{pls}} & \multicolumn{2}{c}{\textbf{eos}} & \multicolumn{2}{c}{\textbf{con}} & \multicolumn{2}{c}{\textbf{pq\_det}} & \multicolumn{2}{c}{\textbf{pq\_seg}} & \multicolumn{2}{c}{\textbf{pq}} & \multicolumn{2}{c}{\textbf{mpq+}} \\ \cline{3-22} 
 &  & \textbf{t-stat} & \textbf{p-val} & \textbf{t-stat} & \textbf{p-val} & \textbf{t-stat} & \textbf{p-val} & \textbf{t-stat} & \textbf{p-val} & \textbf{t-stat} & \textbf{p-val} & \textbf{t-stat} & \textbf{p-val} & \textbf{t-stat} & \textbf{p-val} & \textbf{t-stat} & \textbf{p-val} & \textbf{t-stat} & \textbf{p-val} & \textbf{t-stat} & \textbf{p-val} \\ \hline
\multirow{5}{*}{\textbf{se-hover-net \textgreater{} hover-net}} & glas & 3.1444 & \textbf{0.0017} & 1.7008 & 0.0894 & -0.0829 & 0.9339 & 0.4781 & 0.6327 & 1.4526 & 0.1468 & 11.9248 & \textbf{\textless{}0.0001} & 2.5512 & \textbf{0.0109} & -1.1093 & 0.2677 & \multicolumn{1}{l}{2.5790} & \multicolumn{1}{l}{\textbf{0.0101}} & \multicolumn{1}{l}{6.4410} & \multicolumn{1}{l}{\textbf{\textless{}0.0001}} \\
 & dpath & 2.7979 & \textbf{0.0052} & 18.9057 & \textbf{\textless{}0.0001} & -8.4985 & \textbf{\textless{}0.0001} & 10.3395 & \textbf{\textless{}0.0001} & 4.3424 & \textbf{\textless{}0.0001} & 16.2891 & \textbf{\textless{}0.0001} & 23.8710 & \textbf{\textless{}0.0001} & 27.0760 & \textbf{\textless{}0.0001} & 30.4551 & \textbf{\textless{}0.0001} & 15.5477 & \textbf{\textless{}0.0001} \\
 & cons & 1.2768 & 0.2064 & 0.6534 & 0.5159 & 0.9567 & 0.3424 & -0.0828 & 0.9343 & 3.1539 & \textbf{0.0025} & 0.4014 & 0.6895 & 0.8514 & 0.3970 & 1.0310 & 0.3064 & 0.8308 & 0.4092 & 3.2680 & \textbf{0.0018} \\
 & pann & -0.4930 & 0.6230 & -1.5815 & 0.1166 & -0.9473 & 0.3455 & -2.1102 & \textbf{0.0371} & 2.2412 & \textbf{0.0270} & 1.1240 & 0.2634 & 0.0690 & 0.9446 & 1.0641 & 0.2896 & 0.1777 & 0.8593 & 0.1069 & 0.9150 \\
 & crag & 3.4205 & \textbf{0.0006} & -7.0200 & \textbf{\textless{}0.0001} & -7.4141 & \textbf{\textless{}0.0001} & 2.8204 & \textbf{0.0048} & -1.1873 & 0.2352 & 0.3356 & 0.7372 & 1.5795 & 0.1144 & -2.1720 & \textbf{0.0299} & 0.9240 & 0.3556 & -4.0071 & \textbf{\textless{}0.0001} \\ \hline
\multirow{5}{*}{\textbf{srm-hover-net \textgreater{} se-hover-net}} & glas & -3.1949 & \textbf{0.0015} & -0.8353 & 0.4038 & 4.8248 & \textbf{\textless{}0.0001} & 1.4852 & 0.1379 & -3.2505 & \textbf{0.0012} & -16.7237 & \textbf{\textless{}0.0001} & -4.6036 & \textbf{\textless{}0.0001} & 3.6949 & \textbf{0.0002} & -3.8037 & \textbf{0.0002} & -5.6549 & \textbf{\textless{}0.0001} \\
 & dpath & 0.5955 & 0.5516 & 1.5707 & 0.1164 & -5.4150 & \textbf{\textless{}0.0001} & -5.0276 & \textbf{\textless{}0.0001} & -0.0069 & 0.9945 & -1.6899 & 0.0912 & 3.2503 & \textbf{0.0012} & 10.0651 & \textbf{\textless{}0.0001} & 0.5211 & 0.6023 & -4.7306 & \textbf{\textless{}0.0001} \\
 & cons & -1.4960 & 0.1396 & 0.1797 & 0.8580 & -1.4831 & 0.1430 & 0.4376 & 0.6632 & -1.2644 & 0.2107 & 1.3096 & 0.1951 & 2.0413 & \textbf{0.0454} & -0.6245 & 0.5345 & 1.8040 & 0.0760 & -0.6977 & 0.4880 \\
 & pann & 1.3364 & 0.1841 & 3.5645 & \textbf{\textless{}0.0001} & 2.8376 & \textbf{0.0054} & 0.4112 & 0.6817 & 0.7398 & 0.4610 & 0.6538 & 0.5146 & 2.0001 & \textbf{0.0479} & 1.0702 & 0.2869 & 2.3040 & \textbf{0.0231} & 2.8960 & \textbf{0.0045} \\
 & crag & -1.9654 & \textbf{0.0495} & 16.6869 & \textbf{\textless{}0.0001} & 4.7051 & \textbf{\textless{}0.0001} & 6.7274 & \textbf{\textless{}0.0001} & 3.5850 & \textbf{\textless{}0.0001} & 13.1578 & \textbf{\textless{}0.0001} & 13.6052 & \textbf{\textless{}0.0001} & 4.8841 & \textbf{\textless{}0.0001} & 14.8626 & \textbf{\textless{}0.0001} & 15.0817 & \textbf{\textless{}0.0001} \\ \hline
\multirow{5}{*}{\textbf{srm-hover-net  \textgreater{} hovernet}} & glas & -0.8114 & 0.4174 & 0.8109 & 0.4177 & 4.9793 & \textbf{\textless{}0.0001} & 2.3828 & \textbf{0.0174} & -2.2134 & \textbf{0.0272} & -6.5097 & \textbf{\textless{}0.0001} & -2.7318 & \textbf{0.0065} & 2.1846 & \textbf{0.0292} & -1.8074 & 0.0711 & -0.2850 & 0.7750 \\
 & dpath & 3.3095 & \textbf{\textless{}0.0001} & 20.8255 & \textbf{\textless{}0.0001} & -12.8211 & \textbf{\textless{}0.0001} & 5.0426 & \textbf{\textless{}0.0001} & 5.3716 & \textbf{\textless{}0.0001} & 15.6021 & \textbf{\textless{}0.0001} & 28.4857 & \textbf{\textless{}0.0001} & 19.8131 & \textbf{\textless{}0.0001} & 32.5700 & \textbf{\textless{}0.0001} & 10.2096 & \textbf{\textless{}0.0001} \\
 & cons & 0.2614 & 0.7946 & 0.8104 & 0.4207 & -0.2047 & 0.8384 & 0.4504 & 0.6540 & 2.8279 & \textbf{0.0063} & 1.4373 & 0.1556 & 2.0012 & \textbf{0.0497} & 1.1079 & 0.2721 & 2.2837 & \textbf{0.0258} & 12.7128 & \textbf{\textless{}0.0001} \\
 & pann & 1.2746 & 0.2051 & 3.5341 & \textbf{0.0006} & 1.9661 & 0.0518 & -1.1428 & 0.2556 & 2.7019 & \textbf{0.0080} & 1.3939 & 0.1661 & 1.3300 & 0.1862 & 1.7367 & 0.0852 & 2.0335 & \textbf{0.0444} & 2.8870 & \textbf{0.0047} \\
 & crag & 1.1424 & 0.2534 & 14.2186 & \textbf{\textless{}0.0001} & -2.3907 & \textbf{0.0169} & 9.5351 & \textbf{\textless{}0.0001} & 2.2800 & \textbf{0.0227} & 12.9627 & \textbf{\textless{}0.0001} & 14.1342 & \textbf{\textless{}0.0001} & 2.7589 & \textbf{0.0058} & 14.5832 & \textbf{\textless{}0.0001} & 1.9801 & 0.0521 \\ \hline
\end{tabular}
\end{adjustbox}
\captionof{table}{\label{tab:significance_hovernet} Paired samples t-test for the HoVer-Net experiments. Values in bold indicate significant differences (p-value $<0.05$). The initials ``neu'', ``epi'', ``lym'', ``pls'', ``eos'', and ``con'' indicate $m\mathcal{PQ}^+$ for each nucleus type.}
\end{table*}

\begin{table*}[h]
\captionsetup{font=scriptsize}
\begin{adjustbox}{width=1\textwidth}
\begin{tabular}{cccccccccccc}
\multicolumn{12}{c}{\textbf{Results: \ac{iid} validation. Data source: CRAG \cite{Graham2019a}}} \\ \hline
  $\mathbf{map_{50}}$ \textbf{bbox} &
  $\mathbf{map_{50}}$ \textbf{segm} &
  \multicolumn{1}{c}{\textbf{neu}} &
  \multicolumn{1}{c}{\textbf{epi}} &
  \multicolumn{1}{c}{\textbf{lym}} &
  \multicolumn{1}{c}{\textbf{pls}} &
  \multicolumn{1}{c}{\textbf{eos}} &
  \multicolumn{1}{c}{\textbf{con}} &
  \multicolumn{1}{c}{\textbf{pq}} &
  \multicolumn{1}{c}{\textbf{pq det}} &
  \multicolumn{1}{c}{\textbf{pq seg}} &
  \multicolumn{1}{c}{\textbf{multi pq+}} \\ \hline
\multicolumn{12}{c}{\textbf{mask-rcnn-resnet50-fpn}} \\ \hline
 0.781 $\pm$ 0.090 & 0.804 $\pm$ 0.077 & 0.256 $\pm$ 0.057 & 0.561 $\pm$ 0.021 & 0.551 $\pm$ 0.032 & 0.423 $\pm$ 0.068 & 0.367 $\pm$ 0.072 & 0.549 $\pm$ 0.038 & 0.557 $\pm$ 0.028 & 0.713 $\pm$ 0.034 & 0.734 $\pm$ 0.034 & 0.451 $\pm$ 0.026 \\
\hline
\multicolumn{12}{c}{\textbf{mask-rcnn-resnet50-se-fpn}} \\ \hline
0.806 $\pm$ 0.074 & 0.825 $\pm$ 0.084 & 0.230 $\pm$ 0.047 & 0.559 $\pm$ 0.025 &	0.544 $\pm$ 0.045 & 0.437 $\pm$ 0.058 &	0.352 $\pm$ 0.081 & 0.564 $\pm$ 0.019 &	0.555 $\pm$ 0.031 & 0.711	$\pm$ 0.038 & 0.733 $\pm$ 0.033 & 0.448 $\pm$ 0.024 \\
										
 \hline
\end{tabular}
\end{adjustbox}
\captionof{table}{\label{tab:results_crag} Results of the Mask-RCNN-ResNet50-FPN, and Mask-RCNN-ResNet50-SE-FPN when trained and evaluated on the same data source. The initials ``neu'', ``epi'', ``lym'', ``pls'', ``eos'', and ``con'' indicate $m\mathcal{PQ}^+$ for each nucleus type.}
\end{table*}

\subsubsection*{Implementation Details}

The Mask-RCNN experiments were implemented in \emph{python} with \emph{Pytorch} version 1.10.2. For the Mask-RCNN-ResNet50-FPN, Mask-RCNN-ResNet50-SE-FPN, and Mask-RCNN-SE-ResNet50-FPN we observed that a batch size of 1 resulted in better results, with a learning rate of 0.001, weight decay equal to $5 \times 10^{-4}$, and Stochastic Gradient Descent (\emph{SGD}) optimizer (momentum=0.9). For the Mask-RCNN-Resnet-50-FPN-SRM experiments, we considered 8 attention heads in the transformer layer, with a dropout \citep{Srivastava2014} probability of 0.5 and GeLU activation \citep{Hendrycks2016}. We use the Adaptive Moment with decoupled weight decay (\emph{AdamW} \citep{Loshchilov2017}) optimizer with a learning rate of $5 \times 10^{-6}$ and weight decay equal to $5 \times 10^{-4}$. In turn, for the Mask-RCNN-SRM-ResNet50-FPN we observe a dropout probability of 0.4 is the optimal choice, with learning rate of $5 \times 10^{-5}$, weight decay equal to $5 \times 10^{-4}$, and \emph{SGD} optimizer (momentum=0.9). For all Mask-RCNN variants, we consider 100 traiming epochs and opt for a linear learning rate warm-up in the first 1000 iterations of the first epoch (starting at $1 \times 10^{-6}$ for Mask-RCNN-ResNet50-FPN,  Mask-RCNN-ResNet50-SE-FPN, and Mask-RCNN-SE-ResNet50-SE-FPN, and at $5 \times 10^{-9}$ for the Mask-RCNN-ResNet50-FPN-SRM and $5 \times 10^{-8}$ for Mask-RCNN-SRM-ResNet50-FPN). The learning rate was decreased by a factor of 0.1 if the training loss did not improve for 10 epochs. A simple data augmentation policy of random horizontal flips with a probability of 0.5 was considered. To deal with data imbalance in the classification task, we resorted to the focal loss \cite{Lin2018}, weighted by the intra-tile effective number of samples \citep{Cui2019} (equation~\ref{eq:iifocal_loss}) which promotes the cross-entropy loss to give a higher penalty for harder examples and for the instances that are least represented in a tile.

\begin{equation}
\label{eq:iifocal_loss}
    FL (p_t) = - \frac{1-\beta}{1-\beta^n}(1-p_t)^\gamma log(p_t)
\end{equation}

With $t$ the target class, $n$ the number of intra-tile instances of class $t$, $\beta$ and $\gamma$ hyperparameters, and $p_t$ the posterior probability, $ p(t | X)$. In our experiments, we set $\beta=0.9$ and $\gamma=0.9$. For validation, we selected the model saved at the last epoch.

Regarding the HoVer-Net experiments, we replicate the baseline from the official code base provided for the CoNIC Challenge \citep{graham2023conic}, only extending the method with the SE-blocks and SRM as previously described. We train the model for 50 epochs, and observe the best results with batch sizes of 2 and 6 for, respectively, HoVer-Net/SE-HoVer-Net and SRM-HoVer-Net, with Adaptive Moment Estimation (\emph{ADAM}) as optimizer and a learning rate of $1 \times 10^{-4}$. Besides, the learning rate is decreased by a factor of 0.1 every 25 epochs. As is done with the Mask-RCNN experiments, we replace the standard cross-entropy loss with the focal loss. Furthermore, in SRM-HoVer-Net we use 8 attention heads and dropout probability of 0.2.

The source code will be made publicly available upon acceptance.

\subsection{Results}

Tables \ref{tab:se_vs_baseline}, \ref{tab:se_vs_baseline_hovernet}, and \ref{tab:results_crag} present the results of the models evaluated in this work. In turn, tables \ref{tab:significance_mask_rcnn} and \ref{tab:significance_hovernet} present the results of a paired samples t-test. The experiments for Mask-RCNN with context and attention mechanisms at the level of the ResNet-50 backbone always demonstrated significant differences (p-value $<0.0001$).

\subsection{Discussion}

Pathologists rely on contextual information, at multiple levels, when analysing \acp{WSI}. For instance, they could consider the relationship between tissues and cells \citep{Ryu2023}, the co-occurrence of the same nuclei types \citep{Hassan2022}, or the correlation of tissue subregions with high-diagnostic value \citep{Shao2021}. Therefore, extending \acp{ANN} with context- and attention-based inductive biases could translate to more discriminative features. In this work, we extend Mask-RCNN and HoVer-Net with context and attention-based mechanisms to assess if it could improve the performance and generalization of the models. However, while most authors suggest extending \acp{ConvNet} with context and attention-based mechanisms improves segmentation performance, our findings are inconclusive. While in the HoVer-Net experiments, we observe the \ac{SE}-blocks and \ac{SRM} improve $m\mathcal{PQ}^+$ over the baseline by, respectively, 1.83 \% and 2.93 \%, the results in Mask-RCNN are not stable. Indeed, this is confirmed by the inconsistency of the significance tests.

This is contrary to our intuition that \ac{HE}-stained \acp{WSI} are context-rich and that incorporating context and attention-based inductive bias into algorithm design would enrich the learned representations. Furthermore, the surveyed papers demonstrate performance improvement in instance segmentation and classification with context and attention-based mechanisms \citep{Vahadane2021, ChenZhao2022, Dogar2023}, which is not consistent with our observations. As a matter of fact, the results suggest that depending on how context and attention modules are integrated into the neural network architecture, these could even deteriorate performance. Nonetheless, the performance degradation could be due to overfitting as we could end with an overparameterized network relative to dataset size.

Our findings thus suggest extending state-of-the-art models with simple context and attention-based mechanisms is not sufficient, and that more tailored solutions are required. Incorporating domain knowledge in algorithm design is thus no trivial task, which requires extensive experimentation, or even explicit context annotations. However, the lack of datasets with annotations referring to nuclei context (e.g., tissues, cells, etc.) hinders our ability to explicitly model contextual information. Indeed, while there is a recent effort towards annotating the context of cells (i.e. tissues (hierarchical/global context)) for improving cell detection algorithms \citep{Ryu2023}, we find there is no similar approach for nuclei instance segmentation and classification, besides that other levels of explicit contextual information could also be explored (e.g., long-range context).  On the other hand, there is the hypothesis that when the data semantics is context-rich, traditional \acp{ConvNet} could already learn contextual representations (e.g., through the \acp{FPN}), meaning extending these works with explicit context mechanisms (e.g., transformer layers) would not provide additional information. This hypothesis should be addressed in future work to help clarify these yet unanswered topics while contributing to improving algorithm performance and trustworthiness. It would provide some new insights regarding the inner workings of \acp{ANN} that could guide future developments towards more precise and reliable \ac{ML} in \ac{CPath}.

\section{Conclusion}
\label{sec:conclusion}

We have reviewed context and attention-based neural networks in medical image analysis, with a special focus on  cell nuclei instance segmentation and classification. Despite frequent claims that these mechanisms increase model performance, we found most works regard semantic segmentation, whereas the performance is usually lower in instance segmentation and classification. State-of-the-art models are frequently based on encoder-decoder neural networks, even in transformer neural networks as these often correspond to hybrids with \ac{ConvNet} encoder and transformer decoder, or with transformers cross-attention incorporated in the \ac{ConvNet} layers. Moreover, we find that the challenge of learning with limited annotations is being addressed in \ac{CPath}, where several works suggest exploring domain knowledge in the form of context and attention mechanisms to tackle the gap when compared with the fully supervised setting. Overall, context and attention-based mechanisms are promising approaches to solve many of the challenges of cell nuclei instance segmentation and classification from \ac{HE}-stained brightfield microscopy images like mitigating the effects of spurious contexts (e.g., low contrast between foreground and background instances) or refine instance segmentation masks of touching and overlapping instances. On the other hand, our comparative study of context and attention mechanisms does not support the hypothesis that simple context and attention-based models could increase the performance and generalization of nuclei instance segmentation and classification tasks, meaning more tailored solutions would be required. How to optimally exploit this domain knowledge thus remains an open question but there is also the hypothesis that state-of-the-art \acp{ConvNet} (e.g. Mask-RCNN \citep{He2017}) naturally learn contextual features from the training data. Moreover, while there is work demonstrating \acp{ConvNet}  can achieve the same results as models with an intrinsic context and attention design, namely \ac{ViT},\citep{Liu2022} a more detailed analysis could allow us to more objectively address this topic, especially in cell nuclei instance segmentation and classification. Although pathologists rely on multiscale context while paying attention to specific \acp{RoI} when analysing and annotating \acp{WSI}, our findings suggest translating that domain knowledge into \acp{ANN} requires careful design and extensive experimentation,  but to exploit these mechanisms fully, the scientific understanding of these methods should first be addressed.

\section*{Acknowledgments}

The authors would like to thank Isabel Rio Torto, Pedro Costa and Professor Aurélio Campilho for their contributions to the experimental part of this work, for their ideas and critical perspective.

This work is financed by National Funds through the Portuguese funding agency, FCT - Fundação para a Ciência e a Tecnologia, within the PhD grant 2022.12385.BD.

\section*{List of Acronyms}

\begin{acronym}[WSI]

\acro{ACF}{Adaptive Context Fusion}
\acro{AFPN}{Attention Feature Pyramid Network}
\acro{AHD}{Average Hausdorff Distance}
\acro{AI}{Artificial Intelligence}
\acro{AJI}{Aggregated Jaccard Index}
\acro{ANN}{Artificial Neural Network}
\acro{AP}{Average Precision}
\acro{AUC}{Area Under the Curve}
\acro{BACH2018}[BACH 2018]{Breast Cancer Histology Challenge 2018}
\acro{CANet}[CA-Net]{Comprehensive Attention Convolutional Neural Network}
\acro{CBAM}{Convolutional Block Attention Module}
\acro{CCa}{Colon Cancer}
\acro{CEM}{Context Enhancement Module}
\acro{CL}{Curriculum Learning}
\acro{CPFNet}{Context Pyramid Fusion Network}
\acro{CPM17}[CPM 2017]{Computational Precision Medicine Challenge 2017}
\acro{CPath}{computational pathology}
\acro{CRA}{Colorectal Adenocarcinoma}
\acro{CRC}{Colorectal Cancer}
\acro{CT}{Computerized Tomography}
\acro{CV}{Computer Vision}
\acro{CWM}{Confidence-based Weighting Module}
\acro{CoNSeP}{Colorectal Nuclear Segmentation and Phenotype}
\acro{ConvNet}{Convolutional Neural Network}
\acro{DA}{Domain Adaptation}
\acro{DCA}{Detail Context Attention}
\acro{DCNet}[DC-Net]{Dual Context Network}
\acro{DC}{Dice Similarity Coefficient}
\acro{DDD}{Decreasing Distance Discretization}
\acro{DETR}{Detection Transformer}
\acro{DG}{Domain Generalization}
\acro{dH}{directed Hausdorff Distance}
\acro{DL}{Deep Learning}
\acro{DTA}{Dynamic Texture Attention}
\acro{DeiT}{Data Efficient Image Transformer}
\acro{DigestPath}{Digestive System Pathological Detection and Segmentation Challenge}
\acro{FACA}{Foreground Aware Co-Attention}
\acro{FPN}{Feature Pyramid Network}
\acro{FTN}{Fully Transformer Network}
\acro{GAN}{Generative Adversarial Network}
\acro{GNN}{Graph Neural Network}
\acro{GPG}{Global Pyramid Guidance}
\acro{GPU}{Graphics Processing Unit}
\acro{GlaS}{Gland Segmentation in Colon Histology Images Challenge}
\acro{HCC}{Hepatocellular Carcinoma}
\acro{HD}{Hausdorff Distance}
\acro{HE}[H\&E]{Haematoxylin and Eosin}
\acro{HoVer}{Horizontal-Vertical}
\acro{IAM}{Information Aggregation Module}
\acro{IEN}{Instance Encoding Network}
\acro{IHC}{Immunohistochemical}
\acro{iid}[IID]{independent and identically distributed}
\acro{IOU}[IoU]{Intersection Over Union}
\acro{IRNet}{Instance Relation Network}
\acro{IoU}{Intersection over Union}
\acro{KD}{Knowledge Distillation}
\acro{KUMAR}{ Multi-organ Nuclear Segmentation Challenge}
\acro{LSTM}{Long Short-term Memory}
\acro{mAP}{Mean Average Precision}
\acro{MIL}{Multiple Instance Learning}
\acro{MLPHead}[MLP-Head]{Multi-layer Perceptron Head}
\acro{MLP}{Multilayer Perceptron}
\acro{ML}{Machine Learning}
\acro{MPN}{Message Passing Network}
\acro{MPN}{Message Passing Network}
\acro{MRI}{Magnetic Resonance Imaging}
\acro{MSDU}{Multi-scale Dense Units}
\acro{NC}{Nuclear Classification}
\acro{NF}{Nuclear Foreground}
\acro{NLP}{Natural Language Processing}
\acro{NMS}{Non-Maximum Suppression}
\acro{NO}{Nuclear Ordinal Regression}
\acro{NP}{Nuclear Prediction}
\acro{NT}{Nuclear Type}
\acro{NeurIPS}{Neural Information Processing Systems}
\acro{OOD}{out of distribution}
\acro{PAIP2019}[PAIP 2019]{Pathology Artificial Intelligence Platform Challenge 2019}
\acro{PQ}{Panoptic Quality}
\acro{RCCA}{Recurrent Criss-cross Attention}
\acro{RPN}{Region Proposal Network}
\acro{ReLU}{Rectified Linear Unit}
\acro{ResGANet}{Residual Group Attention Network}
\acrodefplural{RoI}[RoIs]{Regions of Interest}
\acro{RoI}{Region of Interest}
\acro{SAM}{Segment Anything}
\acro{SAPF}{Scale-Aware Pyramid Fusion}
\acro{SCM}{Structural Causal Model}
\acro{SE}{Squeeze and Excitation}
\acro{SL}{Strongly Supervised Learning}
\acro{SPN}{Semantic Proposal Network}
\acro{SPP}{Spatial Pyramid Pooling}
\acro{SPT}{Spatial Pyramid Transformer}
\acro{SRM}{Semantic Relation Module}
\acro{SSL}{Self-supervised Learning}
\acro{SWT}{Sliding Window Tokenization}
\acro{SmSL}{Semi-supervised Learning}
\acro{TCGA}{The Cancer Genomic Atlas}
\acro{TIL}{Tumour-Infiltrating Lymphocyte}
\acro{VAE}{Variational Autoencoder}
\acro{ViT}{Vision Transformer}
\acro{WSI}{Whole Slide Image}
\acro{WSL}{Weakly Supervised Learning}
\acro{WSOL}{Weakly Supervised Object Localization and Segmentation}
\end{acronym}

\bibliographystyle{model2-names.bst}\biboptions{authoryear}
\bibliography{refs}

\end{document}